\documentclass[10pt,twocolumn,letterpaper]{article}

\usepackage[pagenumbers]{iccv} 

\usepackage{multirow}
\usepackage{adjustbox}
\usepackage{float}

\usepackage{textcomp}
\makeatletter
\renewcommand{\@fnsymbol}[1]{\textsubscript{\textbf{\large *}}}  
\makeatother

\definecolor{iccvblue}{rgb}{0.21,0.49,0.74}
\usepackage[pagebackref,breaklinks,colorlinks,allcolors=iccvblue]{hyperref}

\def\paperID{3188} 
\def\confName{ICCV}
\def\confYear{2025}

\title{nnInteractive: Redefining 3D Promptable Segmentation}

\author{Fabian Isensee\textsuperscript{1,4}\thanks{\small Contributed equally. Each co-first author may list themselves as lead author on their CV.}\;,
Maximilian Rokuss\textsuperscript{1,2}\footnotemark[1]\;,
Lars Krämer\textsuperscript{1,4}\footnotemark[1]\;,
Stefan Dinkelacker\textsuperscript{1},
Ashis Ravindran\textsuperscript{1},\\
Florian Stritzke\textsuperscript{5},
Benjamin Hamm\textsuperscript{1,3},
Tassilo Wald\textsuperscript{1,2,4},
Moritz Langenberg\textsuperscript{1,2},
Constantin Ulrich\textsuperscript{1},\\
Jonathan Deissler\textsuperscript{1,2},
Ralf Floca\textsuperscript{1},
Klaus Maier-Hein\textsuperscript{1,2,3,4,6,7}
\\
\and
\textsuperscript{1}German Cancer Research Center, Division of Medical Image Computing, Germany\\
\textsuperscript{2}Faculty of Mathematics and Computer Science and 
\textsuperscript{3}Medical Faculty  - Heidelberg University\\
\textsuperscript{4}Helmholtz Imaging, 
\textsuperscript{5}Department of Radiation Oncology, Heidelberg University Hospital, Germany\\
\textsuperscript{6}HIDSS4Health, Heidelberg
\textsuperscript{7}Pattern Analysis and Learning Group, Heidelberg University Hospital
\\
{\tt\small \{f.isensee, maximilian.rokuss\}@dkfz-heidelberg.de}
}

\begin{document}
\maketitle
\begin{abstract}
Accurate and efficient 3D segmentation is essential for both clinical and research applications. While foundation models like SAM have revolutionized interactive segmentation, their 2D design and domain shift limitations make them ill-suited for 3D medical images. 
Current adaptations address some of these challenges but remain limited, either lacking volumetric awareness, offering restricted interactivity, or supporting only a small set of structures and modalities.
Usability also remains a challenge, as current tools are rarely integrated into established imaging platforms and often rely on cumbersome web-based interfaces with restricted functionality.
We introduce nnInteractive, the first comprehensive 3D interactive open-set segmentation method. It supports diverse prompts—including points, scribbles, boxes, and a novel lasso prompt—while leveraging intuitive 2D interactions to generate full 3D segmentations. Trained on 120+ diverse volumetric 3D datasets (CT, MRI, PET, 3D Microscopy, etc.), nnInteractive sets a new state-of-the-art in accuracy, adaptability, and usability. Crucially, it is the first method integrated into widely used image viewers (e.g., Napari, MITK), ensuring broad accessibility for real-world clinical and research applications.
Extensive benchmarking demonstrates that nnInteractive far surpasses existing methods, setting a new standard for AI-driven interactive 3D segmentation. nnInteractive is publicly available: \href{https://github.com/MIC-DKFZ/napari-nninteractive}{Napari plugin}, \href{https://www.mitk.org/MITK-nnInteractive}{MITK integration}, \href{https://github.com/MIC-DKFZ/nnInteractive}{Python backend}.
\end{abstract}

\begin{figure}[t]
    \begin{center}
        \includegraphics[width=\linewidth]{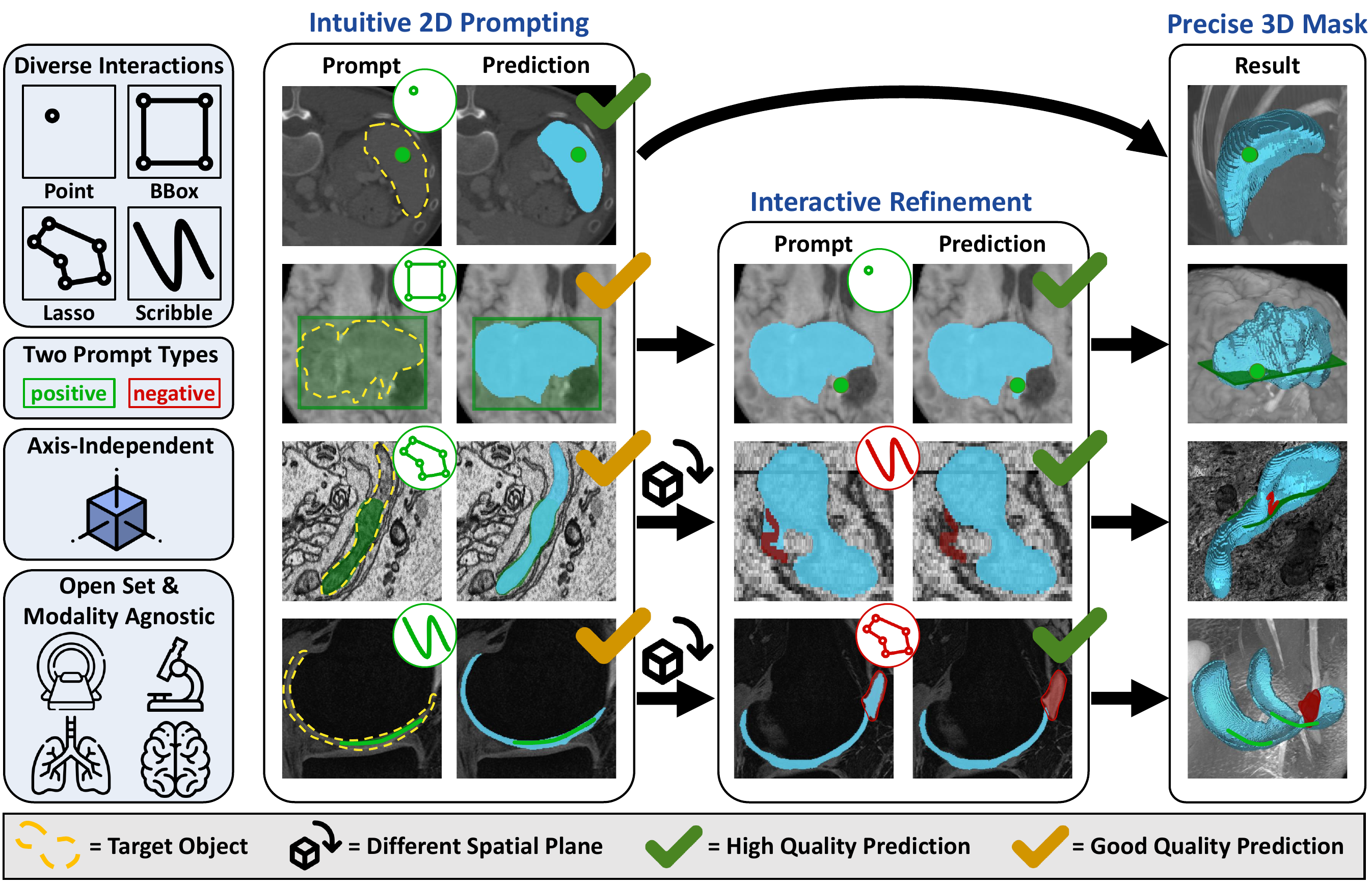}
    \end{center}
    \vspace{-0.5cm}
    \caption{\textbf{nnInteractive} fully unlocks the potential of 3D interactive segmentation. Supporting a diverse set of prompting styles, it generates full 3D segmentations from intuitive 2D interactions. Prompts can be arbitrarily mixed and placed on any axis. nnInteractive is open set and supports all modalities. It quickly adapts to user input to accurately segment any target structure.}
    \vspace{-0.5cm}
    \label{fig:the_real_figure_1}
\end{figure}
\section{Introduction}
\label{sec:intro}

Precise and efficient 3D segmentation is essential across fields such as medical imaging, biology, and industrial inspection, where volumetric data provides critical insights. Unlike 2D segmentation, which processes individual images, 3D segmentation must maintain volumetric consistency while handling high-dimensional data efficiently. This requires specialized methods that can adapt to diverse structures, imaging modalities, and real-world constraints. While fully automated segmentation models have achieved remarkable performance in specific tasks~\citep{isensee_nnu-net_2021, mednext, huang2023stunet,nnunet_revisited,multitalent}, their reliance on predefined training labels and distributions limits generalization to unseen structures and domains.  As a result, interactive segmentation has gained traction as a way to integrate user guidance, enhancing flexibility, accuracy, and real-world applicability.


\noindent Recent advances in foundation models, particularly interactive vision models like SAM~\citep{segment_anything,sam2}, enable general-purpose segmentation across diverse datasets via prompt-based interactions. By using simple inputs like points and bounding boxes these models reduce the need for task-specific training. SAM's direct application to medical imaging has shown promise~\cite{roy2023sammd}, which has driven a plethora of advancements in adapting interactive segmentation methods to the medical domain. Existing models~\citep{medsam, cheng2023sammed2d, scribbleprompt}, however, primarily operate on 2D slices, failing to account for the volumetric nature of medical scans. This leads to inconsistencies across slices and necessitates labor-intensive manual refinements when extending segmentations to full 3D volumes. Furthermore, models that do support volumetric segmentation are often limited to single imaging modalities such as computed tomography (CT)~\citep{segvol, vista3d, rokuss2025lesionlocator} or are constrained to closed-set segmentation, meaning they can only identify structures seen during training~\citep{vista3d, prism}. These limitations hinder their generalizability and practical usability, particularly in clinical workflows that demand adaptability to new and unseen structures and image acquitision protocols. Beyond these challenges, current interactive segmentation frameworks also suffer from restricted interaction paradigms. Most rely on simple prompts such as points and boxes~\citep{3dsamadapter, medsam}, with few methods supporting more intuitive inputs like scribbles~\citep{scribbleprompt}. The lack of user-friendly interactions further complicates usability, as existing tools often require cumbersome 3D bounding box annotations rather than leveraging natural 2D interactions for volumetric segmentation. Additionally, some methods do not support negative prompts~\cite{segvol} or interactive refinement~\cite{medsam}, both of which are essential for practical segmentation workflows. Finally, usability concerns persist, as existing solutions lack user-friendly interfaces integrated into established annotation platforms, limiting their adoption in real-world applications.

\noindent In this work, we introduce \texttt{nnInteractive}, a 3D interactive segmentation framework that systematically addresses key challenges in volumetric annotation. Rather than proposing new architectural innovations, our approach prioritizes usability, interaction diversity, positive and negative prompting, and computational efficiency. To ensure broad applicability and generalization, our model is trained on an unprecedentedly large and multimodal 3D dataset comprising over 120 publicly available datasets with 64,518 volumes spanning multiple imaging modalities and anatomical structures.
\noindent Beyond its technical capabilities, it is designed for seamless real-world adoption. It is integrated into widely used annotation platforms such as Napari~\citep{napari} and MITK Workbench~\citep{MITK}, providing an accessible and efficient tool for both research and clinical workflows. Extensive evaluation across diverse and out-of-distribution datasets demonstrates that \texttt{nnInteractive} establishes a new benchmark in 3D interactive segmentation, combining state-of-the-art performance with practical usability. We present the following \textit{key contributions:}

\begin{itemize}
    \item \texttt{nnInteractive} presents the first 3D interactive open-set segmentation model, supporting a wide range of positive and negative prompts (points, scribbles, bounding boxes, and lasso) across multiple imaging modalities (CT, MR, PET, etc.). It enables full 3D segmentation from intuitive 2D interactions, with an adaptive AutoZoom mechanism for large structures. It builds on nnU-Net’s best practices of carefully curated optimal design choices rather than a new novel architecture.
    \item Extensive benchmarking on 14 datasets shows superior segmentation accuracy, interactive refinement, and clinical usability, with the proposed lasso interaction providing the best guidance signal to the model.
    \item It is trained on an unprecedented scale with over 120 diverse 3D datasets, including a wide range of modalities, anatomical structures and label variations as well as novel SuperVoxels. Natural and simulated label variations enable nnInteractive to resolve segmentation ambiguities based on user intent.
    \item Optimized for real-world usability with \(<\)10 GB VRAM requirement, rapid inference and seamless integration into Napari and MITK Workbench.
\end{itemize}

\section{Method}
We design nnInteractive around three key principles: usability, interaction diversity, generalization and computational efficiency. This section details its network architecture, the transformation of 2D prompts into 3D masks, supported interaction types, user simulation, strategies to handle segmentation ambiguities, and our proposed Auto Zoom to predict targets that exceed the patch size of the model. Further details can be found in Appendix \ref{appendix:training_details}.

\begin{figure*}[t]
  \centering
  \includegraphics[width=0.9\textwidth]{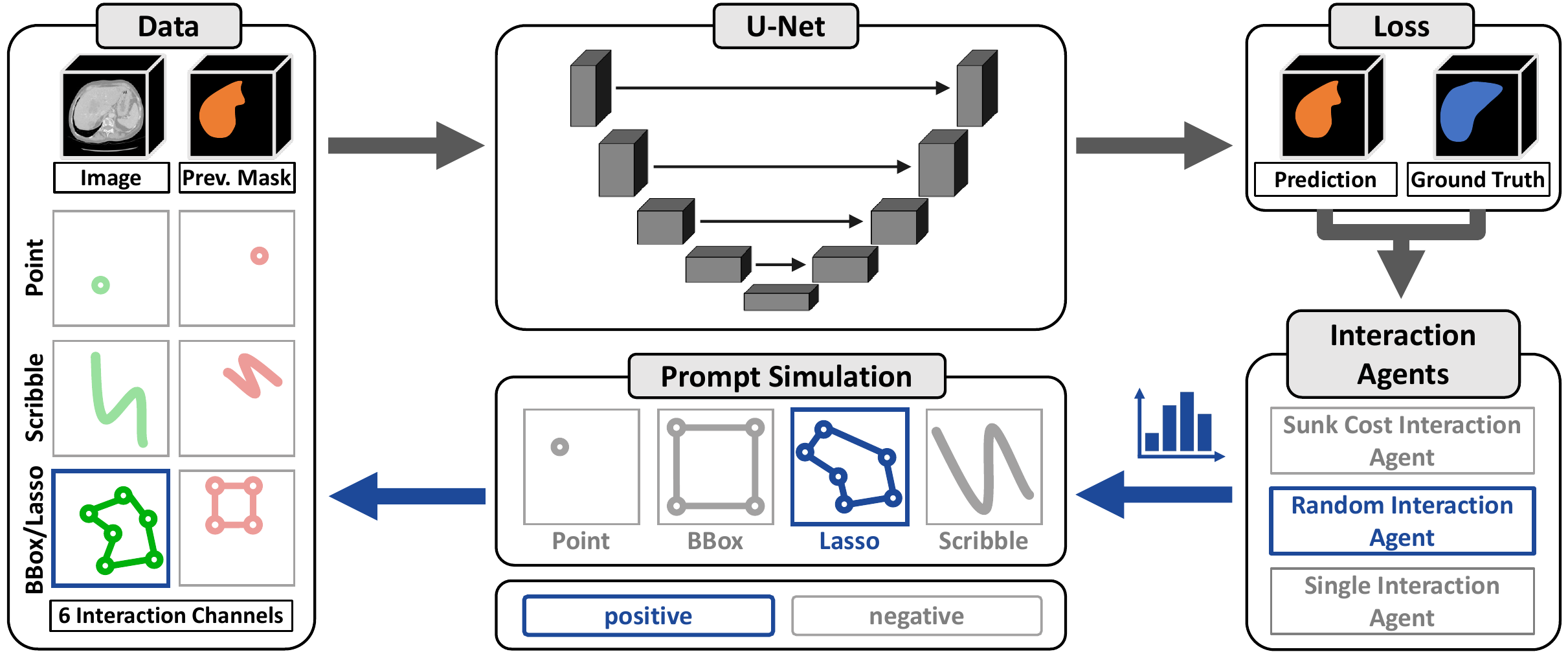}
  \caption{\textbf{Overview of the nnInteractive Training Pipeline.} The model first receives an input image and an initial prompt. The network then generates a prediction, which is used to compute the loss and identify false positive/negative areas. Based on the interaction agent simulating user input, a new prompt is sampled and added to the network input along with the current prediction.}
  \label{fig:nnInteractive_method_overview}
\end{figure*}

\subsection{Network Architecture}

Despite the widespread adoption of Transformer-based models in 2D computer vision, UNet architectures continue to dominate 3D medical image segmentation, consistently delivering state-of-the-art performance in benchmarks and competitions~\cite{bassi2024touchstonebenchmark, nnunet_revisited, uls_challenge, topcowchallenge, lnq2023challenge, autopet}. Given these advantages, we adopt a UNet-based design over Transformer alternatives and build upon the nnU-Net framework~\cite{isensee_nnu-net_2021} employing the Residual Encoder (ResEnc-L) configuration~\cite{nnunet_revisited} as our backbone.
\noindent A key distinction of our approach lies in how prompts are incorporated. Existing interactive segmentation models~\cite{segvol, sammed3d, sam2, segment_anything, medsam, cheng2023sammed2d} typically encode images first and then integrate user-provided prompts in latent space. While this approach has been effective in 2D leveraging large pretrained models, the absence of proper 3D foundation models renders this strategy suboptimal for 3D medical imaging. Instead, we adopt an early prompting strategy similar to \cite{scribbleprompt} (see Fig. \ref{fig:nnInteractive_method_overview}), where user inputs are directly incorporated as additional channels, ensuring prompts influence the entire feature extraction process. This allows the model to learn task-relevant representations from the highest resolution.

\subsection{From 2D Prompt to 3D Mask}

A core principle of nnInteractive is enhancing usability by bridging the gap between intuitive 2D annotation and full 3D segmentation. Modern 2D promptable models require separate prompts for each slice, and existing 3D approaches often rely on volumetric inputs such as 3D bounding boxes, which are challenging to define precisely from 2D views and can introduce segmentation errors due to excess empty space. nnInteractive efficiently predicts 3D masks from lower-dimensional prompts (see Fig. \ref{fig:the_real_figure_1}). Prompts can be either discrete points \( p \in \mathbb{R}^3 \) or structured 2D annotations \( p \in \mathbb{R}^{m \times n} \) (e.g., scribbles, bounding boxes, lassos). These prompts can be placed on any plane, from which nnInteractive generates a complete 3D segmentation mask, significantly minimizing annotation effort.

\subsection{Interaction Simulation}
\label{methods:interactions}
\begin{figure}[t]
  \centering
  \includegraphics[width=\linewidth]{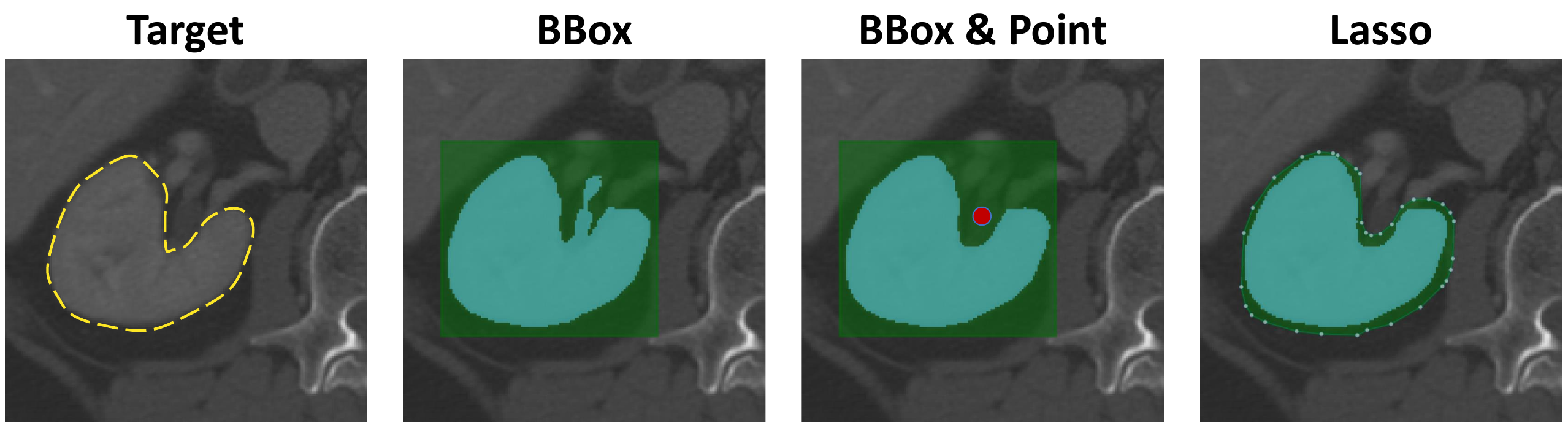}
  \caption{\textbf{Bounding Box vs. Lasso.} A bounding box interaction often requires additional refinement, whereas a fine-grained lasso interaction enables precise segmentation in a single step.}
  \label{fig:lasso}
\end{figure}



nnInteractive supports a comprehensive range of spatial prompt types, including points, bounding boxes, scribbles and lasso selections. 
Each prompt type is encoded in two separate input channels (positive and negative), see \cref{fig:nnInteractive_method_overview}. Lasso and bounding boxes share a pair of input channels. Inspired by Photoshop’s selection tool, the lasso prompt provides a more precise alternative to bounding boxes while requiring comparable annotation effort. Unlike bounding boxes, which often enclose irrelevant structures and necessitate additional refinement (\cref{fig:lasso}), lasso selections allow users to loosely outline the object without exact tracing.

\noindent We now describe nnInteractive’s prompt and interaction generation process. 
During training, the initial prompt simulation is derived from the ground truth, while subsequent interactions are guided by the current prediction error. Since predicted masks often contain both false positives (FP) and false negatives (FN), a connected-component \( V \) is first selected from either a FP or FN region. For lasso, 2D bounding boxes, and scribbles, a representative slice \( S \) is sampled from \( V \). The interaction is then simulated and assigned to the corresponding input channel. Upon adding a followup interaction, existing interactions are decayed by multiplying their intensity with 0.9. The network also receives its latest prediction as additional input.



\paragraph{Identifying Error Regions.}  
Error regions are identified by computing false positive (FP) and false negative (FN) areas from the difference between the ground truth \( y \) and prediction \( \hat{y} \). A random error component \( V \) is then selected with probability proportional to its size.  For bounding boxes and lassos, additional fragmentation is applied to prevent large, thin border regions—caused by slight over- or under-segmentation—from dominating the correction process. This is achieved by multiplying a thresholded Perlin noise mask, which breaks elongated structures before performing connected component analysis.  

\paragraph{2D Slice Sampling.}  
A 2D slice \( S \) is extracted from the selected 3D error component \( V \). Slices are sampled probabilistically based on the foreground volume distribution across axial, sagittal, and coronal planes, with a bias toward slices containing more foreground voxels. The chosen slice \( S \) serves as the foundation for generating scribbles, lassos, and bounding box interactions.  


\paragraph{Point Interactions.}
A Point interaction selects a representative location within the selected error component \( V \). We compute the normalized Euclidean Distance Transform (EDT) $D$ for all voxels $x\in V$, $D(x) \to [0,1]$ with $D$ assigning the highest values to central voxels. The point prompt location is then sampled using either a center-biased approach (\(\alpha=8\)) or uniform sampling (\(\alpha=1\)), with sampling probability:
\[
p(x) = \frac{D(x)^\alpha}{\sum_{z \in V} D(z)^\alpha}.
\]
The sampled point is expanded into a sphere which is then converted to a soft mask with maximum intensity at the center via an additional normalized EDT.

\paragraph{Bounding Box Interactions.}
Bounding boxes enclose the 2D error region in the selected slice \( S \). To introduce variability, the bounding box is randomly augmented:
\begin{itemize}
    \item \textbf{Jittering:} Each boundary is perturbed by an offset sampled from \( \mathcal{U}(-0.05, 0.05) \cdot d \), where \( d \) is the size of the bounding box in that dimension.
    \item \textbf{Shifting:} The entire box is translated within the same range as jittering.
    \item \textbf{Scaling:} A factor \( s \sim \mathcal{U}(0.8, 1.2) \) is applied per dimension, maintaining uniform scaling with probability 0.3.
\end{itemize}

\paragraph{Lasso Interactions.}
Lasso interactions provide a more precise alternative to bounding boxes, allowing users to loosely outline the object of interest. The lasso mask is generated in two steps:  
\begin{enumerate}
    \item \textbf{Coarse mask generation:} An enclosure of the 2D error mask  is formed using closing and dilation, where structuring element sizes are adapted to the object shape via the directional Euclidean distance transform (EDT).
    \item \textbf{Deformation:} A random displacement field is applied, with deformation magnitudes sampled proportionally to the directional EDT, ensuring realistic variability.
\end{enumerate}

\paragraph{Scribble Interactions.}
Inspired by ScribblePrompt~\cite{scribbleprompt}, nnInteractive generates three types of scribbles—center, line, and contour—each selected with equal probability. Unlike ScribblePrompt, which uses Perlin noise to break scribbles into multiple disconnected parts and thus regularly simulates multiple disconnected scribbles at once, our approach avoids such fragmentation by instead truncating scribbles at randomly sampled upper and lower coordinate bounds along each axis. 
Our implementation furthermore improves upon consistency by guaranteeing a fixed scribble width and improves parametrization by tying deformation parameters to object size.

\begin{itemize}
    \item \textbf{Center Scribbles:} Extracting the skeleton of the 2D error mask, then truncating to simulate partial annotation.
    \item \textbf{Line Scribbles:} Connecting two random points from the 2D error mask.
    \item \textbf{Contour Scribbles:} First eroding the 2D error mask and then computing the truncated contour of the eroded object.
\end{itemize}
Each scribble undergoes structured deformation using a random displacement field, where deformation magnitudes are sampled proportionally to the directional Euclidean distance transform (EDT). Skeletonization and dilation are then applied to maintain consistent thickness.




\subsection{User Simulation}

While nnInteractive supports diverse prompting styles, real users tend to follow consistent patterns rather than switching randomly. Disregarding this could lead to overfitting on unrealistic prompt strategies that do not reflect real-world usage. To maximize generalization, we introduce \textit{simulated user agents} that guide interaction sequences over multiple steps during training.
Overall, we introduce three agents:
\begin{enumerate*}[label=\roman*)]
    \item The \textit{Random agent}, selects a different prompting/interaction type at each iteration, representing a user who switches prompts at will. 
    \item The \textit{Sunk Cost agent} represents a user that prefers one interaction type and sticks with it for several iterations before switching to another. We simulate this with a  high probability of keeping the current prompting style and a low probability of randomly switching to a new one. 
    \item The \textit{Single Interaction agent} represents a user that strongly prefers one type of interaction and keeps it throughout the entire refinement process.
\end{enumerate*}

\subsection{Auto zoom}

\begin{figure}[t]
  \centering
  \includegraphics[width=\linewidth]{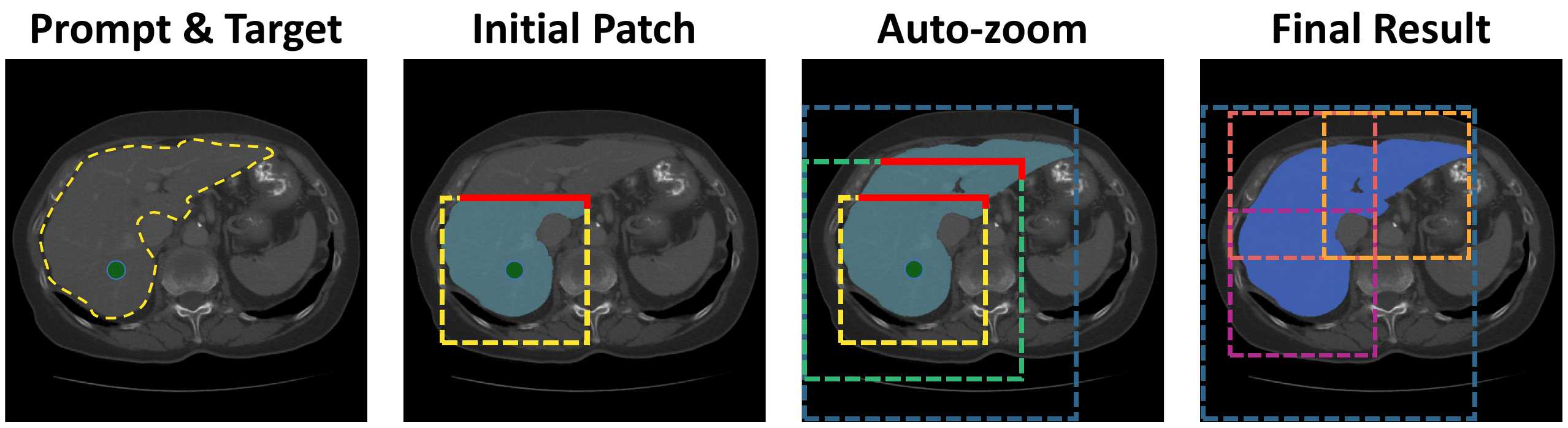}
  \caption{\textbf{Auto Zoom.} nnInteractive adaptively zooms out to ensure complete segmentation of large structures. By detecting border changes and dynamically querying additional regions, it preserves global context while refining local details, mitigating the constraints of patch-wise processing.}
  \vspace{-0.3cm}
  \label{fig:autozoom}
\end{figure}

One key limitation of 3D models is the need for patch-wise processing due to VRAM constraints. Without a dedicated mechanism, objects larger than the patch size get truncated at patch borders, leading to incomplete segmentation. Existing 3D methods either ignore this issue or rely on additional interactions to prompt new patches~\cite{vista3d, sammed3d}. SegVol~\cite{segvol} addresses this by first segmenting at a coarse resolution, heavily downscaling images to fit a \verb+32×256×256+ patch, and then refining predictions via a sliding window at full resolution.
We introduce a more adaptive auto zoom strategy (Fig.~\ref{fig:autozoom}). nnInteractive dynamically expands the region of interest (ROI) based on prediction borders, iteratively zooming out by a factor of 1.5 until the object is fully captured (up to 4× zoom out). The predicted low resolution mask is then resized to the original image resolution and refined with a sliding window approach. To optimize refinement, we process patches from most to least informative as measured by the amount of predicted foreground pixels they contain, ensuring the model starts in well-informed regions and incrementally extends to less constrained areas. The initial zoom factor is determined by the prompt that triggered it, ensuring the ROI encompasses the whole prompt plus a border of 1/6th of the patch size. This adaptive scheme minimizes computational overhead—small objects remain unaffected for faster inference, while large structures like the liver undergo progressive zoom and refinement. Unlike SegVol, nnInteractive dynamically adjusts the zoom level, preserving details, avoiding excessive downsampling for small objects, and maintaining compatibility with large images without exceeding VRAM limits.

\subsection{Ambiguity}

Interactive segmentation models must adapt to user inputs and segment structures based on user intent, which is often ambiguous. For example, in liver tumor segmentation, users may segment the liver with or without the tumor, or the tumor alone. Similarly, in cardiac cine-MRI, the left ventricle may be segmented with or without its lumen. Models trained with rigid predefined class definitions  struggle with such ambiguities (see Fig. \ref{fig:vista_fail_kidney_and_tumor}). To address this, nnInteractive is trained with randomly sampled label variations, exposing the model to realistic anatomical combinations. This allows it to resolve ambiguities based on user interactions and flexibly adapt to different segmentation needs. Additionally, we do not harmonize class definitions across datasets, preserving ambiguities arising from differing annotation conventions and label definitions.
\section{Training Data}

\subsection{Large-Scale, Multi-Domain Training Dataset}

We develop our model on an unprecedented collection of over 120 publicly available 3D segmentation datasets, comprising a total of 64,518 volumes with 717,148 objects (5\% of images held out for internal validation). This diverse collection spans a wide range of structures across multiple imaging modalities (Tab. \ref{tab:training_datasets}). Specifically, the collection spans Computed Tomography (CT)~\cite{antonelli2021medical, BTCV, lidc, BTCV2, structseg, lambert2019segthor, NHpancreas, verse1, verse2, verse3, ripseg, ji2022amos, abdomenatlas8k, totalseg, FLARE22, Jin2021-jo, luo2022word, Ma-2021-AbdomenCT-1K, qu2023abdomenatlas, topcowchallenge, Rister2019tm, deeplesion, covid19_junma, heller2023kits21, FLARE23, lndb, Antonelli2022, nih_lymph, NSCLC-Radiomics, autopet, covid19_challenge, kiser2020plethora, instance22, kanwar2023stress, raudaschl2017evaluation, toothfairy2, luo2023efficient, wolterink2016evaluation, he2021meta, RieraMarin2024, jordan2022pediatric, atm22, roth2014new, bouget2019semantic, bouget2021mediastinal, pieper2024spinemets, IMRAN2024102470, støverud2023aeropath}, different Magnetic Resonance Imaging (MRI) sequences~\cite{Antonelli2022, isles2015, Litjens2014, Bernard2018, Carass2017, CHAOSdata2019, Campello2021, prostatex, Grovik2020-un, Shapey2021-iz, Liew2018-wy, ji2022amos, bratsgli1, bratsgli2, bratsgli3, totalseg_mri, prostate158, remind, upenn-gbm, sudre2024valdo, spider, vallieres2015radiomics, ski10, kuijf2024mr, pace2024hvsmr, lalande2020emidec, Mayr2023, quinton2023tumour, mama_mia, marcus2007open, bloch2015nci, kuijf2019standardized, xiong2021global, kadish2009rationale, alexander2019desikan, buda2019association, payette2021automatic, macdonald_2020_7774566}, 3D Ultrasound~\cite{kronke2022tracked, behboodi2024open, duque2024ultrasound, bernard2015standardized, leclerc2019deep}, Positron Emission Tomography (PET)~\cite{autopet, autopet2024fdgpsma, Andrearczyk2023-ii} and 3D Microscopy~\cite{lapd_mouse, mavska2023cell, zheng2023nis3d, gerhard2013segmented, lucchi2011supervoxel, ljosa2012annotated, svoboda2009generation, svoboda2011generation, lin_yang_2023_7413818, Lin_2021, wei2021axonem}. The scale and diversity of this dataset collection is unmatched in 3D medical image segmentation, providing our model with exposure to a broad range of imaging modalities, (anatomical) structures, and pathological conditions. This enables robust, generalizable representations across varying image scales and clinical scenarios, laying the foundation for a versatile model capable of addressing diverse applications.

\subsection{Label Diversity with SuperVoxels}

\begin{figure}[t]
  \raggedright
  \scriptsize\textbf{\;\;\;\;\;\;\;\; SLIC~\cite{achanta2012slic} \;\;\;\;\;\;\;\;\;\;\;\; Vista3D~\cite{vista3d} \;\;\;\;\;\; Felzenszwalb~\cite{felzenszwalb2004efficient} \;\;\;\;\;\;\;\;\;\; \footnotesize\textit{Ours}}\par
  \centering
  \includegraphics[width=\linewidth]{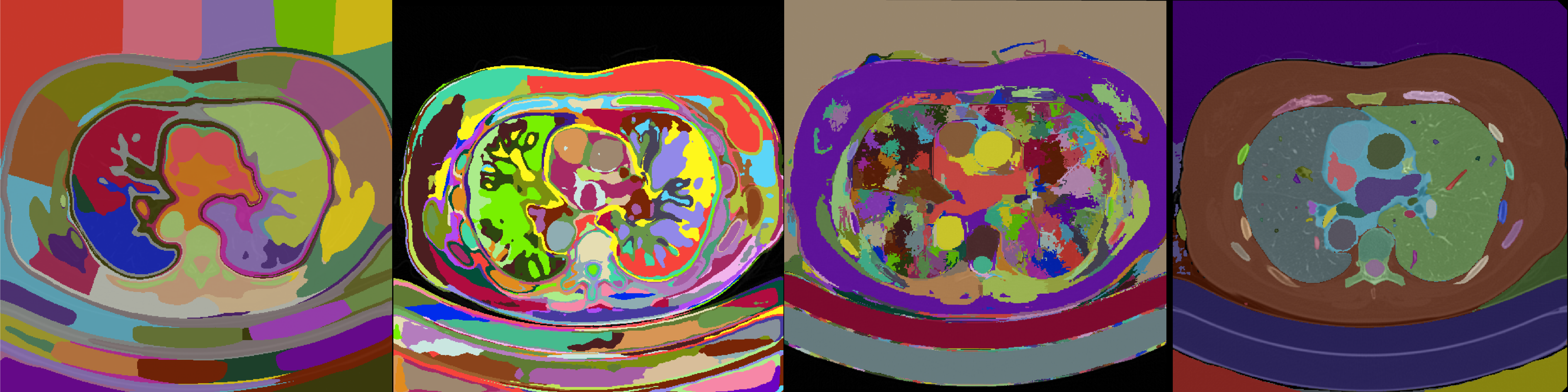}
  \caption{\textbf{SuperVoxels enhancing training label diversity.} Classical algorithms like SLIC and Felzenszwalb produce parcellation or fuzzy boundaries, even using image embeddings (Vista3D), while our approach yields precise variable-sized objects.}
  \vspace{-0.3cm}
  \label{fig:supervoxel}
\end{figure}

Despite the scale and variety of our training distribution, further increasing robustness to unseen structures remains essential. To achieve this we incorporate pseudo-labels sampled with a probability of 0.2. While existing frameworks have relied on traditional computer vision algorithms such as Felzenszwalb~\cite{felzenszwalb2004efficient} or SLIC~\cite{achanta2012slic}, either applied directly to images~\cite{segvol,scribbleprompt} or to encoded representations (Vista3D~\cite{vista3d}), we leverage the capabilities of modern foundation models. Specifically, we utilize SAM’s automatic "segment everything" feature to generate high-confidence SuperVoxels ($ \geq92\% $) on axially sampled slices. We then employ SAM2’s video mask propagation, treating the remaining slices as sequential frames to generate a 3D segmentation. As illustrated in Fig.~\ref{fig:supervoxel}, our approach mitigates common limitations of SLIC, which struggles with image parcellation even when applied to embeddings, and Felzenszwalb, which tends to produce fuzzy borders. In contrast, our method generates high-quality, variable-sized objects, enhancing segmentation accuracy and adaptability across diverse structures.

\section{Experiments}

To evaluate nnInteractive we perform a large-scale comparison against established 2D and 3D models (Sec \ref{experiments:radioactive}), benchmark on human expert scribbles (Sec \ref{experiments:humanScribbles}), compare the performance of supported prompting styles (Sec \ref{experiments:promptstyles}) and perform a User Study involving medical doctors (Sec \ref{experiments:userstudy}).

\subsection{Benchmarking against Established Methods.} 
\label{experiments:radioactive}
For our primary comparison, we utilize the RadioActive Benchmark~\cite{radioactive}, which provides a comprehensive and reproducible evaluation framework with advanced prompting techniques, ensuring a fair and thorough comparison against existing methods. Competing 2D models include SAM \cite{segment_anything} and SAM2 \cite{sam2}, originally trained on natural images, as well as medical adaptations: SAM-Med2D \cite{cheng2023sammed2d}, trained on diverse biomedical modalities; MedSAM \cite{medsam}, trained with box prompts on 1.5M medical segmentations; and ScribblePrompt \cite{scribbleprompt}, trained on 65 medical datasets spanning healthy and pathological structures. For 3D models, we compare against SAM-Med3D \cite{sammed3d}, a 3D extension of SAM, and SegVol \cite{segvol}, trained on multi-organ and lesion datasets. Other notable interactive models, such as Vista3D~\cite{vista3d}, 3D SAM Adapter~\cite{3dsamadapter}, and Prism~\cite{prism}, are designed for closed-set segmentation and were therefore not included in our evaluation. Our comparison is conducted on the ten held-out test datasets proposed by the benchmark~\cite{muslim2022ms,podobnik2023han,hntsmrg2024wahid,rider_lung,lnq2023challenge,livermets,adrenalacc,hcctace,liu2023pelvic,luo2023segrap2023}, covering standard CT and MRI modalities, including various anatomical structures and pathologies such as lesions. To further assess zero-shot adaptability and the generalization capabilities expected of foundation models, we extend the test set with out-of-distribution datasets featuring different imaging modalities and unseen tasks. Specifically, we evaluate models on microCT images of mouse tumors~\cite{mouseCTtumor}, knee MRI ligament structures~\cite{stanford_knee_mri} (unseen during training), microCT scans of insect anatomy~\cite{ant_brain}, and PET images of head \& neck tumors~\cite{Kinahan2019acrin} (Tab \ref{tab:datasets_test}). These datasets introduce significant domain shifts in resolution, contrast, target, and anatomical scale, providing a challenging benchmark for evaluating model robustness beyond conventional medical imaging tasks.\\

\noindent We start by comparing static point and 3D box prompts, as these are the most widely supported prompt types in current methods (e.g., MedSam~\cite{medsam} is limited to non-interactive boxes). In terms of points, 3D models receive a single point per structure of interest, while 2D models receive one point per slice, which increases theoretically invested user effort. For bounding box interactions, each model receives the precise 3D bounding box around the target structure, while 2D models are queried with the corresponding sliced bounding box. Since 3D bounding boxes are impractical for interactive segmentation compromising user-friendliness, we trained a dedicated nnInteractive variant solely for benchmarking against existing models that rely on them (see Appendix \ref{appendix:3dbboxes_suck}). Following the benchmark's guidelines, we also simulate static scribbles as out-of-plane lines/curves, querying the respective point in each slice for 2D models, while 3D models receive the full scribble.\\  
To benchmark interactive refinement capabilities we use point prompts as they are the only prompting style supported across all methods. We place an initial point, followed up by 5 additional clicks sampled according to the model’s generated segmentation mask. Negative points are enabled if supported by the model.

\subsection{Expert Scribble Benchmark.} 
\label{experiments:humanScribbles}
To assess performance on real rather than simulated scribbles, we evaluate on unseen data from the MS-CMRSeg 2019 Challenge~\cite{MSCMRSeg} with expert-provided per-slice axial scribble annotations for the left ventricle (LV), right ventricle (RV), and myocardium (MYO)~\cite{zhang2022cyclemix}. ScribblePrompt~\cite{scribbleprompt}, the only competing model with native scribble support, is prompted in a slice-wise 2D manner, using all scribbled slices. nnInteractive is evaluated using all annotated slices as well, but also tested with \textit{only three scribbles} (top, middle, bottom), leveraging its ability to handle sparse prompts, simulating a \textit{significantly reduced annotation effort}.

\subsection{Evaluation of Prompting Styles}
\label{experiments:promptstyles}
Current state-of-the-art methods are limited in their interactions, making it difficult to directly compare all interactions supported in nnInteractive for which there is no analogue in existing methods. This does not only affect the newly proposed lasso interaction but also the 2D bounding boxes and scribbles where nnInteractive is currently the only method able to accept these in a true 3D setting. We evaluate the usefulness of these advanced interactions on the aforementioned test and OOD datasets. We furthermore test random combination of interactions given to nnInteractive to confirm that it can make effective use of arbitrary combinations of inputs. Point, 2D bbox, lasso and scribble interactions are simulated using the logic outlined in Sec \ref{methods:interactions}.

\subsection{Radiological User Study.}
\label{experiments:userstudy}
To evaluate real-world usability, we conducted a user study on the segmentation of 12 tumor lesions across various anatomical regions, derived from MR and CT scans used in radiation therapy planning. After briefly introducing the radiologist to the prompt types, each lesion was annotated using nnInteractive and compared against expert manual segmentations from two raters —a resident and a specialist radiation oncologist —using their inter-rater variability as a reference. Additionally, we measured the time required for both manual and prompt-guided segmentation to assess efficiency improvements.

\section{Results \& Discussion}

We first present results from the expanded RadioActive~\cite{radioactive} benchmark and expert scribble comparison, followed by an analysis of nnInteractive's unique contributions and our user study.

\begin{figure}[t]
    \centering
    \begin{subfigure}{0.36\linewidth}
        \centering
        \includegraphics[width=\textwidth]{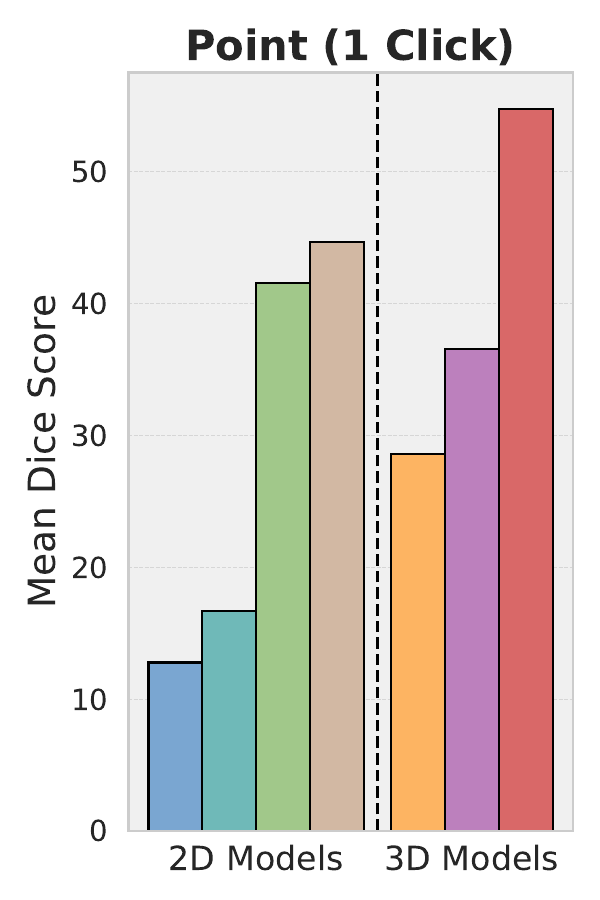}
    \end{subfigure}
    \begin{subfigure}{0.36\linewidth}
        \centering
        \includegraphics[width=\textwidth]{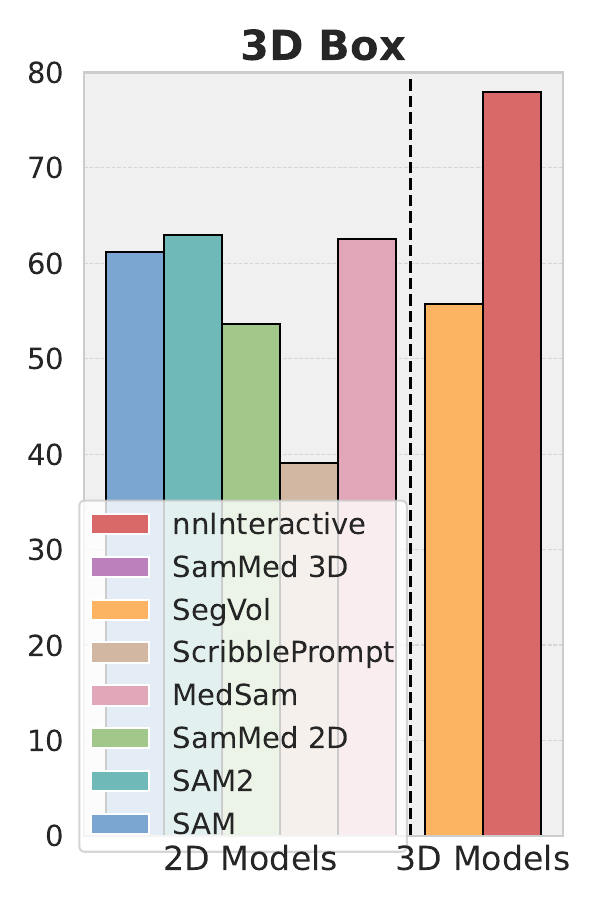}
    \end{subfigure}
    \begin{subfigure}{0.26\linewidth}
        \centering
        \includegraphics[width=\textwidth]{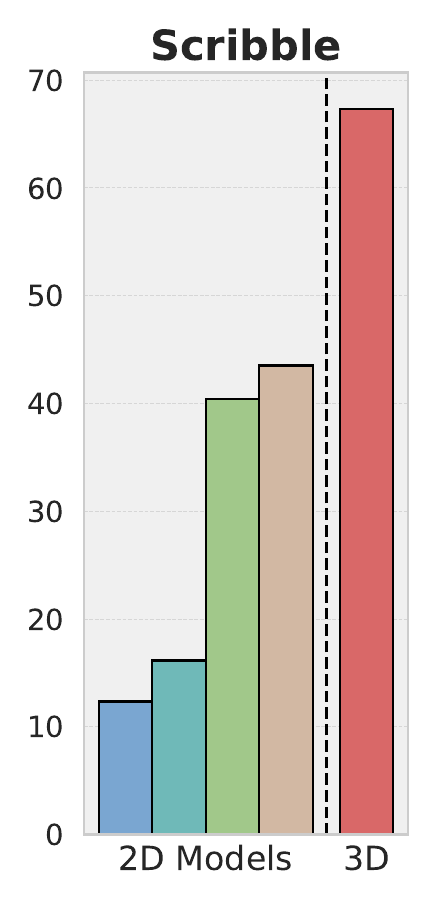}
    \end{subfigure}
    \caption{\textbf{Single-prompt performance.} Mean Dice scores on unseen test data when prompting with (1) a single point (2D models receive one per slice, giving them a theoretical advantage over 3D models), (2) a 3D bounding box (sliced for 2D models), and (3) an out-of-plane scribble (interpreted as points per slice in 2D). nnInteractive consistently outperforms all baselines, demonstrating superior segmentation quality across all interactions.}
    \label{fig:static_results}
\end{figure}

\subsection{Comparison with SOTA}

\begin{figure*}[t]
    \begin{center}
        \includegraphics[width=\linewidth]{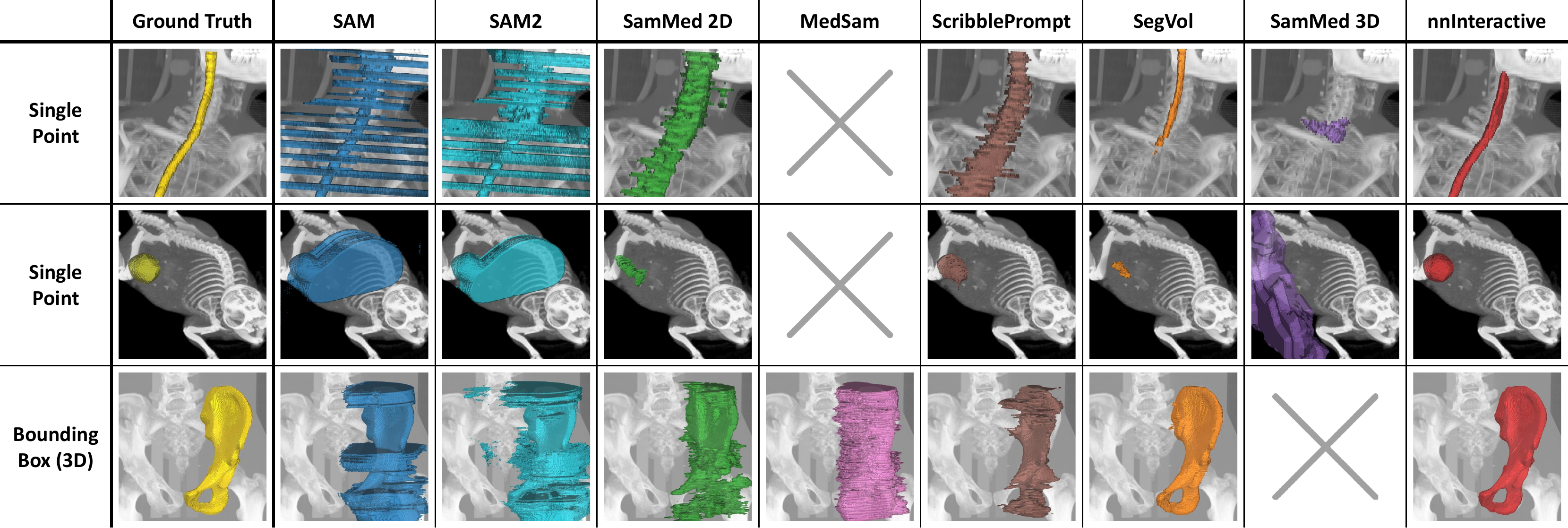}
    \end{center}
    \vspace{-0.5cm}
    \caption{Qualitative comparison of interactive segmentation methods on unseen test images with different static prompting strategies: points and 3D boxes. nnInteractive achieves the highest accuracy, closely matching the ground truth, while others struggle with precision, consistency, or volumetric adaptation. Omitted results ($\times$) indicate unsupported prompt types.}
    \vspace{-0.3cm}
    \label{fig:qualitative_main}
\end{figure*}

\textbf{Static prompts.} Figure \ref{fig:static_results} compares model performance using a single point, box, or out-of-plane scribble. nnInteractive consistently outperforms all baselines across interaction types. For point prompts, it surpasses the closest competitor, ScribblePrompt, by 10.1 Dice points, despite 2D models receiving significantly more prompts. For boxes, it achieves a 14.9 Dice point advantage over the second-best model SAM2, and for scribbles, an impressive 23.8-point lead. Notably, no single competing model consistently ranks second, as their performance highly varies by interaction type.

\begin{figure}[t]
  \centering
  \begin{subfigure}{0.495\linewidth}
        \centering
        \includegraphics[width=\textwidth]{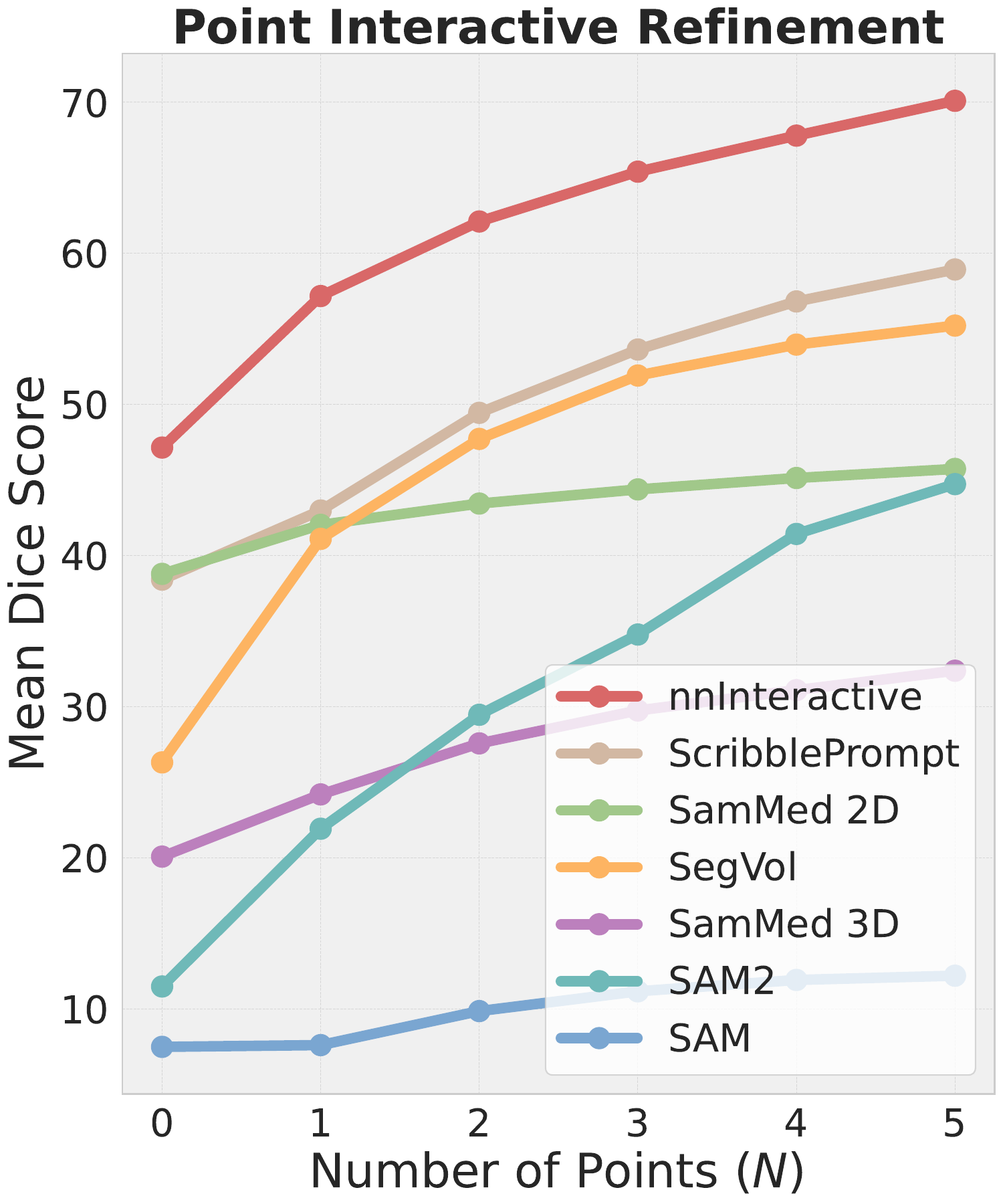}
    \end{subfigure}
    \begin{subfigure}{0.495\linewidth}
        \centering
        \includegraphics[width=\textwidth]{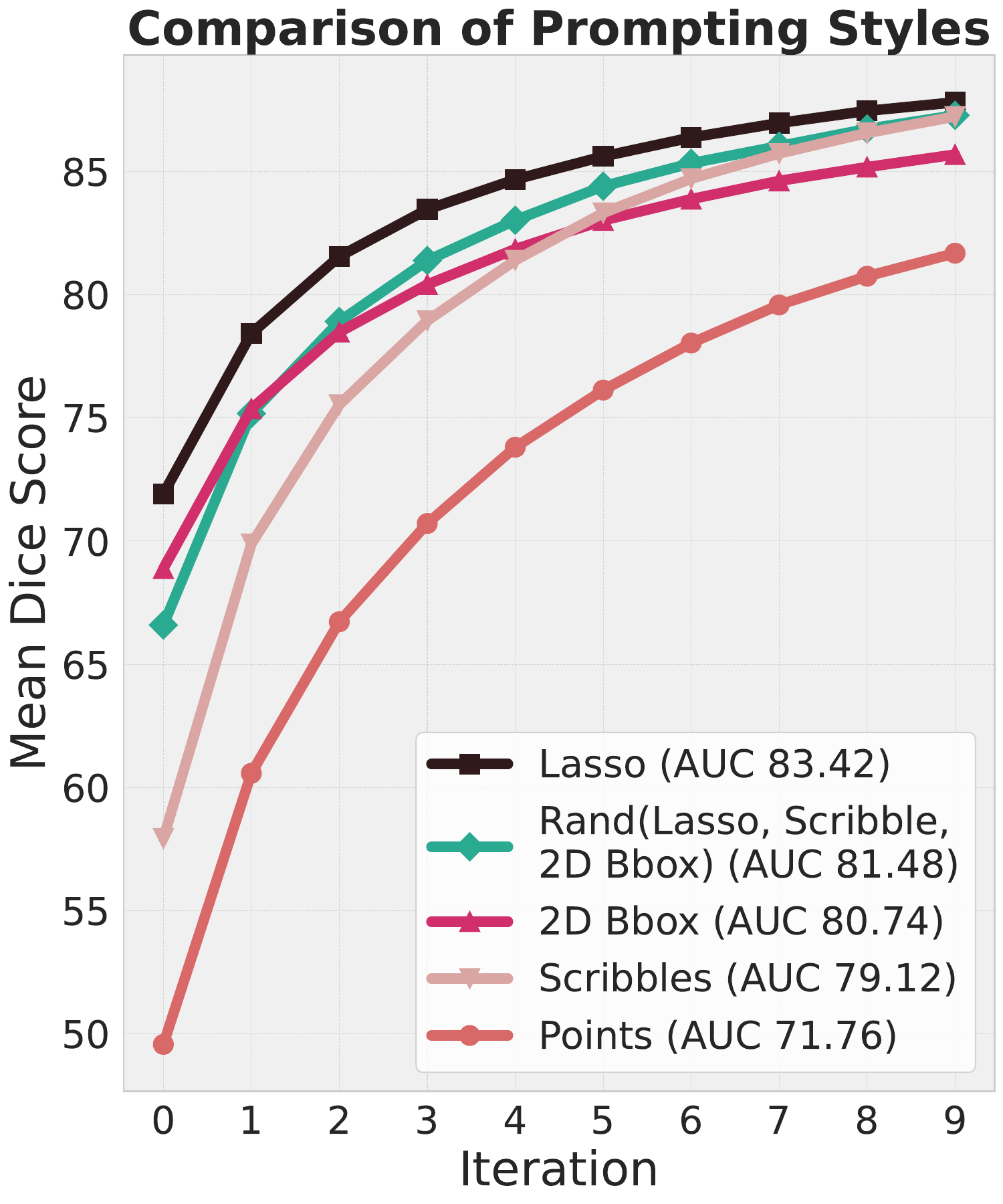}
    \end{subfigure}
  \caption{\textbf{Interactive Performance.} nnInteractive achieves the highest Dice scores in \underline{point-based refinement} (\textit{Left}), with a large gap over all competitors, despite 2D models receiving N points per slice and points being nnInteractive's weakest prompt type (\textit{Right)}. Beyond points, nnInteractive excels with \underline{stronger interactions like scribbles and lasso}. Interactions can be freely mixed for flexibility.}
  \vspace{-0.3cm}
  \label{fig:points_interactive}
\end{figure}

\noindent\textbf{Interactive Prompts.} Fig. \ref{fig:points_interactive} (\textit{Left}) evaluates the interactive refinement capabilities of models using positive and negative point prompts, if supported. nnInteractive consistently achieves the highest performance, starting with a superior initial Dice score and reaching a Dice of more than 70 across all datasets. This surpasses the best-performing 2D model, ScribblePrompt, by 11.2 Dice points despite its significantly higher number of prompts, and outperforms the strongest 3D competitor, SegVol, by 14.9 Dice points.

\noindent\textbf{Expert Scribble Benchmark.} Table \ref{tab:expert_scribble} compares nnInteractive and ScribblePrompt on expert-drawn scribbles across three target classes. When provided with the same number of prompts (scribbles on every slice), nnInteractive outperforms ScribblePrompt by nearly 7 Dice points on average. Notably, even with only three annotated slices, nnInteractive achieves a Dice score of 84.3—surpassing ScribblePrompt by 3 points while requiring significantly less user interaction. This demonstrates nnInteractive’s ability to \textit{achieve high segmentation accuracy with minimal annotation effort}.

\begin{table}[t]
\centering
\begin{adjustbox}{width=\linewidth}
\begin{tabular}{lccccc}
\toprule
Prompt & Model & LV & RV & MYO & Agv.\\
\midrule
All Slices & ScribblePrompt & 66.08 & 90.04 & 86.82 & 80.98 \\
All Slices       & nnInteractive & \textbf{78.86} & \textbf{92.93} & \textbf{90.07} & \textbf{87.29} \\
\midrule
3 Slices & nnInteractive & \underline{74.40} & \underline{91.24} & \underline{87.33} & \underline{84.29} \\
\bottomrule
\end{tabular}
\end{adjustbox}
\caption{\textbf{Expert Scribbles Benchmark.} Performance (Dice) using human axial scribbles as prompts for the left \& right ventricle (LV, RV) and myocardium (MYO). nnInteractive outperforms ScribblePrompt across all structures, even when limited to just three annotated slices instead of all scribbles.}
\vspace{-0.5cm}
\label{tab:expert_scribble}
\end{table}

\subsection{Evaluation nnInteractive's contributions}

\paragraph{Prompting Styles.} We evaluate nnInteractive's prompting styles on the Test and OOD datasets using simulated interactions as described in \ref{methods:interactions}. Due to the absence of competing methods capable of processing 2D bounding boxes, scribbles, and lasso in a 3D setting, we ground results using the only common interaction type: points. As shown in \cref{fig:points_interactive} (\textit{Right}), among all interaction types, lasso performs best, achieving the highest Dice scores across all iterations with an AUC of 83.42. 2D bounding boxes yield high initial Dice scores but fall behind with more iterations, likely due to their coarse guidance being less effective for refinement. Scribbles initially provide less information than lasso or bounding boxes but surpass bounding boxes after five iterations, demonstrating their strength in precise refinement. Points perform the weakest, with an AUC of 71.76, consistently lagging behind other interaction types. This finding is notable as nnInteractive substantially outperforms current state-of-the-art models in points (both in static and interactive setting), despite points being its worst performing prompting style. Finally, we tested randomly selecting interactions each iteration, showing that interactions can be freely mixed in nnInteractive, with Rand(Lasso, Scribble, 2D Bbox) achieving Dice scores between the respective non-random interaction simulations.

\paragraph{AutoZoom.} As shown in \cref{fig:appendix_autozoom}, AutoZoom with refinement effectively captures large objects and accelerates convergence with fewer user interactions. While AUC improvement on the Test and OOD datasets is minimal (82.72 vs 82.36) due to the predominance of small objects, substantial gains are observed for datasets with larger objects, such as HCC Tace liver \cite{hcctace} (AUC 95.40 vs 91.92) and InsectAnatomy \cite{ant_brain} (AUC 94.81 vs 92.54). Further information in \cref{appendix:autozoom}.


\paragraph{Resolving Ambiguity.} Existing methods struggle with ambiguities due to overfitting on learned classes, for example failing to segment structures like kidneys and kidney tumors simultaneously (see \cref{fig:vista_fail_kidney_and_tumor}). nnInteractive overcomes this limitation and can dynamically adapt to user input, efficiently resolving ambiguities with minimal iterations. We demonstrate this on two organs with and without tumors in previously unseen images (Fig. \ref{fig:nninteractive_ambiguities_win}).

\paragraph{Inference Time.} Optimized for broad adoption, our implementation maintains VRAM usage below 10 GB (\(<6\) GB for small objects). Small structures like tumors and organs take merely 120–200 ms on an NVIDIA RTX 4090. For larger objects, AutoZoom iteratively zooms out and then refines segmentations, increasing inference time up to 1160 ms for a liver CT and 3700 ms in rare high-resolution cases. See Appendix \ref{appendix:runtime}.

\subsection{Real-World Impact on Radiological Tasks}

To assess the clinical applicability of nnInteractive, we compared its segmentation accuracy and efficiency against expert manual tumor annotation. Using Dice similarity coefficients, we found that segmentations generated using nnInteractive were as consistent with specialist annotations as inter-expert agreement itself. Median Dice scores were 0.842\footnotesize±0.058\normalsize\;between resident and specialist annotations, 0.794\footnotesize±0.040\normalsize\; between specialist annotations and nnInteractive, and 0.853\footnotesize±0.068\normalsize\; between the resident and nnInteractive. Wilcoxon tests confirmed no significant differences between these comparisons (\( p=0.577 \) and \( p=0.365 \)), indicating that \textit{nnInteractive achieves expert-level performance}. Beyond accuracy, it dramatically improves efficiency: Experts completed per case segmentations in 179\footnotesize±114\normalsize\;seconds using nnInteractive—72\% faster than the 635\footnotesize±343\normalsize\;seconds required for manual annotation. This substantial time reduction demonstrates nnInteractive’s potential to streamline clinical workflows, reducing workload while maintaining expert-grade precision.


\section{Conclusion}

We introduced nnInteractive, a universal 3D promptable segmentation framework that sets a new standard for AI-driven interactive segmentation. By supporting diverse prompt types—including points, scribbles, bounding boxes, and lasso—nnInteractive bridges the gap between intuitive 2D interactions and full 3D volumetric segmentation. Trained on an unprecedented dataset of 120+ multimodal 3D datasets, our model achieves superior performance and usability across a wide range of imaging tasks. Extensive benchmarking demonstrates that our approach far surpasses existing methods, offering state-of-the-art segmentation accuracy while significantly reducing annotation effort. Integrated into established imaging platforms such as Napari and MITK Workbench, it ensures seamless real-world adoption in clinical and research workflows, paving the way for more efficient and accessible 3D segmentation.

{
    \small
    \bibliographystyle{ieeenat_fullname}
    \bibliography{main}

\begin{thebibliography}{154}
\providecommand{\natexlab}[1]{#1}
\providecommand{\url}[1]{\texttt{#1}}
\expandafter\ifx\csname urlstyle\endcsname\relax
  \providecommand{\doi}[1]{doi: #1}\else
  \providecommand{\doi}{doi: \begingroup \urlstyle{rm}\Url}\fi

\bibitem[Achanta et~al.(2012)Achanta, Shaji, Smith, Lucchi, Fua, and S{\"u}sstrunk]{achanta2012slic}
Radhakrishna Achanta, Appu Shaji, Kevin Smith, Aurelien Lucchi, Pascal Fua, and Sabine S{\"u}sstrunk.
\newblock Slic superpixels compared to state-of-the-art superpixel methods.
\newblock \emph{IEEE transactions on pattern analysis and machine intelligence}, 34\penalty0 (11):\penalty0 2274--2282, 2012.

\bibitem[Aerts et~al.(2019)Aerts, Wee, Rios~Velazquez, Leijenaar, Parmar, Grossmann, Carvalho, Bussink, Monshouwer, Haibe-Kains, Rietveld, Hoebers, Rietbergen, Leemans, Dekker, Quackenbush, Gillies, and Lambin]{NSCLC-Radiomics}
Hugo J. W.~L. Aerts, Leonard Wee, Emmanuel Rios~Velazquez, Ralph T.~H. Leijenaar, Chintan Parmar, Patrick Grossmann, Sara Carvalho, Johan Bussink, Ren Monshouwer, Benjamin Haibe-Kains, Derek Rietveld, Frank Hoebers, Michelle~M. Rietbergen, C.~Ren Leemans, Andre Dekker, John Quackenbush, Robert~J. Gillies, and Philippe Lambin.
\newblock Data from nsclc-radiomics, 2019.

\bibitem[Alexander et~al.(2019)Alexander, Loh, Matthews, Murray, Adamson, Beare, Chen, Kelly, Anderson, Doyle, et~al.]{alexander2019desikan}
Bonnie Alexander, Wai~Yen Loh, Lillian~G Matthews, Andrea~L Murray, Chris Adamson, Richard Beare, Jian Chen, Claire~E Kelly, Peter~J Anderson, Lex~W Doyle, et~al.
\newblock Desikan-killiany-tourville atlas compatible version of m-crib neonatal parcellated whole brain atlas: The m-crib 2.0.
\newblock \emph{Frontiers in Neuroscience}, 13:\penalty0 34, 2019.

\bibitem[Andrearczyk et~al.(2023)Andrearczyk, Oreiller, Abobakr, Akhavanallaf, Balermpas, Boughdad, Capriotti, Castelli, Le~Rest, and Decazes]{Andrearczyk2023-ii}
Vincent Andrearczyk, Valentin Oreiller, Moamen Abobakr, Azadeh Akhavanallaf, Panagiotis Balermpas, Sarah Boughdad, Leo Capriotti, Joel Castelli, Catherine~Cheze Le~Rest, and Pierre et~al. Decazes.
\newblock Overview of the {HECKTOR} challenge at {MICCAI} 2022: Automatic head and neck {TumOR} segmentation and outcome prediction in {PET/CT}.
\newblock \emph{Head Neck Tumor Chall (2022)}, 2023.

\bibitem[Antonelli et~al.(2021)Antonelli, Reinke, Bakas, Farahani, Kopp-Schneider, Landman, Litjens, Menze, Ronneberger, Summers, and et~al.]{antonelli2021medical}
Michela Antonelli, Annika Reinke, Spyridon Bakas, Keyvan Farahani, Annette Kopp-Schneider, Bennett~A. Landman, Geert Litjens, Bjoern Menze, Olaf Ronneberger, Ronald~M. Summers, and et al.
\newblock The medical segmentation decathlon.
\newblock \emph{arXiv:2106.05735}, 2021.

\bibitem[Antonelli et~al.(2022)Antonelli, Reinke, Bakas, Farahani, Kopp-Schneider, Landman, Litjens, Menze, Ronneberger, Summers, and et~al]{Antonelli2022}
Michela Antonelli, Annika Reinke, Spyridon Bakas, Keyvan Farahani, Annette Kopp-Schneider, Bennett~A. Landman, Geert Litjens, Bjoern Menze, Olaf Ronneberger, Ronald~M. Summers, and et al.
\newblock The medical segmentation decathlon.
\newblock \emph{Nature Communications}, 2022.

\bibitem[{Armato III, Samuel G.} et~al.(2015){Armato III, Samuel G.}, McLennan, Bidaut, McNitt-Gray, Meyer, Reeves, Zhao, Aberle, Henschke, and Hoffman]{lidc}
{Armato III, Samuel G.}, Geoffrey McLennan, Luc Bidaut, Michael~F. McNitt-Gray, Charles~R. Meyer, Anthony~P. Reeves, Binsheng Zhao, Denise~R. Aberle, Claudia~I. Henschke, and Eric A. et~al. Hoffman.
\newblock Data from lidc-idri, 2015.

\bibitem[Baid et~al.(2021)Baid, Ghodasara, Mohan, Bilello, Calabrese, Colak, Farahani, Kalpathy-Cramer, Kitamura, Pati, and et~al.]{bratsgli3}
Ujjwal Baid, Satyam Ghodasara, Suyash Mohan, Michel Bilello, Evan Calabrese, Errol Colak, Keyvan Farahani, Jayashree Kalpathy-Cramer, Felipe~C Kitamura, Sarthak Pati, and Spyridon et al.
\newblock The {RSNA-ASNR-MICCAI} {BraTS} 2021 benchmark on brain tumor segmentation and radiogenomic classification.
\newblock \emph{arXiv}, 2021.

\bibitem[Bakas et~al.(2017)Bakas, Akbari, Sotiras, Bilello, Rozycki, Kirby, Freymann, Farahani, and Davatzikos]{bratsgli1}
Spyridon Bakas, Hamed Akbari, Aristeidis Sotiras, Michel Bilello, Martin Rozycki, Justin~S Kirby, John~B Freymann, Keyvan Farahani, and Christos Davatzikos.
\newblock Advancing the cancer genome atlas glioma {MRI} collections with expert segmentation labels and radiomic features.
\newblock \emph{Sci. Data}, 2017.

\bibitem[Bakas et~al.(2022)Bakas, Sako, Akbari, Bilello, Sotiras, Shukla, Rudie, Santamar{\'\i}a, Kazerooni, Pati, Rathore, Mamourian, Ha, Parker, Doshi, Baid, Bergman, Binder, Verma, Lustig, Desai, Bagley, Mourelatos, Morrissette, Watt, Brem, Wolf, Melhem, Nasrallah, Mohan, O'Rourke, and Davatzikos]{upenn-gbm}
Spyridon Bakas, Chiharu Sako, Hamed Akbari, Michel Bilello, Aristeidis Sotiras, Gaurav Shukla, Jeffrey~D. Rudie, Natali~Flores Santamar{\'\i}a, Anahita~Fathi Kazerooni, Sarthak Pati, Saima Rathore, Elizabeth Mamourian, Sung~Min Ha, William Parker, Jimit Doshi, Ujjwal Baid, Mark Bergman, Zev~A. Binder, Ragini Verma, Robert~A. Lustig, Arati~S. Desai, Stephen~J. Bagley, Zissimos Mourelatos, Jennifer Morrissette, Christopher~D. Watt, Steven Brem, Ronald~L. Wolf, Elias~R. Melhem, MacLean~P. Nasrallah, Suyash Mohan, Donald~M. O'Rourke, and Christos Davatzikos.
\newblock The university of pennsylvania glioblastoma (upenn-gbm) cohort: advanced mri, clinical, genomics, \& radiomics.
\newblock \emph{Scientific Data}, 9\penalty0 (1):\penalty0 453, 2022.

\bibitem[Bassi et~al.(2024)Bassi, Li, Tang, Isensee, Wang, Chen, Chou, Kirchhoff, Rokuss, Huang, Ye, He, Wald, Ulrich, Baumgartner, Roy, Maier-Hein, Jaeger, Ye, Xie, Zhang, Chen, Xia, Xing, Zhu, Sadegheih, Bozorgpour, Kumari, Azad, Merhof, Shi, Ma, Du, Bai, Huang, Zhao, Wang, Li, Gu, Dong, Yang, Mazurowski, Gupta, Wu, Zhuang, Chen, Roth, Xu, Blaschko, Decherchi, Cavalli, Yuille, and Zhou]{bassi2024touchstonebenchmark}
Pedro R. A.~S. Bassi, Wenxuan Li, Yucheng Tang, Fabian Isensee, Zifu Wang, Jieneng Chen, Yu-Cheng Chou, Yannick Kirchhoff, Maximilian Rokuss, Ziyan Huang, Jin Ye, Junjun He, Tassilo Wald, Constantin Ulrich, Michael Baumgartner, Saikat Roy, Klaus~H. Maier-Hein, Paul Jaeger, Yiwen Ye, Yutong Xie, Jianpeng Zhang, Ziyang Chen, Yong Xia, Zhaohu Xing, Lei Zhu, Yousef Sadegheih, Afshin Bozorgpour, Pratibha Kumari, Reza Azad, Dorit Merhof, Pengcheng Shi, Ting Ma, Yuxin Du, Fan Bai, Tiejun Huang, Bo Zhao, Haonan Wang, Xiaomeng Li, Hanxue Gu, Haoyu Dong, Jichen Yang, Maciej~A. Mazurowski, Saumya Gupta, Linshan Wu, Jiaxin Zhuang, Hao Chen, Holger Roth, Daguang Xu, Matthew~B. Blaschko, Sergio Decherchi, Andrea Cavalli, Alan~L. Yuille, and Zongwei Zhou.
\newblock Touchstone benchmark: Are we on the right way for evaluating ai algorithms for medical segmentation?, 2024.

\bibitem[Behboodi et~al.(2024)Behboodi, Carton, Chabanas, de~Ribaupierre, Solheim, Munkvold, Rivaz, Xiao, and Reinertsen]{behboodi2024open}
Bahareh Behboodi, Francois-xavier Carton, Matthieu Chabanas, Sandrine de Ribaupierre, Ole Solheim, Bodil~KR Munkvold, Hassan Rivaz, Yiming Xiao, and Ingerid Reinertsen.
\newblock Open access segmentations of intraoperative brain tumor ultrasound images.
\newblock \emph{Medical Physics}, 51\penalty0 (9):\penalty0 6525--6532, 2024.

\bibitem[Beichel et~al.(2019)Beichel, Glenny, Bauer, and Krueger]{lapd_mouse}
Reinhard~R Beichel, Robb~W Glenny, Christian Bauer, and Melissa~A Krueger.
\newblock Lung anatomy + particle deposition (lapd) mouse archive, 2019.

\bibitem[Bernard et~al.(2015)Bernard, Bosch, Heyde, Alessandrini, Barbosa, Camarasu-Pop, Cervenansky, Valette, Mirea, Bernier, et~al.]{bernard2015standardized}
Olivier Bernard, Johan~G Bosch, Brecht Heyde, Martino Alessandrini, Daniel Barbosa, Sorina Camarasu-Pop, Frederic Cervenansky, S{\'e}bastien Valette, Oana Mirea, Michel Bernier, et~al.
\newblock Standardized evaluation system for left ventricular segmentation algorithms in 3d echocardiography.
\newblock \emph{IEEE transactions on medical imaging}, 35\penalty0 (4):\penalty0 967--977, 2015.

\bibitem[Bernard et~al.(2018)Bernard, Lalande, Zotti, Cervenansky, Yang, Heng, Cetin, Lekadir, Camara, and Gonzalez~Ballester]{Bernard2018}
Olivier Bernard, Alain Lalande, Clement Zotti, Frederick Cervenansky, Xin Yang, Pheng-Ann Heng, Irem Cetin, Karim Lekadir, Oscar Camara, and et~al. Gonzalez~Ballester.
\newblock Deep learning techniques for automatic mri cardiac multi-structures segmentation and diagnosis: Is the problem solved?
\newblock \emph{IEEE Transactions on Medical Imaging}, 2018.

\bibitem[Bloch et~al.(2015)Bloch, Madabhushi, Huisman, Freymann, Kirby, Grauer, Enquobahrie, Jaffe, Clarke, and Farahani]{bloch2015nci}
N. Bloch, A. Madabhushi, H. Huisman, J. Freymann, J. Kirby, M. Grauer, A. Enquobahrie, C. Jaffe, L. Clarke, and K. Farahani.
\newblock Nci-isbi 2013 challenge: Automated segmentation of prostate structures, 2015.

\bibitem[Bolelli et~al.(2024)Bolelli, Lumetti, Vinayahalingam, Bartolomeo, Pellacani, Marchesini, Van~Nistelrooij, Van~Lierop, Xi, Liu, Xin, Yang, Wang, Wang, Xu, Cui, Wodzinski, Müller, Kirchhoff, Rokuss, Maier-Hein, Han, Kim, Ahn, Szczepański, Grzeszczyk, Korzeniowski, Ballester, Burgos-Artizzu, Carrasco, Berge, Van~Ginneken, Anesi, and Grana]{toothfairy2}
Federico Bolelli, Luca Lumetti, Shankeeth Vinayahalingam, Mattia~Di Bartolomeo, Arrigo Pellacani, Kevin Marchesini, Niels Van~Nistelrooij, Pieter Van~Lierop, Tong Xi, Yusheng Liu, Rui Xin, Tao Yang, Lisheng Wang, Haoshen Wang, Chenfan Xu, Zhiming Cui, Marek Wodzinski, Henning Müller, Yannick Kirchhoff, Maximilian~R. Rokuss, Klaus Maier-Hein, Jaehwan Han, Wan Kim, Hong-Gi Ahn, Tomasz Szczepański, Michal~K. Grzeszczyk, Przemyslaw Korzeniowski, Vicent~Caselles Ballester, Xavier~Paolo Burgos-Artizzu, Ferran~Prados Carrasco, Stefaan Berge, Bram Van~Ginneken, Alexandre Anesi, and Costantino Grana.
\newblock Segmenting the inferior alveolar canal in cbcts volumes: the toothfairy challenge.
\newblock \emph{IEEE Transactions on Medical Imaging}, pages 1--1, 2024.

\bibitem[Bouget et~al.(2019)Bouget, J{\o}rgensen, Kiss, Leira, and Lang{\o}]{bouget2019semantic}
David Bouget, Arve J{\o}rgensen, Gabriel Kiss, Haakon~Olav Leira, and Thomas Lang{\o}.
\newblock Semantic segmentation and detection of mediastinal lymph nodes and anatomical structures in ct data for lung cancer staging.
\newblock \emph{International journal of computer assisted radiology and surgery}, 14:\penalty0 977--986, 2019.

\bibitem[Bouget et~al.(2022)Bouget, Pedersen, Vanel, Leira, and Langø]{bouget2021mediastinal}
David Bouget, André Pedersen, Johanna Vanel, Haakon~O. Leira, and Thomas Langø.
\newblock Mediastinal lymph nodes segmentation using 3d convolutional neural network ensembles and anatomical priors guiding.
\newblock \emph{Computer Methods in Biomechanics and Biomedical Engineering: Imaging \& Visualization}, 0\penalty0 (0):\penalty0 1--15, 2022.

\bibitem[Boyd(2014)]{dino_head}
Clint~A Boyd.
\newblock The cranial anatomy of the neornithischian dinosaur thescelosaurus neglectus.
\newblock \emph{PeerJ}, 2:\penalty0 e669, 2014.

\bibitem[Buda et~al.(2019)Buda, Saha, and Mazurowski]{buda2019association}
Mateusz Buda, Ashirbani Saha, and Maciej~A Mazurowski.
\newblock Association of genomic subtypes of lower-grade gliomas with shape features automatically extracted by a deep learning algorithm.
\newblock \emph{Computers in biology and medicine}, 109:\penalty0 218--225, 2019.

\bibitem[Campello et~al.(2021)Campello, Gkontra, Izquierdo, Martin-Isla, Sojoudi, Full, Maier-Hein, Zhang, He, and Ma]{Campello2021}
Victor~M. Campello, Polyxeni Gkontra, Cristian Izquierdo, Carlos Martin-Isla, Alireza Sojoudi, Peter~M. Full, Klaus Maier-Hein, Yao Zhang, Zhiqiang He, and et~.al Ma.
\newblock Multi-centre, multi-vendor and multi-disease cardiac segmentation: The mms challenge.
\newblock \emph{IEEE Transactions on Medical Imaging}, 2021.

\bibitem[Carass et~al.(2017)Carass, Roy, Jog, Cuzzocreo, Magrath, Gherman, Button, Nguyen, Prados, and Sudre]{Carass2017}
Aaron Carass, Snehashis Roy, Amod Jog, Jennifer~L. Cuzzocreo, Elizabeth Magrath, Adrian Gherman, Julia Button, James Nguyen, Ferran Prados, and et~al. Sudre.
\newblock Longitudinal multiple sclerosis lesion segmentation: Resource and challenge.
\newblock \emph{NeuroImage}, 2017.

\bibitem[Cheng et~al.(2023)Cheng, Ye, Deng, Chen, Li, Wang, Su, Huang, Chen, Jiang, Sun, He, Zhang, Zhu, and Qiao]{cheng2023sammed2d}
Junlong Cheng, Jin Ye, Zhongying Deng, Jianpin Chen, Tianbin Li, Haoyu Wang, Yanzhou Su, Ziyan Huang, Jilong Chen, Lei Jiang, Hui Sun, Junjun He, Shaoting Zhang, Min Zhu, and Yu Qiao.
\newblock Sam-med2d, 2023.

\bibitem[Clark et~al.(2013)Clark, Vendt, Smith, Freymann, Kirby, Koppel, Moore, Phillips, Maffitt, Pringle, Tarbox, and Prior]{clark_cancer_2013}
Kenneth Clark, Bruce Vendt, Kirk Smith, John Freymann, Justin Kirby, Paul Koppel, Stephen Moore, Stanley Phillips, David Maffitt, Michael Pringle, Lawrence Tarbox, and Fred Prior.
\newblock The {Cancer} {Imaging} {Archive} ({TCIA}): {Maintaining} and {Operating} a {Public} {Information} {Repository}.
\newblock \emph{Journal of Digital Imaging}, 2013.

\bibitem[D'Antonoli et~al.(2024)D'Antonoli, Berger, Indrakanti, Vishwanathan, Weiß, Jung, Berkarda, Rau, Reisert, Küstner, Walter, Merkle, Segeroth, Cyriac, Yang, and Wasserthal]{totalseg_mri}
Tugba~Akinci D'Antonoli, Lucas~K. Berger, Ashraya~K. Indrakanti, Nathan Vishwanathan, Jakob Weiß, Matthias Jung, Zeynep Berkarda, Alexander Rau, Marco Reisert, Thomas Küstner, Alexandra Walter, Elmar~M. Merkle, Martin Segeroth, Joshy Cyriac, Shan Yang, and Jakob Wasserthal.
\newblock Totalsegmentator mri: Sequence-independent segmentation of 59 anatomical structures in mr images, 2024.

\bibitem[de~Grauw et~al.(2024)de~Grauw, Scholten, Smit, Rutten, Prokop, van Ginneken, and Hering]{uls_challenge}
MJJ de Grauw, E~Th Scholten, EJ Smit, MJCM Rutten, M Prokop, B van Ginneken, and A Hering.
\newblock The uls23 challenge: a baseline model and benchmark dataset for 3d universal lesion segmentation in computed tomography.
\newblock \emph{arXiv preprint arXiv:2406.05231}, 2024.

\bibitem[Desai et~al.(2022)Desai, Schmidt, Rubin, Sandino, Black, Mazzoli, Stevens, Boutin, Ré, Gold, Hargreaves, and Chaudhari]{stanford_knee_mri}
Arjun~D Desai, Andrew~M Schmidt, Elka~B Rubin, Christopher~M Sandino, Marianne~S Black, Valentina Mazzoli, Kathryn~J Stevens, Robert Boutin, Christopher Ré, Garry~E Gold, Brian~A Hargreaves, and Akshay~S Chaudhari.
\newblock Skm-tea: A dataset for accelerated mri reconstruction with dense image labels for quantitative clinical evaluation, 2022.

\bibitem[Dorent et~al.(2024)Dorent, Khajavi, Idris, Ziegler, Somarouthu, Jacene, LaCasce, Deissler, Ehrhardt, Engelson, Fischer, Gu, Handels, Kasai, Kondo, Maier-Hein, Schnabel, Wang, Wang, Wald, Yang, Zhang, Zhang, Pieper, Harris, Kikinis, and Kapur]{lnq2023challenge}
Reuben Dorent, Roya Khajavi, Tagwa Idris, Erik Ziegler, Bhanusupriya Somarouthu, Heather Jacene, Ann LaCasce, Jonathan Deissler, Jan Ehrhardt, Sofija Engelson, Stefan~M. Fischer, Yun Gu, Heinz Handels, Satoshi Kasai, Satoshi Kondo, Klaus Maier-Hein, Julia~A. Schnabel, Guotai Wang, Litingyu Wang, Tassilo Wald, Guang-Zhong Yang, Hanxiao Zhang, Minghui Zhang, Steve Pieper, Gordon Harris, Ron Kikinis, and Tina Kapur.
\newblock Lnq 2023 challenge: Benchmark of weakly-supervised techniques for mediastinal lymph node quantification, 2024.

\bibitem[Du et~al.(2023)Du, Bai, Huang, and Zhao]{segvol}
Yuxin Du, Fan Bai, Tiejun Huang, and Bo Zhao.
\newblock Segvol: Universal and interactive volumetric medical image segmentation.
\newblock \emph{arXiv preprint arXiv:2311.13385}, 2023.

\bibitem[Duque et~al.(2024)Duque, Marquardt, Velikova, Lacourpaille, Nordez, Crouzier, Lee, Mateus, and Navab]{duque2024ultrasound}
Vanessa~Gonzalez Duque, Alexandra Marquardt, Yordanka Velikova, Lilian Lacourpaille, Antoine Nordez, Marion Crouzier, Hong~Joo Lee, Diana Mateus, and Nassir Navab.
\newblock Ultrasound segmentation analysis via distinct and completed anatomical borders.
\newblock \emph{International Journal of Computer Assisted Radiology and Surgery}, 19\penalty0 (7):\penalty0 1419--1427, 2024.

\bibitem[Felzenszwalb and Huttenlocher(2004)]{felzenszwalb2004efficient}
Pedro~F Felzenszwalb and Daniel~P Huttenlocher.
\newblock Efficient graph-based image segmentation.
\newblock \emph{International journal of computer vision}, 59:\penalty0 167--181, 2004.

\bibitem[Garrucho et~al.(2024)Garrucho, Reidel, Kushibar, Joshi, Osuala, Tsirikoglou, Bobowicz, del Riego, Catanese, Gwo{\'z}dziewicz, et~al.]{mama_mia}
Lidia Garrucho, Claire-Anne Reidel, Kaisar Kushibar, Smriti Joshi, Richard Osuala, Apostolia Tsirikoglou, Maciej Bobowicz, Javier del Riego, Alessandro Catanese, Katarzyna Gwo{\'z}dziewicz, et~al.
\newblock Mama-mia: A large-scale multi-center breast cancer dce-mri benchmark dataset with expert segmentations.
\newblock \emph{arXiv preprint arXiv:2406.13844}, 2024.

\bibitem[Gatidis and Kuestner(2022)]{Gatidis2022-ms}
Sergios Gatidis and Thomas Kuestner.
\newblock A whole-body {FDG-PET/CT} dataset with manually annotated tumor lesions ({FDG-PET-CT-Lesions}), 2022.

\bibitem["Gatidis~S(2022)]{autopet}
andKuestner~T." "Gatidis~S.
\newblock A whole-body fdg-pet/ct dataset with manually annotated tumor lesions (fdg-pet-ct-lesions).
\newblock The Cancer Imaging Archive,, 2022.

\bibitem[Gerhard et~al.(2013)Gerhard, Funke, Martel, Cardona, and Fetter]{gerhard2013segmented}
Stephan Gerhard, Jan Funke, Julien Martel, Albert Cardona, and Richard Fetter.
\newblock Segmented anisotropic sstem dataset of neural tissue.
\newblock \emph{figshare}, pages 0--0, 2013.

\bibitem[Gibson et~al.(2018)Gibson, Giganti, Hu, Bonmati, Bandula, Gurusamy, Davidson, Pereira, Clarkson, and Barratt]{BTCV2}
Eli Gibson, Francesco Giganti, Yipeng Hu, Ester Bonmati, Steve Bandula, Kurinchi Gurusamy, Brian Davidson, Stephen~P. Pereira, Matthew~J. Clarkson, and Dean~C. Barratt.
\newblock Automatic multi-organ segmentation on abdominal ct with dense v-networks.
\newblock \emph{IEEE Transactions on Medical Imaging}, 2018.

\bibitem[Gong et~al.(2024)Gong, Zhong, Ma, Li, Wang, Zhang, Heng, and Dou]{3dsamadapter}
Shizhan Gong, Yuan Zhong, Wenao Ma, Jinpeng Li, Zhao Wang, Jingyang Zhang, Pheng-Ann Heng, and Qi Dou.
\newblock 3dsam-adapter: Holistic adaptation of sam from 2d to 3d for promptable tumor segmentation.
\newblock \emph{Medical Image Analysis}, 98:\penalty0 103324, 2024.

\bibitem[Gotkowski et~al.(2024)Gotkowski, Gupta, Godinho, Tochtrop, Maier-Hein, and Isensee]{gotkowski2024particleseg3d}
Karol Gotkowski, Shuvam Gupta, Jose~RA Godinho, Camila~GS Tochtrop, Klaus~H Maier-Hein, and Fabian Isensee.
\newblock Particleseg3d: a scalable out-of-the-box deep learning segmentation solution for individual particle characterization from micro ct images in mineral processing and recycling.
\newblock \emph{Powder Technology}, 434:\penalty0 119286, 2024.

\bibitem[Gr{\o}vik et~al.(2020)Gr{\o}vik, Yi, Iv, Tong, Rubin, and Zaharchuk]{Grovik2020-un}
Endre Gr{\o}vik, Darvin Yi, Michael Iv, Elizabeth Tong, Daniel Rubin, and Greg Zaharchuk.
\newblock Deep learning enables automatic detection and segmentation of brain metastases on multisequence {MRI}.
\newblock \emph{J. Magn. Reson. Imaging}, 2020.

\bibitem[He et~al.(2021)He, Yang, Yang, Ge, Kong, Zhu, Zhang, Shao, Shu, Dillenseger, et~al.]{he2021meta}
Yuting He, Guanyu Yang, Jian Yang, Rongjun Ge, Youyong Kong, Xiaomei Zhu, Shaobo Zhang, Pengfei Shao, Huazhong Shu, Jean-Louis Dillenseger, et~al.
\newblock Meta grayscale adaptive network for 3d integrated renal structures segmentation.
\newblock \emph{Medical image analysis}, 71:\penalty0 102055, 2021.

\bibitem[He et~al.(2024)He, Guo, Tang, Myronenko, Nath, Xu, Yang, Zhao, Simon, Belue, Harmon, Turkbey, Xu, and Li]{vista3d}
Yufan He, Pengfei Guo, Yucheng Tang, Andriy Myronenko, Vishwesh Nath, Ziyue Xu, Dong Yang, Can Zhao, Benjamin Simon, Mason Belue, Stephanie Harmon, Baris Turkbey, Daguang Xu, and Wenqi Li.
\newblock Vista3d: Versatile imaging segmentation and annotation model for 3d computed tomography, 2024.

\bibitem[Heller et~al.(2023)Heller, Isensee, Trofimova, Tejpaul, Zhao, Chen, Wang, Golts, Khapun, Shats, and et~al.]{heller2023kits21}
Nicholas Heller, Fabian Isensee, Dasha Trofimova, Resha Tejpaul, Zhongchen Zhao, Huai Chen, Lisheng Wang, Alex Golts, Daniel Khapun, Daniel Shats, and et al.
\newblock The kits21 challenge: Automatic segmentation of kidneys, renal tumors, and renal cysts in corticomedullary-phase ct, 2023.

\bibitem[Huang et~al.(2023)Huang, Wang, Deng, Ye, Su, Sun, He, Gu, Gu, Zhang, and Qiao]{huang2023stunet}
Ziyan Huang, Haoyu Wang, Zhongying Deng, Jin Ye, Yanzhou Su, Hui Sun, Junjun He, Yun Gu, Lixu Gu, Shaoting Zhang, and Yu Qiao.
\newblock Stu-net: Scalable and transferable medical image segmentation models empowered by large-scale supervised pre-training, 2023.

\bibitem[Imran et~al.(2024)Imran, Krebs, Gopu, Fazzone, Sivaraman, Kumar, Viscardi, Heithaus, Shickel, Zhou, Cooper, and Shao]{IMRAN2024102470}
Muhammad Imran, Jonathan~R. Krebs, Veera Rajasekhar~Reddy Gopu, Brian Fazzone, Vishal~Balaji Sivaraman, Amarjeet Kumar, Chelsea Viscardi, Robert~Evans Heithaus, Benjamin Shickel, Yuyin Zhou, Michol~A. Cooper, and Wei Shao.
\newblock Cis-unet: Multi-class segmentation of the aorta in computed tomography angiography via context-aware shifted window self-attention.
\newblock \emph{Computerized Medical Imaging and Graphics}, 118:\penalty0 102470, 2024.

\bibitem[Isensee et~al.(2021)Isensee, Jaeger, Kohl, Petersen, and Maier-Hein]{isensee_nnu-net_2021}
Fabian Isensee, Paul~F. Jaeger, Simon A.~A. Kohl, Jens Petersen, and Klaus~H. Maier-Hein.
\newblock {nnU}-net: a self-configuring method for deep learning-based biomedical image segmentation.
\newblock \emph{Nature Methods}, 18(2)\penalty0 (2):\penalty0 203--211, 2021.

\bibitem[Isensee et~al.(2024)Isensee, Wald, Ulrich, Baumgartner, Roy, Maier-Hein, and Jaeger]{nnunet_revisited}
Fabian Isensee, Tassilo Wald, Constantin Ulrich, Michael Baumgartner, Saikat Roy, Klaus Maier-Hein, and Paul~F. Jaeger.
\newblock nnu-net revisited: A call for rigorous validation in 3d medical image segmentation, 2024.

\bibitem[Jaus et~al.(2023)Jaus, Seibold, Hermann, Walter, Giske, Haubold, Kleesiek, and Stiefelhagen]{jaus2023towards}
Alexander Jaus, Constantin Seibold, Kelsey Hermann, Alexandra Walter, Kristina Giske, Johannes Haubold, Jens Kleesiek, and Rainer Stiefelhagen.
\newblock Towards unifying anatomy segmentation: Automated generation of a full-body ct dataset via knowledge aggregation and anatomical guidelines.
\newblock \emph{arXiv preprint arXiv:2307.13375}, 2023.

\bibitem[Jensen et~al.(2024)Jensen, Clemmensen, Hansen, van Krimpen~Mortensen, Christensen, Kjaer, and Ripa]{mouseCTtumor}
Malte Jensen, Andreas Clemmensen, Jacob~Gorm Hansen, Julie van Krimpen~Mortensen, Emil~N. Christensen, Andreas Kjaer, and Rasmus~Sejersten Ripa.
\newblock 3d whole body preclinical micro-ct database of subcutaneous tumors in mice with annotations from 3 annotators.
\newblock \emph{Scientific Data}, 11\penalty0 (1):\penalty0 1021, 2024.

\bibitem[Ji et~al.(2022)Ji, Bai, GE, Yang, Zhu, Zhang, Li, Zhanng, Ma, Wan, and Luo]{ji2022amos}
Yuanfeng Ji, Haotian Bai, Chongjian GE, Jie Yang, Ye Zhu, Ruimao Zhang, Zhen Li, Lingyan Zhanng, Wanling Ma, Xiang Wan, and Ping Luo.
\newblock Amos: A large-scale abdominal multi-organ benchmark for versatile medical image segmentation.
\newblock In \emph{Advances in Neural Information Processing Systems}, 2022.

\bibitem[Jin et~al.(2021)Jin, Pepe, Li, Gsaxner, Zhao, Pomykala, Kleesiek, Frangi, and Egger]{Jin2021-jo}
Yuan Jin, Antonio Pepe, Jianning Li, Christina Gsaxner, Fen-Hua Zhao, Kelsey~L Pomykala, Jens Kleesiek, Alejandro~F Frangi, and Jan Egger.
\newblock {AI-based} aortic vessel tree segmentation for cardiovascular diseases treatment: Status quo.
\newblock \emph{arXiv}, 2021.

\bibitem[Jordan et~al.(2022)Jordan, Adamson, Bhattbhatt, Beriwal, Shen, Radermecker, Bose, Strain, Offe, Fraley, et~al.]{jordan2022pediatric}
Petr Jordan, Philip~M Adamson, Vrunda Bhattbhatt, Surabhi Beriwal, Sangyu Shen, Oskar Radermecker, Supratik Bose, Linda~S Strain, Michael Offe, David Fraley, et~al.
\newblock Pediatric chest-abdomen-pelvis and abdomen-pelvis ct images with expert organ contours.
\newblock \emph{Medical physics}, 49\penalty0 (5):\penalty0 3523--3528, 2022.

\bibitem[Jun et~al.(2020)Jun, Cheng, Yixin, Xingle, Jiantao, Ziqi, Minqing, Xin, Xueyuan, Shucheng, Hao, Sen, Xiaoyu, Ziwei, Chen, Lu, Yuntao, Qiongjie, Guoqiang, and Jian]{covid19_junma}
Ma Jun, Ge Cheng, Wang Yixin, An Xingle, Gao Jiantao, Yu Ziqi, Zhang Minqing, Liu Xin, Deng Xueyuan, Cao Shucheng, Wei Hao, Mei Sen, Yang Xiaoyu, Nie Ziwei, Li Chen, Tian Lu, Zhu Yuntao, Zhu Qiongjie, Dong Guoqiang, and He Jian.
\newblock {COVID-19 CT Lung and Infection Segmentation Dataset}, 2020.

\bibitem[Juvekar et~al.(2024)Juvekar, Dorent, K{\"o}gl, Torio, Barr, Rigolo, Galvin, Jowkar, Kazi, Haouchine, Cheema, Navab, Pieper, Wells, Bi, Golby, Frisken, and Kapur]{remind}
Parikshit Juvekar, Reuben Dorent, Fryderyk K{\"o}gl, Erickson Torio, Colton Barr, Laura Rigolo, Colin Galvin, Nick Jowkar, Anees Kazi, Nazim Haouchine, Harneet Cheema, Nassir Navab, Steve Pieper, William~M. Wells, Wenya~Linda Bi, Alexandra Golby, Sarah Frisken, and Tina Kapur.
\newblock Remind: The brain resection multimodal imaging database.
\newblock \emph{Scientific Data}, 11\penalty0 (1):\penalty0 494, 2024.

\bibitem[Kadish et~al.(2009)Kadish, Bello, Finn, Bonow, Schaechter, Subacius, Albert, Daubert, Fonseca, and Goldberger]{kadish2009rationale}
Alan~H Kadish, David Bello, J~Paul Finn, Robert~O Bonow, Andi Schaechter, Haris Subacius, Christine Albert, James~P Daubert, Carissa~G Fonseca, and Jeffrey~J Goldberger.
\newblock Rationale and design for the defibrillators to reduce risk by magnetic resonance imaging evaluation (determine) trial.
\newblock \emph{Journal of cardiovascular electrophysiology}, 20\penalty0 (9):\penalty0 982--987, 2009.

\bibitem[Kanwar et~al.(2023)Kanwar, Merz, Claunch, Rana, Hung, and Thompson]{kanwar2023stress}
Aasheesh Kanwar, Brandon Merz, Cheryl Claunch, Shushan Rana, Arthur Hung, and Reid~F Thompson.
\newblock Stress-testing pelvic autosegmentation algorithms using anatomical edge cases.
\newblock \emph{Physics and Imaging in Radiation Oncology}, 25:\penalty0 100413, 2023.

\bibitem[Karargyris et~al.(2023)Karargyris, Umeton, Sheller, Aristizabal, George, Wuest, Pati, Kassem, Zenk, Baid, and et~al.]{Karargyris2023}
Alexandros Karargyris, Renato Umeton, Micah~J. Sheller, Alejandro Aristizabal, Johnu George, Anna Wuest, Sarthak Pati, Hasan Kassem, Maximilian Zenk, Ujjwal Baid, and et al.
\newblock Federated benchmarking of medical artificial intelligence with medperf.
\newblock \emph{Nature Machine Intelligence}, 2023.

\bibitem[Kavur et~al.(2019)Kavur, Selver, Dicle, Baris, and Gezer]{CHAOSdata2019}
Ali~Emre Kavur, M.~Alper Selver, Oguz Dicle, Mustafa Baris, and N.~Sinem Gezer.
\newblock Chaos - combined (ct-mr) healthy abdominal organ segmentation challenge data, 2019.

\bibitem[Kinahan et~al.(2019)Kinahan, Muzi, Bialecki, and Coombs]{Kinahan2019acrin}
P. Kinahan, M. Muzi, B. Bialecki, and L. Coombs.
\newblock Data from the acrin 6685 trial hnscc-fdg-pet/ct.
\newblock Data set, 2019.

\bibitem[Kirillov et~al.(2023)Kirillov, Mintun, Ravi, Mao, Rolland, Gustafson, Xiao, Whitehead, Berg, Lo, et~al.]{segment_anything}
Alexander Kirillov, Eric Mintun, Nikhila Ravi, Hanzi Mao, Chloe Rolland, Laura Gustafson, Tete Xiao, Spencer Whitehead, Alexander~C Berg, Wan-Yen Lo, et~al.
\newblock Segment anything.
\newblock In \emph{Proceedings of the IEEE/CVF International Conference on Computer Vision}, pages 4015--4026, 2023.

\bibitem[Kiser et~al.(2020)Kiser, Ahmed, Stieb, Mohamed, Elhalawani, Park, Doyle, Wang, Barman, Li, et~al.]{kiser2020plethora}
Kendall~J Kiser, Sara Ahmed, Sonja Stieb, Abdallah~SR Mohamed, Hesham Elhalawani, Peter~YS Park, Nathan~S Doyle, Brandon~J Wang, Arko Barman, Zhao Li, et~al.
\newblock Plethora: Pleural effusion and thoracic cavity segmentations in diseased lungs for benchmarking chest ct processing pipelines.
\newblock \emph{Medical physics}, 47\penalty0 (11):\penalty0 5941--5952, 2020.

\bibitem[Kr{\"o}nke et~al.(2022)Kr{\"o}nke, Eilers, Dimova, K{\"o}hler, Buschner, Schweiger, Konstantinidou, Makowski, Nagarajah, Navab, et~al.]{kronke2022tracked}
Markus Kr{\"o}nke, Christine Eilers, Desislava Dimova, Melanie K{\"o}hler, Gabriel Buschner, Lilit Schweiger, Lemonia Konstantinidou, Marcus Makowski, James Nagarajah, Nassir Navab, et~al.
\newblock Tracked 3d ultrasound and deep neural network-based thyroid segmentation reduce interobserver variability in thyroid volumetry.
\newblock \emph{Plos one}, 17\penalty0 (7):\penalty0 e0268550, 2022.

\bibitem[Kuijf et~al.(2024)Kuijf, Bennink, Vincken, Weaver, Biessels, and Viergever]{kuijf2024mr}
HJ Kuijf, E Bennink, KL Vincken, N Weaver, GJ Biessels, and MA Viergever.
\newblock Mr brain segmentation challenge 2018 data.
\newblock \emph{DOI: https://doi. org/10.34894/E0U32Q}, 2024.

\bibitem[Kuijf et~al.(2019)Kuijf, Biesbroek, De~Bresser, Heinen, Andermatt, Bento, Berseth, Belyaev, Cardoso, Casamitjana, et~al.]{kuijf2019standardized}
Hugo~J Kuijf, J~Matthijs Biesbroek, Jeroen De~Bresser, Rutger Heinen, Simon Andermatt, Mariana Bento, Matt Berseth, Mikhail Belyaev, M~Jorge Cardoso, Adria Casamitjana, et~al.
\newblock Standardized assessment of automatic segmentation of white matter hyperintensities and results of the wmh segmentation challenge.
\newblock \emph{IEEE transactions on medical imaging}, 38\penalty0 (11):\penalty0 2556--2568, 2019.

\bibitem[Lalande et~al.(2020)Lalande, Chen, Decourselle, Qayyum, Pommier, Lorgis, de~La~Rosa, Cochet, Cottin, Ginhac, et~al.]{lalande2020emidec}
Alain Lalande, Zhihao Chen, Thomas Decourselle, Abdul Qayyum, Thibaut Pommier, Luc Lorgis, Ezequiel de La~Rosa, Alexandre Cochet, Yves Cottin, Dominique Ginhac, et~al.
\newblock Emidec: a database usable for the automatic evaluation of myocardial infarction from delayed-enhancement cardiac mri.
\newblock \emph{Data}, 5\penalty0 (4):\penalty0 89, 2020.

\bibitem[Lambert et~al.(2019)Lambert, Petitjean, Dubray, and Ruan]{lambert2019segthor}
Z. Lambert, C. Petitjean, B. Dubray, and S. Ruan.
\newblock Segthor: Segmentation of thoracic organs at risk in ct images.
\newblock \emph{arXiv:1912.05950}, 2019.

\bibitem[Landman et~al.(2015)Landman, Xu, Igelsias, Styner, and et~al.]{BTCV}
Bennett Landman, Zhoubing Xu, Juan~Eugenio Igelsias, Martin Styner, and et al.
\newblock 2015 miccai multi-atlas labeling beyond the cranial vault workshop and challenge.
\newblock In \emph{Proc. MICCAI Multi-Atlas Labeling Beyond Cranial Vault—Workshop Challenge}, 2015.

\bibitem[Leclerc et~al.(2019)Leclerc, Smistad, Pedrosa, {\O}stvik, Cervenansky, Espinosa, Espeland, Berg, Jodoin, Grenier, et~al.]{leclerc2019deep}
Sarah Leclerc, Erik Smistad, Joao Pedrosa, Andreas {\O}stvik, Frederic Cervenansky, Florian Espinosa, Torvald Espeland, Erik Andreas~Rye Berg, Pierre-Marc Jodoin, Thomas Grenier, et~al.
\newblock Deep learning for segmentation using an open large-scale dataset in 2d echocardiography.
\newblock \emph{IEEE transactions on medical imaging}, 38\penalty0 (9):\penalty0 2198--2210, 2019.

\bibitem[Li et~al.(2019)Li, Zhou, Deng, and Chen]{structseg}
Hongsheng Li, Jinghao Zhou, Jincheng Deng, and Ming Chen.
\newblock Automatic structure segmentation for radiotherapy planning challenge, 2019.
\newblock https://structseg2019.grand-challenge.org/ 25/02/2022).

\bibitem[Li et~al.(2024{\natexlab{a}})Li, Liu, Hu, Wang, and Oguz]{prism}
Hao Li, Han Liu, Dewei Hu, Jiacheng Wang, and Ipek Oguz.
\newblock Prism: A promptable and robust interactive segmentation model with visual prompts, 2024{\natexlab{a}}.

\bibitem[Li et~al.(2024{\natexlab{b}})Li, Yuille, and Zhou]{li2024well}
Wenxuan Li, Alan Yuille, and Zongwei Zhou.
\newblock How well do supervised models transfer to 3d image segmentation?
\newblock In \emph{The Twelfth International Conference on Learning Representations}, 2024{\natexlab{b}}.

\bibitem[Li et~al.(2023)Li, Luo, Wang, Wang, Liu, Liang, Jiang, Song, Zheng, Chi, Xu, He, Ma, Guo, Liu, Li, Chen, Siddiquee, Myronenko, Sanner, Mukhopadhyay, Othman, Zhao, Liu, Zhang, Ma, Liu, MacIntosh, Liang, Mazher, Qayyum, Abramova, Lladó, and Li]{instance22}
Xiangyu Li, Gongning Luo, Kuanquan Wang, Hongyu Wang, Jun Liu, Xinjie Liang, Jie Jiang, Zhenghao Song, Chunyue Zheng, Haokai Chi, Mingwang Xu, Yingte He, Xinghua Ma, Jingwen Guo, Yifan Liu, Chuanpu Li, Zeli Chen, Md~Mahfuzur~Rahman Siddiquee, Andriy Myronenko, Antoine~P. Sanner, Anirban Mukhopadhyay, Ahmed~E. Othman, Xingyu Zhao, Weiping Liu, Jinhuang Zhang, Xiangyuan Ma, Qinghui Liu, Bradley~J. MacIntosh, Wei Liang, Moona Mazher, Abdul Qayyum, Valeriia Abramova, Xavier Lladó, and Shuo Li.
\newblock The state-of-the-art 3d anisotropic intracranial hemorrhage segmentation on non-contrast head ct: The instance challenge, 2023.

\bibitem[Liao et~al.(2023)Liao, Luo, He, Dong, Li, Li, Zhang, Zhang, Wang, and Xiao]{liao2023comprehensive}
Wenjun Liao, Xiangde Luo, Yuan He, Ye Dong, Churong Li, Kang Li, Shichuan Zhang, Shaoting Zhang, Guotai Wang, and Jianghong Xiao.
\newblock Comprehensive evaluation of a deep learning model for automatic organs-at-risk segmentation on heterogeneous computed tomography images for abdominal radiation therapy.
\newblock \emph{International Journal of Radiation Oncology* Biology* Physics}, 117\penalty0 (4):\penalty0 994--1006, 2023.

\bibitem[Liebl et~al.(2021)Liebl, Schinz, Sekuboyina, Malagutti, L\"{o}ffler, Bayat, Husseini, Tetteh, Grau, and Niederreiter]{verse3}
Hans Liebl, David Schinz, Anjany Sekuboyina, Luca Malagutti, Maximilian~T. L\"{o}ffler, Amirhossein Bayat, Malek~El Husseini, Giles Tetteh, Katharina Grau, and et~al. Niederreiter.
\newblock A computed tomography vertebral segmentation dataset with anatomical variations and multi-vendor scanner data, 2021.

\bibitem[Liew et~al.(2018)Liew, Anglin, Banks, Sondag, Ito, Kim, Chan, Ito, Jung, Khoshab, and at~al.]{Liew2018-wy}
Sook-Lei Liew, Julia~M Anglin, Nick~W Banks, Matt Sondag, Kaori~L Ito, Hosung Kim, Jennifer Chan, Joyce Ito, Connie Jung, Nima Khoshab, and at al.
\newblock A large, open source dataset of stroke anatomical brain images and manual lesion segmentations.
\newblock \emph{Sci. Data}, 2018.

\bibitem[Lin et~al.(2021)Lin, Wei, Petkova, Wu, Ahmed, K, Zou, Wendt, Boulanger-Weill, Wang, Dhanyasi, Arganda-Carreras, Engert, Lichtman, and Pfister]{Lin_2021}
Zudi Lin, Donglai Wei, Mariela~D. Petkova, Yuelong Wu, Zergham Ahmed, Krishna~Swaroop K, Silin Zou, Nils Wendt, Jonathan Boulanger-Weill, Xueying Wang, Nagaraju Dhanyasi, Ignacio Arganda-Carreras, Florian Engert, Jeff Lichtman, and Hanspeter Pfister.
\newblock \emph{NucMM Dataset: 3D Neuronal Nuclei Instance Segmentation at Sub-Cubic Millimeter Scale}, page 164–174.
\newblock Springer International Publishing, 2021.

\bibitem[Litjens et~al.(2014)Litjens, Toth, van~de Ven, Hoeks, Kerkstra, van Ginneken, Vincent, Guillard, Birbeck, and Zhang]{Litjens2014}
Geert Litjens, Robert Toth, Wendy van~de Ven, Caroline Hoeks, Sjoerd Kerkstra, Bram van Ginneken, Graham Vincent, Gwenael Guillard, Neil Birbeck, and et~al. Zhang.
\newblock Evaluation of prostate segmentation algorithms for mri: The promise12 challenge.
\newblock \emph{Medical Image Analysis}, 2014.

\bibitem[Litjens et~al.(2017)Litjens, Debats, Barentsz, Karssemeijer, and Huisman]{prostatex}
Geert Litjens, Oscar Debats, Jelle Barentsz, Nico Karssemeijer, and Henkjan Huisman.
\newblock Spie-aapm prostatex challenge data, 2017.

\bibitem[Liu et~al.(2023)Liu, Yibulayimu, Sang, Zhu, Wang, Zhao, and Wu]{liu2023pelvic}
Yanzhen Liu, Sutuke Yibulayimu, Yudi Sang, Gang Zhu, Yu Wang, Chunpeng Zhao, and Xinbao Wu.
\newblock Pelvic fracture segmentation using a multi-scale distance-weighted neural network.
\newblock In \emph{International Conference on Medical Image Computing and Computer-Assisted Intervention}, pages 312--321. Springer, 2023.

\bibitem[Ljosa et~al.(2012)Ljosa, Sokolnicki, and Carpenter]{ljosa2012annotated}
Vebjorn Ljosa, Katherine~L Sokolnicki, and Anne~E Carpenter.
\newblock Annotated high-throughput microscopy image sets for validation.
\newblock \emph{Nature methods}, 9\penalty0 (7):\penalty0 637, 2012.

\bibitem[L\"{o}ffler et~al.(2020)L\"{o}ffler, Sekuboyina, Jacob, Grau, Scharr, El~Husseini, Kallweit, Zimmer, Baum, and Kirschke]{verse2}
Maximilian~T. L\"{o}ffler, Anjany Sekuboyina, Alina Jacob, Anna-Lena Grau, Andreas Scharr, Malek El~Husseini, Mareike Kallweit, Claus Zimmer, Thomas Baum, and Jan~S. Kirschke.
\newblock A vertebral segmentation dataset with fracture grading.
\newblock \emph{Radiology: Artificial Intelligence}, 2020.

\bibitem[Lucchi et~al.(2011)Lucchi, Smith, Achanta, Knott, and Fua]{lucchi2011supervoxel}
Aur{\'e}lien Lucchi, Kevin Smith, Radhakrishna Achanta, Graham Knott, and Pascal Fua.
\newblock Supervoxel-based segmentation of mitochondria in em image stacks with learned shape features.
\newblock \emph{IEEE transactions on medical imaging}, 31\penalty0 (2):\penalty0 474--486, 2011.

\bibitem[Luo et~al.(2023{\natexlab{a}})Luo, Wang, Liu, Li, Liang, Li, Gan, Wang, Dong, Wang, et~al.]{luo2023efficient}
Gongning Luo, Kuanquan Wang, Jun Liu, Shuo Li, Xinjie Liang, Xiangyu Li, Shaowei Gan, Wei Wang, Suyu Dong, Wenyi Wang, et~al.
\newblock Efficient automatic segmentation for multi-level pulmonary arteries: The parse challenge.
\newblock \emph{arXiv preprint arXiv:2304.03708}, 2023{\natexlab{a}}.

\bibitem[Luo et~al.(2022)Luo, Liao, Xiao, Chen, Song, Zhang, Li, Metaxas, Wang, and Zhang]{luo2022word}
Xiangde Luo, Wenjun Liao, Jianghong Xiao, Jieneng Chen, Tao Song, Xiaofan Zhang, Kang Li, Dimitris~N. Metaxas, Guotai Wang, and Shaoting Zhang.
\newblock Word: A large scale dataset, benchmark and clinical applicable study for abdominal organ segmentation from ct image.
\newblock \emph{Medical Image Analysis}, 2022.

\bibitem[Luo et~al.(2023{\natexlab{b}})Luo, Fu, Zhong, Liu, Han, Astaraki, Bendazzoli, Toma-Dasu, Ye, Chen, et~al.]{luo2023segrap2023}
Xiangde Luo, Jia Fu, Yunxin Zhong, Shuolin Liu, Bing Han, Mehdi Astaraki, Simone Bendazzoli, Iuliana Toma-Dasu, Yiwen Ye, Ziyang Chen, et~al.
\newblock Segrap2023: A benchmark of organs-at-risk and gross tumor volume segmentation for radiotherapy planning of nasopharyngeal carcinoma.
\newblock \emph{arXiv preprint arXiv:2312.09576}, 2023{\natexlab{b}}.

\bibitem[Ma et~al.(2022)Ma, Zhang, Gu, Zhu, Ge, Zhang, An, Wang, Wang, Liu, Cao, Zhang, Liu, Wang, Li, He, and Yang]{Ma-2021-AbdomenCT-1K}
Jun Ma, Yao Zhang, Song Gu, Cheng Zhu, Cheng Ge, Yichi Zhang, Xingle An, Congcong Wang, Qiyuan Wang, Xin Liu, Shucheng Cao, Qi Zhang, Shangqing Liu, Yunpeng Wang, Yuhui Li, Jian He, and Xiaoping Yang.
\newblock Abdomenct-1k: Is abdominal organ segmentation a solved problem?
\newblock \emph{IEEE Transactions on Pattern Analysis and Machine Intelligence}, 2022.

\bibitem[Ma et~al.(2023)Ma, Zhang, Gu, Ge, Ma, Young, Zhu, Meng, Yang, Huang, Zhang, Liu, and ant~et al.]{FLARE22}
Jun Ma, Yao Zhang, Song Gu, Cheng Ge, Shihao Ma, Adamo Young, Cheng Zhu, Kangkang Meng, Xin Yang, Ziyan Huang, Fan Zhang, Wentao Liu, and YuanKe~Pan ant~et al.
\newblock Unleashing the strengths of unlabeled data in pan-cancer abdominal organ quantification: the flare22 challenge.
\newblock \emph{arXiv preprint arXiv:2308.05862}, 2023.

\bibitem[Ma et~al.(2024{\natexlab{a}})Ma, He, Li, Han, You, and Wang]{medsam}
Jun Ma, Yuting He, Feifei Li, Lin Han, Chenyu You, and Bo Wang.
\newblock Segment anything in medical images.
\newblock \emph{Nature Communications}, 15\penalty0 (1):\penalty0 654, 2024{\natexlab{a}}.

\bibitem[Ma et~al.(2024{\natexlab{b}})Ma, Zhang, Gu, Ge, Wang, Zhou, Huang, Lyu, He, and Wang]{FLARE23}
Jun Ma, Yao Zhang, Song Gu, Cheng Ge, Ershuai Wang, Qin Zhou, Ziyan Huang, Pengju Lyu, Jian He, and Bo Wang.
\newblock Automatic organ and pan-cancer segmentation in abdomen ct: the flare 2023 challenge, 2024{\natexlab{b}}.

\bibitem[Macdonald et~al.(2020)Macdonald, Zhu, Konkel, Mazurowski, Wiggins, and Bashir]{macdonald_2020_7774566}
Jacob~A. Macdonald, Zhe Zhu, Brandon Konkel, Maciej Mazurowski, Walter Wiggins, and Mustafa Bashir.
\newblock Duke liver dataset (mri) v2, 2020.

\bibitem[Maier et~al.(2015)Maier, Wilms, {von der Gablentz}, Krämer, Münte, and Handels]{isles2015}
Oskar Maier, Matthias Wilms, Janina {von der Gablentz}, Ulrike~M. Krämer, Thomas~F. Münte, and Heinz Handels.
\newblock Extra tree forests for sub-acute ischemic stroke lesion segmentation in mr sequences.
\newblock \emph{Journal of Neuroscience Methods}, 2015.

\bibitem[Marcus et~al.(2007)Marcus, Wang, Parker, Csernansky, Morris, and Buckner]{marcus2007open}
Daniel~S Marcus, Tracy~H Wang, Jamie Parker, John~G Csernansky, John~C Morris, and Randy~L Buckner.
\newblock Open access series of imaging studies (oasis): cross-sectional mri data in young, middle aged, nondemented, and demented older adults.
\newblock \emph{Journal of cognitive neuroscience}, 19\penalty0 (9):\penalty0 1498--1507, 2007.

\bibitem[Martín-Isla et~al.(2023)Martín-Isla, Campello, Izquierdo, Kushibar, Sendra-Balcells, Gkontra, Sojoudi, Fulton, Arega, and Punithakumar]{10103611}
Carlos Martín-Isla, Víctor~M. Campello, Cristian Izquierdo, Kaisar Kushibar, Carla Sendra-Balcells, Polyxeni Gkontra, Alireza Sojoudi, Mitchell~J. Fulton, Tewodros~Weldebirhan Arega, and et~al. Punithakumar.
\newblock Deep learning segmentation of the right ventricle in cardiac mri: The mms challenge.
\newblock \emph{IEEE Journal of Biomedical and Health Informatics}, 2023.

\bibitem[Ma{\v{s}}ka et~al.(2023)Ma{\v{s}}ka, Ulman, Delgado-Rodriguez, G{\'o}mez-de Mariscal, Ne{\v{c}}asov{\'a}, Guerrero~Pe{\~n}a, Ren, Meyerowitz, Scherr, L{\"o}ffler, et~al.]{mavska2023cell}
Martin Ma{\v{s}}ka, Vladim{\'\i}r Ulman, Pablo Delgado-Rodriguez, Estibaliz G{\'o}mez-de Mariscal, Tereza Ne{\v{c}}asov{\'a}, Fidel~A Guerrero~Pe{\~n}a, Tsang~Ing Ren, Elliot~M Meyerowitz, Tim Scherr, Katharina L{\"o}ffler, et~al.
\newblock The cell tracking challenge: 10 years of objective benchmarking.
\newblock \emph{Nature Methods}, 20\penalty0 (7):\penalty0 1010--1020, 2023.

\bibitem[Mayr et~al.(2023)Mayr, Yuh, Bowen, Harkenrider, Knopp, Lee, Leung, Lo, Small~Jr., and Wolfson]{Mayr2023}
N. Mayr, W.~T.~C. Yuh, S. Bowen, M. Harkenrider, M.~V. Knopp, E.~Y.-P. Lee, E. Leung, S.~S. Lo, W. Small~Jr., and A.~H. Wolfson.
\newblock Cervical cancer – tumor heterogeneity: Serial functional and molecular imaging across the radiation therapy course in advanced cervical cancer (version 1), 2023.
\newblock Data set.

\bibitem[Menze et~al.(2015)Menze, Jakab, Bauer, Kalpathy-Cramer, Farahani, Kirby, Burren, Porz, and et~al.]{bratsgli2}
Bjoern~H Menze, Andras Jakab, Stefan Bauer, Jayashree Kalpathy-Cramer, Keyvan Farahani, Justin Kirby, Yuliya Burren, Nicole Porz, and et al.
\newblock The multimodal brain tumor image segmentation benchmark ({BRATS}).
\newblock \emph{IEEE Trans. Med. Imaging}, 2015.

\bibitem[Moawad et~al.(2021)Moawad, Fuentes, Morshid, Khalaf, Elmohr, Abusaif, Hazle, Kaseb, Hassan, Mahvash, Szklaruk, Qayyom, and Elsayes]{hcctace}
Ahmed~W. Moawad, David Fuentes, Ali Morshid, Ahmed~M. Khalaf, Mohab~M. Elmohr, Abdelrahman Abusaif, John~D. Hazle, Ahmed~O. Kaseb, Manal Hassan, Armeen Mahvash, Janio Szklaruk, Aliyya Qayyom, and Khaled Elsayes.
\newblock Multimodality annotated hcc cases with and without advanced imaging segmentation, 2021.

\bibitem[Moawad et~al.(2023)Moawad, Ahmed, ElMohr, Eltaher, Habra, Fisher, Perrier, Zhang, Fuentes, and Elsayes]{adrenalacc}
Ahmed~W. Moawad, Ayahallah~A. Ahmed, Mohab ElMohr, Mohamed Eltaher, Mouhammed~Amir Habra, Sarah Fisher, Nancy Perrier, Miao Zhang, David Fuentes, and Khaled Elsayes.
\newblock Voxel-level segmentation of pathologically-proven adrenocortical carcinoma with ki-67 expression (adrenal-acc-ki67-seg), 2023.

\bibitem[Muslim et~al.(2022)Muslim, Mashohor, Gawwam, Mahmud, binti Hanafi, Alnuaimi, Josephine, and Almutairi]{muslim2022ms}
Ali~M. Muslim, Syamsiah Mashohor, Gheyath~Al Gawwam, Rozi Mahmud, Marsyita binti Hanafi, Osama Alnuaimi, Raad Josephine, and Abdullah~Dhaifallah Almutairi.
\newblock Brain mri dataset of multiple sclerosis with consensus manual lesion segmentation and patient meta information.
\newblock \emph{Data in Brief}, 2022.

\bibitem[Nolden et~al.(2013)Nolden, Zelzer, Seitel, Wald, M{\"u}ller, Franz, Maleike, Fangerau, Baumhauer, Maier-Hein, et~al.]{MITK}
Marco Nolden, Sascha Zelzer, Alexander Seitel, Diana Wald, Michael M{\"u}ller, Alfred~M Franz, Daniel Maleike, Markus Fangerau, Matthias Baumhauer, Lena Maier-Hein, et~al.
\newblock The medical imaging interaction toolkit: challenges and advances: 10 years of open-source development.
\newblock \emph{International journal of computer assisted radiology and surgery}, 8:\penalty0 607--620, 2013.

\bibitem[Pace et~al.(2024)Pace, Contreras, Romanowicz, Ghelani, Rahaman, Zhang, Gao, Jubair, Yeh, Golland, et~al.]{pace2024hvsmr}
Danielle~F Pace, Hannah~TM Contreras, Jennifer Romanowicz, Shruti Ghelani, Imon Rahaman, Yue Zhang, Patricia Gao, Mohammad~Imrul Jubair, Tom Yeh, Polina Golland, et~al.
\newblock Hvsmr-2.0: A 3d cardiovascular mr dataset for whole-heart segmentation in congenital heart disease.
\newblock \emph{Scientific Data}, 11\penalty0 (1):\penalty0 721, 2024.

\bibitem[Payette et~al.(2021)Payette, de~Dumast, Kebiri, Ezhov, Paetzold, Shit, Iqbal, Khan, Kottke, Grehten, et~al.]{payette2021automatic}
Kelly Payette, Priscille de Dumast, Hamza Kebiri, Ivan Ezhov, Johannes~C Paetzold, Suprosanna Shit, Asim Iqbal, Romesa Khan, Raimund Kottke, Patrice Grehten, et~al.
\newblock An automatic multi-tissue human fetal brain segmentation benchmark using the fetal tissue annotation dataset.
\newblock \emph{Scientific data}, 8\penalty0 (1):\penalty0 167, 2021.

\bibitem[Pedrosa et~al.(2021)Pedrosa, Aresta, Ferreira, Atwal, Phoulady, Chen, Chen, Li, Wang, Galdran, et~al.]{lndb}
Joao Pedrosa, Guilherme Aresta, Carlos Ferreira, Gurraj Atwal, Hady~Ahmady Phoulady, Xiaoyu Chen, Rongzhen Chen, Jiaoliang Li, Liansheng Wang, Adrian Galdran, et~al.
\newblock Lndb challenge on automatic lung cancer patient management.
\newblock \emph{Medical image analysis}, 70:\penalty0 102027, 2021.

\bibitem[Pepe et~al.(2020)Pepe, Li, Rolf-Pissarczyk, Gsaxner, Chen, Holzapfel, and Egger]{Pepe2020-fh}
Antonio Pepe, Jianning Li, Malte Rolf-Pissarczyk, Christina Gsaxner, Xiaojun Chen, Gerhard~A Holzapfel, and Jan Egger.
\newblock Detection, segmentation, simulation and visualization of aortic dissections: A review.
\newblock \emph{Med. Image Anal.}, 2020.

\bibitem[Pieper et~al.(2024)Pieper, Haouchine, Hackney, Wells, Sanhinova, Balboni, Spektor, Huynh, Tanguturi, Kim, Guenette, Kozono, Czajkowski, Caplan, Doyle, Kang, and Alkalay]{pieper2024spinemets}
S. Pieper, N. Haouchine, D.~B. Hackney, W.~M. Wells, M. Sanhinova, T. Balboni, A. Spektor, M. Huynh, S. Tanguturi, E. Kim, J.~P. Guenette, D.~E. Kozono, B. Czajkowski, S. Caplan, P. Doyle, H. Kang, and R.~N. Alkalay.
\newblock Spine metastatic bone cancer: pre and post radiotherapy ct (spine-mets-ct-seg) [dataset] (version 1), 2024.

\bibitem[Podobnik et~al.(2023)Podobnik, Strojan, Peterlin, Ibragimov, and Vrtovec]{podobnik2023han}
Ga{\v{s}}per Podobnik, Primo{\v{z}} Strojan, Primo{\v{z}} Peterlin, Bulat Ibragimov, and Toma{\v{z}} Vrtovec.
\newblock Han-seg: The head and neck organ-at-risk ct and mr segmentation dataset.
\newblock \emph{Medical physics}, 50\penalty0 (3):\penalty0 1917--1927, 2023.

\bibitem[Qu et~al.(2023{\natexlab{a}})Qu, Zhang, Qiao, Liu, Tang, Yuille, and Zhou]{abdomenatlas8k}
Chongyu Qu, Tiezheng Zhang, Hualin Qiao, Jie Liu, Yucheng Tang, Alan Yuille, and Zongwei Zhou.
\newblock Abdomenatlas-8k: Annotating 8,000 ct volumes for multi-organ segmentation in three weeks, 2023{\natexlab{a}}.

\bibitem[Qu et~al.(2023{\natexlab{b}})Qu, Zhang, Qiao, Tang, Yuille, and Zhou]{qu2023abdomenatlas}
Chongyu Qu, Tiezheng Zhang, Hualin Qiao, Yucheng Tang, Alan~L Yuille, and Zongwei Zhou.
\newblock Abdomenatlas-8k: Annotating 8,000 ct volumes for multi-organ segmentation in three weeks.
\newblock \emph{Advances in Neural Information Processing Systems}, 2023{\natexlab{b}}.

\bibitem[Quinton et~al.(2023)Quinton, Popoff, Presles, Leclerc, Meriaudeau, Nodari, Lopez, Pellegrinelli, Chevallier, Ginhac, et~al.]{quinton2023tumour}
F{\'e}lix Quinton, Romain Popoff, Beno{\^\i}t Presles, Sarah Leclerc, Fabrice Meriaudeau, Guillaume Nodari, Olivier Lopez, Julie Pellegrinelli, Olivier Chevallier, Dominique Ginhac, et~al.
\newblock A tumour and liver automatic segmentation (atlas) dataset on contrast-enhanced magnetic resonance imaging for hepatocellular carcinoma.
\newblock \emph{Data}, 8\penalty0 (5):\penalty0 79, 2023.

\bibitem[Radl et~al.(2022)Radl, Jin, Pepe, Li, Gsaxner, Zhao, and Egger]{Radl2022-fz}
Lukas Radl, Yuan Jin, Antonio Pepe, Jianning Li, Christina Gsaxner, Fen-Hua Zhao, and Jan Egger.
\newblock {AVT}: Multicenter aortic vessel tree {CTA} dataset collection with ground truth segmentation masks.
\newblock \emph{Data Brief}, 2022.

\bibitem[Raudaschl et~al.(2017)Raudaschl, Zaffino, Sharp, Spadea, Chen, Dawant, Albrecht, Gass, Langguth, L{\"u}thi, et~al.]{raudaschl2017evaluation}
Patrik~F Raudaschl, Paolo Zaffino, Gregory~C Sharp, Maria~Francesca Spadea, Antong Chen, Benoit~M Dawant, Thomas Albrecht, Tobias Gass, Christoph Langguth, Marcel L{\"u}thi, et~al.
\newblock Evaluation of segmentation methods on head and neck ct: auto-segmentation challenge 2015.
\newblock \emph{Medical physics}, 44\penalty0 (5):\penalty0 2020--2036, 2017.

\bibitem[Ravi et~al.(2024)Ravi, Gabeur, Hu, Hu, Ryali, Ma, Khedr, Rädle, Rolland, Gustafson, Mintun, Pan, Alwala, Carion, Wu, Girshick, Dollár, and Feichtenhofer]{sam2}
Nikhila Ravi, Valentin Gabeur, Yuan-Ting Hu, Ronghang Hu, Chaitanya Ryali, Tengyu Ma, Haitham Khedr, Roman Rädle, Chloe Rolland, Laura Gustafson, Eric Mintun, Junting Pan, Kalyan~Vasudev Alwala, Nicolas Carion, Chao-Yuan Wu, Ross Girshick, Piotr Dollár, and Christoph Feichtenhofer.
\newblock Sam 2: Segment anything in images and videos, 2024.

\bibitem[Riera-Marín et~al.(2024)Riera-Marín, Kleiß, Aubanell, and Antolín]{RieraMarin2024}
M. Riera-Marín, J.-M. Kleiß, A. Aubanell, and A. Antolín.
\newblock Curvas dataset (v1.0.1).
\newblock MEDICAL IMAGE COMPUTING AND COMPUTER ASSISTED INTERVENTION (MICCAI), Marrakesch, 2024.
\newblock Data set.

\bibitem[Rister et~al.(2019)Rister, Shivakumar, Nobashi, and Rubin]{Rister2019tm}
Blaine Rister, Kaushik Shivakumar, Tomomi Nobashi, and Daniel~L Rubin.
\newblock {CT-ORG}: A dataset of {CT} volumes with multiple organ segmentations, 2019.

\bibitem[Rokuss et~al.(2024)Rokuss, Kovacs, Kirchhoff, Xiao, Ulrich, Maier-Hein, and Isensee]{autopet2024fdgpsma}
Maximilian Rokuss, Balint Kovacs, Yannick Kirchhoff, Shuhan Xiao, Constantin Ulrich, Klaus~H. Maier-Hein, and Fabian Isensee.
\newblock From fdg to psma: A hitchhiker's guide to multitracer, multicenter lesion segmentation in pet/ct imaging, 2024.

\bibitem[Rokuss et~al.(2025)Rokuss, Kirchhoff, Akbal, Kovacs, Roy, Ulrich, Wald, Rotkopf, Schlemmer, and Maier-Hein]{rokuss2025lesionlocator}
Maximilian Rokuss, Yannick Kirchhoff, Seval Akbal, Balint Kovacs, Saikat Roy, Constantin Ulrich, Tassilo Wald, Lukas~T. Rotkopf, Heinz-Peter Schlemmer, and Klaus Maier-Hein.
\newblock Lesionlocator: Zero-shot universal tumor segmentation and tracking in 3d whole-body imaging, 2025.

\bibitem[Roth et~al.(2014{\natexlab{a}})Roth, Lu, Seff, Cherry, Hoffman, Wang, Liu, Turkbey, and Summers]{nih_lymph}
Holger~R. Roth, Le Lu, Ari Seff, Kevin~M. Cherry, Joanne Hoffman, Shijun Wang, Jiamin Liu, Evrim Turkbey, and Ronald~M. Summers.
\newblock \emph{A New 2.5D Representation for Lymph Node Detection Using Random Sets of Deep Convolutional Neural Network Observations}, page 520–527.
\newblock Springer International Publishing, 2014{\natexlab{a}}.

\bibitem[Roth et~al.(2014{\natexlab{b}})Roth, Lu, Seff, Cherry, Hoffman, Wang, Liu, Turkbey, and Summers]{roth2014new}
Holger~R Roth, Le Lu, Ari Seff, Kevin~M Cherry, Joanne Hoffman, Shijun Wang, Jiamin Liu, Evrim Turkbey, and Ronald~M Summers.
\newblock A new 2.5 d representation for lymph node detection using random sets of deep convolutional neural network observations.
\newblock In \emph{Medical Image Computing and Computer-Assisted Intervention--MICCAI 2014: 17th International Conference, Boston, MA, USA, September 14-18, 2014, Proceedings, Part I 17}, pages 520--527. Springer, 2014{\natexlab{b}}.

\bibitem[Roth et~al.(2015)Roth, Lu, Farag, Shin, Liu, Turkbey, and Summers]{NHpancreas}
Holger~R. Roth, Le Lu, Amal Farag, Hoo-Chang Shin, Jiamin Liu, Evrim~B. Turkbey, and Ronald~M. Summers.
\newblock Deeporgan: Multi-level deep convolutional networks for automated pancreas segmentation.
\newblock In \emph{Medical Image Computing and Computer-Assisted Intervention -- MICCAI 2015}, 2015.

\bibitem[Roth et~al.(2022)Roth, Xu, Tor-Díez, {Sanchez Jacob}, Zember, Molto, Li, Xu, Turkbey, Turkbey, Yang, Harouni, Rieke, Hu, Isensee, Tang, Yu, Sölter, Zheng, Liauchuk, Zhou, Moltz, Oliveira, Xia, Maier-Hein, Li, Husch, Zhang, Kovalev, Kang, Hering, Vilaça, Flores, Xu, Wood, and Linguraru]{covid19_challenge}
Holger~R. Roth, Ziyue Xu, Carlos Tor-Díez, Ramon {Sanchez Jacob}, Jonathan Zember, Jose Molto, Wenqi Li, Sheng Xu, Baris Turkbey, Evrim Turkbey, Dong Yang, Ahmed Harouni, Nicola Rieke, Shishuai Hu, Fabian Isensee, Claire Tang, Qinji Yu, Jan Sölter, Tong Zheng, Vitali Liauchuk, Ziqi Zhou, Jan~Hendrik Moltz, Bruno Oliveira, Yong Xia, Klaus~H. Maier-Hein, Qikai Li, Andreas Husch, Luyang Zhang, Vassili Kovalev, Li Kang, Alessa Hering, João~L. Vilaça, Mona Flores, Daguang Xu, Bradford Wood, and Marius~George Linguraru.
\newblock Rapid artificial intelligence solutions in a pandemic—the covid-19-20 lung ct lesion segmentation challenge.
\newblock \emph{Medical Image Analysis}, 82:\penalty0 102605, 2022.

\bibitem[Roy et~al.(2023)Roy, Wald, Koehler, Rokuss, Disch, Holzschuh, Zimmerer, and Maier-Hein]{roy2023sammd}
Saikat Roy, Tassilo Wald, Gregor Koehler, Maximilian~R. Rokuss, Nico Disch, Julius Holzschuh, David Zimmerer, and Klaus~H. Maier-Hein.
\newblock Sam.md: Zero-shot medical image segmentation capabilities of the segment anything model, 2023.

\bibitem[Roy et~al.(2024)Roy, Koehler, Ulrich, Baumgartner, Petersen, Isensee, Jaeger, and Maier-Hein]{mednext}
Saikat Roy, Gregor Koehler, Constantin Ulrich, Michael Baumgartner, Jens Petersen, Fabian Isensee, Paul~F. Jaeger, and Klaus Maier-Hein.
\newblock Mednext: Transformer-driven scaling of convnets for medical image segmentation, 2024.

\bibitem[Sekuboyina et~al.(2021)Sekuboyina, Husseini, Bayat, L\"{o}ffler, Liebl, Li, Tetteh, Kukačka, Payer, Štern, Urschler, Chen, Cheng, and Lessmann]{verse1}
Anjany Sekuboyina, Malek~E. Husseini, Amirhossein Bayat, Maximilian L\"{o}ffler, Hans Liebl, Hongwei Li, Giles Tetteh, Jan Kukačka, Christian Payer, Darko Štern, Martin Urschler, Maodong Chen, Dalong Cheng, and et~al. Lessmann.
\newblock Verse: A vertebrae labelling and segmentation benchmark for multi-detector ct images.
\newblock \emph{Medical Image Analysis}, 2021.

\bibitem[Shapey et~al.(2021)Shapey, Kujawa, Dorent, Wang, Bisdas, Dimitriadis, Grishchuck, Paddick, Kitchen, Bradford, Saeed, Ourselin, and Vercauteren]{Shapey2021-iz}
Jonathan Shapey, Aaron Kujawa, Reuben Dorent, Guotai Wang, Sotirios Bisdas, Alexis Dimitriadis, Diana Grishchuck, Ian Paddick, Neil Kitchen, Robert Bradford, Shakeel Saeed, Sebastien Ourselin, and Tom Vercauteren.
\newblock Segmentation of vestibular schwannoma from magnetic resonance imaging: An open annotated dataset and baseline algorithm ({vestibular-schwannoma-SEG}), 2021.

\bibitem[Simpson et~al.(2019)Simpson, Antonelli, Bakas, Bilello, Farahani, van Ginneken, Kopp-Schneider, Landman, Litjens, Menze, and et~al.]{simpson2019large}
Amber~L. Simpson, Michela Antonelli, Spyridon Bakas, Michel Bilello, Keyvan Farahani, Bram van Ginneken, Annette Kopp-Schneider, Bennett~A. Landman, Geert Litjens, Bjoern Menze, and et al.
\newblock A large annotated medical image dataset for the development and evaluation of segmentation algorithms.
\newblock \emph{arXiv:1902.09063}, 2019.

\bibitem[Simpson et~al.(2023)Simpson, Peoples, Creasy, Fichtinger, Gangai, Lasso, Keshava~Murthy, Shia, D'Angelica, and Do]{livermets}
Amber~L. Simpson, Jacob Peoples, John~M. Creasy, Gabor Fichtinger, Natalie Gangai, Andras Lasso, Krishna~Nand Keshava~Murthy, Jinru Shia, Michael~I. D'Angelica, and Richard K.~G. Do.
\newblock Preoperative ct and survival data for patients undergoing resection of colorectal liver metastases (colorectal-liver-metastases), 2023.

\bibitem[Sofroniew et~al.(2025)Sofroniew, Lambert, Bokota, Nunez-Iglesias, Sobolewski, Sweet, Gaifas, Evans, Burt, Doncila~Pop, Yamauchi, Weber~Mendonça, Buckley, Vierdag, Royer, Can~Solak, Harrington, Ahlers, Althviz~Moré, Amsalem, Anderson, Annex, Boone, Bragantini, Bussonnier, Caporal, Eglinger, Eisenbarth, Freeman, Gohlke, Gunalan, Har-Gil, Harfouche, Hilsenstein, Hutchings, Lauer, Lichtner, Liu, Liu, Lowe, Marconato, Martin, McGovern, Migas, Miller, Muñoz, Müller, Nauroth-Kreß, Palecek, Pape, Perlman, Pevey, Peña-Castellanos, Pierré, Pinto, Rodríguez-Guerra, Ross, Russell, Ryan, Selzer, Smith, Smith, Sofiiuk, Soltwedel, Stansby, Vanaret, Wadhwa, Weigert, Windhager, Winston, and Zhao]{napari}
Nicholas Sofroniew, Talley Lambert, Grzegorz Bokota, Juan Nunez-Iglesias, Peter Sobolewski, Andrew Sweet, Lorenzo Gaifas, Kira Evans, Alister Burt, Draga Doncila~Pop, Kevin Yamauchi, Melissa Weber~Mendonça, Genevieve Buckley, Wouter-Michiel Vierdag, Loic Royer, Ahmet Can~Solak, Kyle I.~S. Harrington, Jannis Ahlers, Daniel Althviz~Moré, Oren Amsalem, Ashley Anderson, Andrew Annex, Peter Boone, Jordão Bragantini, Matthias Bussonnier, Clément Caporal, Jan Eglinger, Andreas Eisenbarth, Jeremy Freeman, Christoph Gohlke, Kabilar Gunalan, Hagai Har-Gil, Mark Harfouche, Volker Hilsenstein, Katherine Hutchings, Jessy Lauer, Gregor Lichtner, Ziyang Liu, Lucy Liu, Alan Lowe, Luca Marconato, Sean Martin, Abigail McGovern, Lukasz Migas, Nadalyn Miller, Hector Muñoz, Jan-Hendrik Müller, Christopher Nauroth-Kreß, David Palecek, Constantin Pape, Eric Perlman, Kim Pevey, Gonzalo Peña-Castellanos, Andrea Pierré, David Pinto, Jaime Rodríguez-Guerra, David Ross, Craig~T. Russell, James Ryan, Gabriel Selzer, MB Smith,
  Paul Smith, Konstantin Sofiiuk, Johannes Soltwedel, David Stansby, Jules Vanaret, Pam Wadhwa, Martin Weigert, Jonas Windhager, Philip Winston, and Rubin Zhao.
\newblock napari: a multi-dimensional image viewer for python, 2025.

\bibitem[Støverud et~al.(2023)Støverud, Bouget, Pedersen, Leira, Langø, and Hofstad]{støverud2023aeropath}
Karen-Helene Støverud, David Bouget, Andre Pedersen, Håkon~Olav Leira, Thomas Langø, and Erlend~Fagertun Hofstad.
\newblock {AeroPath: An airway segmentation benchmark dataset with challenging pathology}, 2023.

\bibitem[Sudre et~al.(2024)Sudre, Van~Wijnen, Dubost, Adams, Atkinson, Barkhof, Birhanu, Bron, Camarasa, Chaturvedi, et~al.]{sudre2024valdo}
Carole~H Sudre, Kimberlin Van~Wijnen, Florian Dubost, Hieab Adams, David Atkinson, Frederik Barkhof, Mahlet~A Birhanu, Esther~E Bron, Robin Camarasa, Nish Chaturvedi, et~al.
\newblock Where is valdo? vascular lesions detection and segmentation challenge at miccai 2021.
\newblock \emph{Medical Image Analysis}, 91:\penalty0 103029, 2024.

\bibitem[Sunoqrot et~al.(2022)Sunoqrot, Saha, Hosseinzadeh, Elschot, and Huisman]{prostate158}
Mohammed R.~S. Sunoqrot, Anindo Saha, Matin Hosseinzadeh, Mattijs Elschot, and Henkjan Huisman.
\newblock Artificial intelligence for prostate mri: open datasets, available applications, and grand challenges.
\newblock \emph{European Radiology Experimental}, 6\penalty0 (1):\penalty0 35, 2022.

\bibitem[Svoboda et~al.(2009)Svoboda, Kozubek, and Stejskal]{svoboda2009generation}
David Svoboda, Michal Kozubek, and Stanislav Stejskal.
\newblock Generation of digital phantoms of cell nuclei and simulation of image formation in 3d image cytometry.
\newblock \emph{Cytometry Part A: The Journal of the International Society for Advancement of Cytometry}, 75\penalty0 (6):\penalty0 494--509, 2009.

\bibitem[Svoboda et~al.(2011)Svoboda, Homola, and Stejskal]{svoboda2011generation}
David Svoboda, Ond{\v{r}}ej Homola, and Stanislav Stejskal.
\newblock Generation of 3d digital phantoms of colon tissue.
\newblock In \emph{Image Analysis and Recognition: 8th International Conference, ICIAR 2011, Burnaby, BC, Canada, June 22-24, 2011. Proceedings, Part II 8}, pages 31--39. Springer, 2011.

\bibitem[Toulkeridou et~al.(2023)Toulkeridou, Gutierrez, Baum, Doya, and Economo]{ant_brain}
Evropi Toulkeridou, Carlos~Enrique Gutierrez, Daniel Baum, Kenji Doya, and Evan~P. Economo.
\newblock Automated segmentation of insect anatomy from micro-ct images using deep learning.
\newblock \emph{Natural Sciences}, 3\penalty0 (4):\penalty0 e20230010, 2023.

\bibitem[Ulrich et~al.(2023)Ulrich, Isensee, Wald, Zenk, Baumgartner, and Maier-Hein]{multitalent}
Constantin Ulrich, Fabian Isensee, Tassilo Wald, Maximilian Zenk, Michael Baumgartner, and Klaus~H. Maier-Hein.
\newblock Multitalent: A multi-dataset approach to medical image segmentation.
\newblock In \emph{Medical Image Computing and Computer Assisted Intervention -- MICCAI 2023}, 2023.

\bibitem[Ulrich et~al.(2024)Ulrich, Wald, Tempus, Rokuss, Jaeger, and Maier-Hein]{radioactive}
Constantin Ulrich, Tassilo Wald, Emily Tempus, Maximilian Rokuss, Paul~F. Jaeger, and Klaus Maier-Hein.
\newblock Radioactive: 3d radiological interactive segmentation benchmark, 2024.

\bibitem[Valli{\`e}res et~al.(2015)Valli{\`e}res, Freeman, Skamene, and El~Naqa]{vallieres2015radiomics}
Martin Valli{\`e}res, Carolyn~R Freeman, Sonia~R Skamene, and Issam El~Naqa.
\newblock A radiomics model from joint fdg-pet and mri texture features for the prediction of lung metastases in soft-tissue sarcomas of the extremities.
\newblock \emph{Physics in Medicine \& Biology}, 60\penalty0 (14):\penalty0 5471, 2015.

\bibitem[van~der Graaf et~al.(2023)van~der Graaf, van Hooff, Buckens, Rutten, van Susante, Kroeze, de~Kleuver, van Ginneken, and Lessmann]{spider}
Jasper~Willem van~der Graaf, Miranda~L. van Hooff, Constantinus F.~M. Buckens, Matthieu Rutten, Job L.~C. van Susante, Robert~Jan Kroeze, Marinus de Kleuver, Bram van Ginneken, and Nikolas Lessmann.
\newblock Spider - lumbar spine segmentation in mr images: a dataset and a public benchmark, 2023.

\bibitem[van Ginneken(2021)]{ski10}
Bram van Ginneken.
\newblock Ski10 papers, 2021.

\bibitem[Wahid et~al.(2024)Wahid, Dede, Naser, and Fuller]{hntsmrg2024wahid}
Kareem Wahid, Cem Dede, Mohamed Naser, and Clifton Fuller.
\newblock Training dataset for hntsmrg 2024 challenge, 2024.

\bibitem[Wang et~al.(2024)Wang, Guo, Ye, Deng, Cheng, Li, Chen, Su, Huang, Shen, Fu, Zhang, He, and Qiao]{sammed3d}
Haoyu Wang, Sizheng Guo, Jin Ye, Zhongying Deng, Junlong Cheng, Tianbin Li, Jianpin Chen, Yanzhou Su, Ziyan Huang, Yiqing Shen, Bin Fu, Shaoting Zhang, Junjun He, and Yu Qiao.
\newblock Sam-med3d: Towards general-purpose segmentation models for volumetric medical images, 2024.

\bibitem[Wasserthal et~al.(2023)Wasserthal, Breit, Meyer, Pradella, Hinck, Sauter, Heye, Boll, Cyriac, Yang, Bach, and Segeroth]{totalseg}
Jakob Wasserthal, Hanns-Christian Breit, Manfred Meyer, Maurice Pradella, daniel Hinck, Alexander~W. Sauter, Tobias Heye, Daniel~T. Boll, Joshy Cyriac, Shan Yang, Michael Bach, and Martin Segeroth.
\newblock Totalsegmentator: Robust segmentation of 104 anatomic structures in ct images.
\newblock \emph{Radiol Artif Intell.}, 2023.

\bibitem[Wei et~al.(2021)Wei, Lee, Li, Lu, Bae, Liu, Zhang, dos Santos, Lin, Uram, Wang, Arganda-Carreras, Matejek, Kasthuri, Lichtman, and Pfister]{wei2021axonem}
Donglai Wei, Kisuk Lee, Hanyu Li, Ran Lu, J.~Alexander Bae, Zequan Liu, Lifu Zhang, Márcia dos Santos, Zudi Lin, Thomas Uram, Xueying Wang, Ignacio Arganda-Carreras, Brian Matejek, Narayanan Kasthuri, Jeff Lichtman, and Hanspeter Pfister.
\newblock Axonem dataset: 3d axon instance segmentation of brain cortical regions, 2021.

\bibitem[Wolterink et~al.(2016)Wolterink, Leiner, De~Vos, Coatrieux, Kelm, Kondo, Salgado, Shahzad, Shu, Snoeren, et~al.]{wolterink2016evaluation}
Jelmer~M Wolterink, Tim Leiner, Bob~D De~Vos, Jean-Louis Coatrieux, B~Michael Kelm, Satoshi Kondo, Rodrigo~A Salgado, Rahil Shahzad, Huazhong Shu, Miranda Snoeren, et~al.
\newblock An evaluation of automatic coronary artery calcium scoring methods with cardiac ct using the orcascore framework.
\newblock \emph{Medical physics}, 43\penalty0 (5):\penalty0 2361--2373, 2016.

\bibitem[Wong et~al.(2024)Wong, Rakic, Guttag, and Dalca]{scribbleprompt}
Hallee~E. Wong, Marianne Rakic, John Guttag, and Adrian~V. Dalca.
\newblock Scribbleprompt: Fast and flexible interactive segmentation for any biomedical image, 2024.

\bibitem[Xiong et~al.(2021)Xiong, Xia, Hu, Huang, Bian, Zheng, Vesal, Ravikumar, Maier, Yang, et~al.]{xiong2021global}
Zhaohan Xiong, Qing Xia, Zhiqiang Hu, Ning Huang, Cheng Bian, Yefeng Zheng, Sulaiman Vesal, Nishant Ravikumar, Andreas Maier, Xin Yang, et~al.
\newblock A global benchmark of algorithms for segmenting the left atrium from late gadolinium-enhanced cardiac magnetic resonance imaging.
\newblock \emph{Medical image analysis}, 67:\penalty0 101832, 2021.

\bibitem[Yan et~al.(2017)Yan, Wang, Lu, and Summers]{deeplesion}
Ke Yan, Xiaosong Wang, Le Lu, and Ronald~M. Summers.
\newblock Deeplesion: Automated deep mining, categorization and detection of significant radiology image findings using large-scale clinical lesion annotations, 2017.

\bibitem[Yang et~al.(2021)Yang, Gu, Wei, Pfister, and Ni]{ripseg}
Jiancheng Yang, Shixuan Gu, Donglai Wei, Hanspeter Pfister, and Bingbing Ni.
\newblock Ribseg dataset and strong point cloud baselines for rib segmentation from ct scans.
\newblock In \emph{Medical Image Computing and Computer Assisted Intervention -- MICCAI 2021}, 2021.

\bibitem[Yang et~al.(2024)Yang, Musio, Ma, Juchler, Paetzold, Al-Maskari, Höher, Li, Hamamci, Sekuboyina, and et~al.]{topcowchallenge}
Kaiyuan Yang, Fabio Musio, Yihui Ma, Norman Juchler, Johannes~C. Paetzold, Rami Al-Maskari, Luciano Höher, Hongwei~Bran Li, Ibrahim~Ethem Hamamci, Anjany Sekuboyina, and et al.
\newblock Benchmarking the cow with the topcow challenge: Topology-aware anatomical segmentation of the circle of willis for cta and mra, 2024.

\bibitem[Yang et~al.(2023)Yang, Schmid, and Isensee]{lin_yang_2023_7413818}
Lin Yang, Otmar Schmid, and Fabian Isensee.
\newblock Lungvis1.0: Active learning ai-powered 3d imaging ecosystem for spatial profiling of lung geometry and pulmonary nanoparticle delivery, 2023.

\bibitem[Zhang and Zhuang(2022)]{zhang2022cyclemix}
Ke Zhang and Xiahai Zhuang.
\newblock Cyclemix: A holistic strategy for medical image segmentation from scribble supervision.
\newblock \emph{arXiv preprint arXiv:2203.01475}, 2022.

\bibitem[Zhang et~al.(2022)Zhang, Wu, Zhang, Qin, Zheng, Sun, Yang, and Gu]{atm22}
M Zhang, Y Wu, H Zhang, Y Qin, H Zheng, J Sun, GZ Yang, and Y Gu.
\newblock Multi-site multi-domain airway tree modeling (atm'22).
\newblock In \emph{Proc. MICCAI Challenge}, 2022.

\bibitem[Zhao et~al.(2015)Zhao, Schwartz, Kris, and Riely]{rider_lung}
Binsheng Zhao, Lawrence~H Schwartz, Mark~G Kris, and Gregory~J Riely.
\newblock Coffee-break lung ct collection with scan images reconstructed at multiple imaging parameters.
\newblock In \emph{The Cancer Imaging Archive}, 2015.

\bibitem[Zheng et~al.(2023)Zheng, Peng, Hou, Lyu, Wang, Mi, Qiao, Wan, and Yu]{zheng2023nis3d}
Wei Zheng, Cheng Peng, Zeyuan Hou, Boyu Lyu, Mengfan Wang, Xuelong Mi, Shuoxuan Qiao, Yinan Wan, and Guoqiang Yu.
\newblock Nis3d: A completely annotated benchmark for dense 3d nuclei image segmentation.
\newblock \emph{Advances in Neural Information Processing Systems}, 36:\penalty0 4741--4752, 2023.

\bibitem[Zhuang et~al.(2021)Zhuang, Xu, Luo, Chen, Ouyang, Rueckert, Campello, Lekadir, Vesal, RaviKumar, Liu, Luo, Chen, Li, Ly, Sermesant, Roth, Zhu, Wang, Ding, Wang, Yang, and Li]{MSCMRSeg}
Xiahai Zhuang, Jiahang Xu, Xinzhe Luo, Chen Chen, Cheng Ouyang, Daniel Rueckert, Victor~M. Campello, Karim Lekadir, Sulaiman Vesal, Nishant RaviKumar, Yashu Liu, Gongning Luo, Jingkun Chen, Hongwei Li, Buntheng Ly, Maxime Sermesant, Holger Roth, Wentao Zhu, Jiexiang Wang, Xinghao Ding, Xinyue Wang, Sen Yang, and Lei Li.
\newblock Cardiac segmentation on late gadolinium enhancement mri: A benchmark study from multi-sequence cardiac mr segmentation challenge, 2021.

\end{thebibliography}
}

\appendix
\setcounter{section}{0}
\renewcommand{\thesection}{A\arabic{section}} 
\setcounter{figure}{0}
\renewcommand{\thefigure}{A\arabic{figure}}   
\setcounter{table}{0}
\renewcommand{\thetable}{A\arabic{table}}     

\clearpage 
\twocolumn[
        \centering
        \Large
        \textbf{\thetitle}\\
        \vspace{0.5em}Supplementary Material \\
        \vspace{1.0em}
       ]

\section*{Overview}


This appendix provides more details and supporting materials to complement the main text. It includes implementation specifics of nnInteractive, training configurations, dataset descriptions, and additional experimental results.


\begin{itemize}
    \item \textbf{Implementation Details:} Section~\ref{appendix:training_details} outlines preprocessing, training, and data augmentation for nnInteractive, including label instance conversion, patch sampling, and user simulation.

    \item \textbf{3D Bounding Box Variant of nnInteractive:} Section~\ref{appendix:3dbboxes_suck} presents details of the 3D bounding box variant developed for benchmarking, while highlighting the practicality of 2D bounding boxes.

    \item \textbf{Dataset:} Section~\ref{appendix:dataset} lists the datasets used for training and evaluation, including in-distribution (ID) datasets and challenging out-of-distribution (OOD) test datasets for assessing generalization.

    \item \textbf{Ambiguity:} Section~\ref{appendix:ambiguity} discusses how nnInteractive dynamically resolves segmentation ambiguities with minimal user interaction.

    \item \textbf{Run Time:} Section~\ref{appendix:run_time} presents inference speed measurements from our Napari Plugin, detailing response times across different datasets, object sizes, and interaction types.

    \item \textbf{Additional Results:} Section~\ref{appendix:additional_results} provides AutoZoom’s impact on segmentation performance and qualitative comparisons demonstrating nnInteractive’s superior generalization to OOD data.
\end{itemize}
\noindent The last two images at the bottom showcase out-of-distribution data segmented using our \textbf{Napari plugin}.

\section{Implementation Details}
\label{appendix:training_details}

\paragraph{Intensity Normalization.} We perform image-level z-score normalization, meaning for each image the voxel intensities are normalized by subtracting the mean and dividing by the standard deviation.

\paragraph{Resampling.} nnInteractive processes all images at their native resolution. No resampling to harmonize the voxels spacing is performed. This is done to ensure compatibility with out of distribution datasets where the voxel spacing may lie well outside the range common for 3D medical images.

\paragraph{Model training.}
nnInteractive is trained for a total of 5000 epochs, where each epoch is defined as 250 iterations with batch size 24. We use an input patch size of 192x192x192 voxels during training. The same patch size is also used in inference. During training, data augmentation is applied, where we extend the default nnU-Net scheme: 
\begin{itemize}
    \item \textbf{Scaling.} Scaling probability is increased from 0.2 to 0.3 per sample. With probability 0.6 we disable axis synchronization and instead sample independent scaling values for each axis. We furthermore increase the scaling range from \([0.7, 1.4]\) to \([0.5, 2]\). These changes are intended to make nnInteractive less sensitive to changes in spacing (remember that we do not resample!).
    \item \textbf{Transpose}. With probability 0.5 we transpose random axes to simulate different imgage orientations
    \item \textbf{Intensity inversion}. With probability 0.1 we invert the intensity values of samples.
\end{itemize}

\noindent Other augmentations are unchanged, following nnU-Net's defaults.

\noindent The remaining training settings, for example loss function, learning rate schedule etc remain identical to nnU-Net \cite{isensee_nnu-net_2021}. The network architecture follows ResEnc L from \cite{nnunet_revisited}.

\noindent Training is performed on 8x Nvidia A100 40GB PCIe GPUs with a batch size of 3 per GPU. With an epoch time of about 200s, this results in total training time of 11-12 days. 

\paragraph{Training Labels.}
nnInteractive exclusively performs instance segmentation, meaning that the segmentation maps encode instances of objects, rather than semantic masks. Semantic segmentations are converted to instance segmentations via connected component analysis with optionally applied morphological opening or closing depending on the dataset. nnInteractive receives no information about what sematic object a mask belongs to. Segmentations for training are stored as consecutive integers with each integer encoding one instance.

\noindent We model ambiguities by combining instances where it makes sense from a task or anatomical perspective. For example in BraTS, we use the whole tumor, tumor core, edema, enhancing and necrosis labels as targets, instead of just the edema, enhancing and necrosis. 

\noindent Psudolabels via SuperVoxels: For each training case, up to 20 SuperVoxels are generated and stored separately. Since the SuperVoxels generation may fail for very large images due to insufficient VRAM, not all training cases could be augmented with additional labels. SuperVoxels were successfully generated for 64,387/64,518 volumes. For some cases insufficient high confidence masks were generated resulting in a lower number of generated SuperVoxels. The average number of generated SuperVoxels per case is 17.46. 

\paragraph{Patch sampling.}
During training patches are sampled by first randomly selecting a training case, then an object within that case and finally a pixel within that object. Sampling probabilities of training cases are determined using a mix of heuristics and manual adjustments. Three terms determine the sampling probability for each case: 
\begin{itemize}
    \item \textbf{What dataset it belongs to.} Each dataset is allocated a sampling budget equal to the square root of the number of training cases, giving datasets with more cases more representation in the training. This budget is distributed across all training cases, thus giving cases from large datasets overall a lower sampling probability than cases from smaller datasets
    \item \textbf{How many object each sample contains.} With the sampling probability for each case being adjusted proportional to the square root of the number of objects within them. 
    \item \textbf{Manual dataset weights.} Datasets are up- or down weighted based on subjective perception of their usefulness for nnInteractive training. Large uniform datasets with repetitive targets (for example CAP~\cite{kadish2009rationale} or FiloData3D~\cite{ljosa2012annotated}) receive lower sampling probabilities while interesting, small dataset receive higher weights (for example SegThy 1~\cite{kronke2022tracked}).
\end{itemize}

\noindent Once a training case was selected, a random object from that case is determined. Pseudolabels are sampled with probability 0.2. From the generated SuperVoxels one object is selected at random. For the remaining p=0.8 we sample real training labels. Hereby, sampling probabilities for available objects are distributed uniformly for most datasets, except for some where important classes could potentially be under-represented. For example, in Task3 from the MSD \cite{antonelli2021medical} we sample the liver with and without tumors with fixed probability 0.25 each, while distributing the remaining 0.5 across all present tumor instances.

\noindent Given a selected object a training patch is selected by picking a center-biased random voxel of that class and constructing the 192x192x192 patch around it.

\noindent Patches are first fed through the standard nnU-Net dataloading pipeline. At the end, an initial interaction is simulated and a user simulating agent is randomly selected.

\paragraph{User simulation and followup interactions.}
In parallel to the standard dataloading infrastructure there is a process pool for handling the user simulation. It receives prediction from the network along with the corresponding patch and previous interactions and applies the selected user agent which in turn selected an interaction to be applied for refinement (see Fig. \ref{fig:nnInteractive_method_overview}). 

\noindent During training, nnInteractive first checks whether a follow-up interaction is ready for processing. If it is it fetches that from the user simulation process. If not it proceeds to draw a new training sample from the standard data augmentation pipeline. After each prediction, the current batch is given with a probability \(p\) to the user simulation for further refinement. The followup interaction probability \(p\) is initialized with 0.3 at epoch 0 and linearly increased to 0.75 during the training, unlocking more refinement steps as the model learns. 

\section{3D Bounding Box Variant of nnInteractive}
\label{appendix:3dbboxes_suck}
The 3D bounding box version of nnInteractive was developed solely for benchmarking against existing 3D interactive segmentation models that rely on 3D bounding boxes.\\

\noindent In this version, 2D bounding boxes and lasso interactions were disabled, replaced entirely by 3D bounding boxes. These were simulated during training using the same methodology as 2D bounding boxes. Due to computational constraints, the model was trained for 2,000 epochs, instead of the normal 5,000 epochs, which would likely yield even better performance. Despite this, the 2K-epoch model effectively demonstrated that nnInteractive’s design principles extend well to 3D bounding boxes, outperforming all baselines by a large margin.\\

\noindent However,\textit{ we do not consider 3D bounding boxes a practical choice for interactive segmentation}. They are cumbersome for users to create and, in our experiments, offered no measurable advantage over 2D bounding boxes. On the contrary, they often led to false positive predictions due to the excess empty space within the bounding volume, particularly for objects oriented diagonally.\\  

\noindent For these reasons, 3D bounding box support was intentionally excluded from the final version of nnInteractive. Beyond their inefficiency, they negatively impacted 2D bounding box and lasso performance when all interactions shared the same input channel. Due to the disparity in the number of positive input voxels, incorporating 3D bounding boxes would have required separate input channels, increasing the total input channels from 7 to 9, adding unnecessary complexity without clear benefits.

\section{Dataset}\label{appendix:dataset}
Table \ref{tab:training_datasets} provides an overview of all datasets used for training. Test and OOD datasets are presented in Table \ref{tab:datasets_test}.

\thispagestyle{empty}
\begin{table*}[h!]
\centering
\scriptsize
\caption{Overview of the 120 datasets used for model training, covering names, image counts, modalities, targets, and access links.}
\label{tab:training_datasets}
\resizebox{0.93\textwidth}{!}{ 
\begin{tabular}{lccrr}
\toprule
Name & Images & Modality & Target & Link \\
\midrule
Decathlon Task 2 \cite{antonelli2021medical,simpson2019large}& 20 & MRI & Heart & \url{http://medicaldecathlon.com} \\
Decathlon Task 3 \cite{antonelli2021medical,simpson2019large}& 131 & CT & Liver, L. Tumor & \url{http://medicaldecathlon.com} \\
Decathlon Task 4 \cite{antonelli2021medical,simpson2019large}& 208 & MRI & Hippocampus & \url{http://medicaldecathlon.com} \\
Decathlon Task 5 \cite{antonelli2021medical,simpson2019large}& 32 & MRI & Prostate & \url{http://medicaldecathlon.com} \\
Decathlon Task 6 \cite{antonelli2021medical,simpson2019large}& 63 & CT & Lung Lesion & \url{http://medicaldecathlon.com} \\
Decathlon Task 7 \cite{antonelli2021medical,simpson2019large}& 281 & CT & Pancreas, P. Tumor & \url{http://medicaldecathlon.com} \\
Decathlon Task 8 \cite{antonelli2021medical,simpson2019large}& 303 & CT & Hepatic Vessel, H. Tumor & \url{http://medicaldecathlon.com} \\
Decathlon Task 9 \cite{antonelli2021medical,simpson2019large}& 41 & CT & Spleen & \url{http://medicaldecathlon.com} \\
Decathlon Task 10 \cite{antonelli2021medical,simpson2019large}& 126 & CT & Colon Tumor & \url{http://medicaldecathlon.com} \\
ISLES2015 \cite{isles2015}& 28 & MRI & Stroke Lesion & \url{http://www.isles-challenge.org/ISLES2015} \\
BTCV \cite{BTCV}& 30 & CT & 13 Abdominal Organs & \url{https://www.synapse.org/Synapse:syn3193805/wiki/89480} \\
LIDC \cite{lidc}& 1010 & CT & Lung Lesion & \url{https://www.cancerimagingarchive.net/collection/lidc-idri} \\
Promise12 \cite{Litjens2014}& 50 & MRI & Prostate & \url{https://zenodo.org/records/8026660} \\
ACDC \cite{Bernard2018}& 200 & MRI & RV Cavity, LV Myocardium, LV Cavity & \url{https://www.creatis.insa-lyon.fr/Challenge/acdc/databases.html} \\
ISBILesion2015 \cite{Carass2017}& 42 & MRI & MS Lesion & \url{https://iacl.ece.jhu.edu/index.php/MSChallenge} \\
CHAOS \cite{CHAOSdata2019}& 60 & MRI & Liver, Kidney (L and R), Spleen & \url{https://zenodo.org/records/3431873} \\
BTCV 2 \cite{BTCV2}& 63 & CT & 9 Abdominal Organs & \url{https://zenodo.org/records/1169361} \\
StructSeg Task1 \cite{structseg} & 50 & CT & 22 OAR Head \& Neck & \url{https://structseg2019.grand-challenge.org} \\
StructSeg Task2 \cite{structseg} & 50 & CT & Nasopharynx Cancer & \url{https://structseg2019.grand-challenge.org} \\
StructSeg Task3 \cite{structseg} & 50 & CT & 6 OAR Lung & \url{https://structseg2019.grand-challenge.org} \\
StructSeg Task4 \cite{structseg} & 50 & CT & Lung Cancer & \url{https://structseg2019.grand-challenge.org} \\
SegTHOR \cite{lambert2019segthor} & 40 & CT & Heart, Aorta, Trachea, Esophagus & \url{https://competitions.codalab.org/competitions/21145} \\
NIH-Pan \cite{clark_cancer_2013,NHpancreas}  & 82 & CT & Pancreas & \url{https://wiki.cancerimagingarchive.net/display/Public/Pancreas-CT} \\
VerSe2020 \cite{verse1,verse2,verse3} & 113 & CT & 28 Vertebrae & \url{https://github.com/anjany/verse} \\
M\&Ms \cite{Campello2021,10103611}& 300 & MRI & Left \& Right Ventricle, Myocardium & \url{https://www.ub.edu/mnms} \\
ProstateX \cite{prostatex} & 140 & MRI & Prostate Lesion & \url{https://www.aapm.org/GrandChallenge/PROSTATEx-2} \\
RibSeg \cite{ripseg}& 370 & CT & Ribs & \url{https://github.com/M3DV/RibSeg?tab=readme-ov-file} \\
BrainMetShare \cite{Grovik2020-un} & 84 & MRI & Brain Metastases & \url{https://aimi.stanford.edu/brainmetshare} \\
CrossModa22 \cite{Shapey2021-iz}& 168 & MRI & Vestibular Schwannoma, Cochlea & \url{https://crossmoda2022.grand-challenge.org} \\
Atlas22 \cite{Liew2018-wy}& 524 & MRI & Stroke Lesion & \url{https://atlas.grand-challenge.org} \\
KiTs23 \cite{heller2023kits21} & 489 & CT & Kidneys, K. Tumor, Cysts & \url{https://kits-challenge.org/kits23} \\
AutoPet2 \cite{Gatidis2022-ms}& 1014 & PET,CT & Lesions & \url{https://autopet-ii.grand-challenge.org} \\
AMOS \cite{ji2022amos}& 360 & CT,MRI & 15 Abdominal Organs & \url{https://amos22.grand-challenge.org} \\
BraTS24 \cite{Karargyris2023,bratsgli1,bratsgli2,bratsgli3} & 1251 & MRI & Glioblastoma & \url{https://www.synapse.org/Synapse:syn51156910/wiki/621282} \\
AbdomenAtlas1.1Mini \cite{li2024well,qu2023abdomenatlas}& 5195 & CT & 8 Abdominal Organs & \url{https://huggingface.co/datasets/AbdomenAtlas/_AbdomenAtlas1.1Mini} \\
TotalSegmentatorV2 \cite{totalseg}& 1180 & CT & 117 Classes of Whole Body & \url{https://github.com/wasserth/TotalSegmentator} \\
Hecktor2022 \cite{Andrearczyk2023-ii}& 524 & PET,CT & Head and Neck Tumor & \url{https://hecktor.grand-challenge.org} \\
FLARE \cite{FLARE22} & 50 & CT & 13 Abdominal Organs & \url{https://flare22.grand-challenge.org} \\
SegA \cite{Radl2022-fz,Jin2021-jo,Pepe2020-fh}& 56 & CT & Aorta & \url{https://multicenteraorta.grand-challenge.org/data} \\
WORD \cite{luo2022word,liao2023comprehensive}& 120 & CT & 16 Abdominal Organs & \url{https://github.com/HiLab-git/WORD} \\
AbdomenCT1K \cite{Ma-2021-AbdomenCT-1K} & 996 & CT & Liver, Kidney, Spleen, Pancreas & \url{https://github.com/JunMa11/AbdomenCT-1K} \\
DAP-ATLAS \cite{jaus2023towards}& 533 & CT & 142 Classes of Whole Body & \url{https://github.com/alexanderjaus/AtlasDataset} \\
CTORG \cite{Rister2019tm} & 140 & CT & Lung, Brain, Bones, Liver, Kidney, Bladder & \url{https://www.cancerimagingarchive.net/collection/ct-org} \\
TopCow \cite{topcowchallenge}& 200 & CT,MRI & Vessel Components of CoW & \url{https://topcow23.grand-challenge.org} \\
AortaSeg24~\cite{IMRAN2024102470} & 50 & CT & Aorta & \url{https://aortaseg24.grand-challenge.org} \\
Duke Liver~\cite{macdonald_2020_7774566} & 310 & MRI & Liver  & \url{https://zenodo.org/records/7774566} \\
Aero Path~\cite{støverud2023aeropath} & 27 & CT & Lungs, Airways & \url{https://github.com/raidionics/AeroPath} \\
AxonEM~\cite{wei2021axonem} & 18 & El. Microscopy & Axon Instances & \url{https://axonem.grand-challenge.org} \\
MitoEM~\cite{wei2021axonem} & 4 & El. Microscopy & Mitochondria Instances & \url{https://mitoem.grand-challenge.org} \\
NucMM~\cite{Lin_2021} & 62 & El. Microscopy & Neuronal Nuclei  & \url{https://nucmm.grand-challenge.org} \\
LungVis1.0~\cite{lin_yang_2023_7413818} & 22 & Fl. Microscopy & Airway & \url{https://zenodo.org/records/7413818} \\
BBBC024 HL60 Cell line~\cite{svoboda2009generation} & 240 & Fl. Microscopy & Cell Nuclei & \url{https://bbbc.broadinstitute.org/BBBC024} \\
BBBC027 Colon Tissue~\cite{svoboda2011generation} & 60 & Fl. Microscopy & Colon Tissue & \url{https://bbbc.broadinstitute.org/BBBC027} \\
BBBC032 MouseEmbryoBlastocyst~\cite{ljosa2012annotated} & 1 & Fl. Microscopy & Blastocyst Cells & \url{https://bbbc.broadinstitute.org/BBBC032} \\
BBBC033 MouseTrophoblast~\cite{ljosa2012annotated} & 1 & Microscopy & Trophoblast & \url{https://bbbc.broadinstitute.org/BBBC033} \\
BBBC034 PluripStemCells~\cite{ljosa2012annotated} & 1 & Microscopy & Stem Cells & \url{https://bbbc.broadinstitute.org/BBBC034} \\
BBBC046 FiloData3D~\cite{ljosa2012annotated} & 5400 & Fl. Microscopy & Lung Cancer Cells & \url{https://bbbc.broadinstitute.org/BBBC046} \\
BBBC050 MouseEmbryoNuclei~\cite{ljosa2012annotated} & 165 & Fl. Microscopy & Mouse Embryonic Cells & \url{https://bbbc.broadinstitute.org/BBBC050} \\
CAMUS~\cite{leclerc2019deep} & 1000 & US & Endocardium,  Epicardium, Atrium & \url{https://www.creatis.insa-lyon.fr/Challenge/camus/index.html} \\
CETUS~\cite{bernard2015standardized} & 90 & US & LV Lumen & \url{https://www.creatis.insa-lyon.fr/Challenge/CETUS/databases.html} \\
EPFL Mito~\cite{lucchi2011supervoxel} & 1 & El. Microscopy & Mitochondria & \url{https://www.epfl.ch/labs/cvlab/data/data-em} \\
FETA~\cite{payette2021automatic} & 120 & MRI & Brain Regions & \url{https://fetachallenge.github.io/pages/Data_description} \\
Drosophila~\cite{gerhard2013segmented} & 1 & El. Microscopy & Mitochondria, Synapses & \url{https://github.com/unidesigner/groundtruth-drosophila-vnc} \\
Leg3DUS~\cite{duque2024ultrasound} & 44 & US & Lower-Limb Leg  & \url{https://www.cs.cit.tum.de/camp/publications/leg-3d-us-dataset} \\
LGGMRISeg~\cite{buda2019association} & 110 & MRI & Tumor & \url{https://www.kaggle.com/datasets/mateuszbuda/lgg-mri-segmentation/data} \\
M-CRIB~\cite{alexander2019desikan} & 10 & MRI & Neonatal Brain Atlas & \url{https://osf.io/4vthr} \\
ParticleSeg3D~\cite{gotkowski2024particleseg3d} & 54 & MicroCT & Mineral Samples & \url{https://syncandshare.desy.de/index.php/s/wjiDQ49KangiPj5} \\
RESECT~\cite{behboodi2024open} & 69 & US & Cerebral Tumor & \url{https://osf.io/jv8bk} \\
CAP~\cite{kadish2009rationale} & 1637 & MRI & Left Ventricle & \url{https://www.cardiacatlas.org/lv-segmentation-challenge} \\
AtriaSeg2018~\cite{xiong2021global} & 100 & MRI & Left Atrium & \url{https://www.cardiacatlas.org/atriaseg2018-challenge/atria-seg-data} \\
NIS3D~\cite{zheng2023nis3d} & 6 & Fl. Microscopy & Cell Nuclei & \url{https://zenodo.org/records/11456029} \\
SegThy 1~\cite{kronke2022tracked} & 14 & MRI & Thyroid, Carotid, Jugular Vein & \url{https://www.cs.cit.tum.de/camp/publications/segthy-dataset} \\
SegThy 2~\cite{kronke2022tracked} & 32 & US & Thyroid, Carotid, Jugular Vein & \url{https://www.cs.cit.tum.de/camp/publications/segthy-dataset} \\
Fluo C3DH A549~\cite{mavska2023cell} & 90 & Fl. Microscopy & Cell & \url{https://celltrackingchallenge.net/3d-datasets} \\
Fluo N3DH~\cite{mavska2023cell} & 230 & Fl. Microscopy & Cell, Border & \url{https://celltrackingchallenge.net/3d-datasets} \\
Spine-Mets~\cite{pieper2024spinemets} & 55 & CT & Vertebra & \url{https://www.cancerimagingarchive.net/collection/spine-mets-ct-seg} \\
WMHSegChallenge~\cite{kuijf2019standardized} & 60 & MRI & White Matter Hyperintensities & \url{https://dataverse.nl/dataset.xhtml?persistentId=doi:10.34894/AECRSD} \\
NCI-ISBI~\cite{bloch2015nci} & 59 & MRI & Prostate & \url{https://www.cancerimagingarchive.net/analysis-result/isbi-mr-prostate-2013} \\
OASIS~\cite{marcus2007open} & 436 & MRI & Brain Regions & \url{https://sites.wustl.edu/oasisbrains/home/oasis-1} \\
MediaLymph~\cite{bouget2021mediastinal} & 15 & CT & Mediastinal Lymph Nodes & \url{https://github.com/dbouget/ct_mediastinal_structures_segmentation} \\
MediaStruct~\cite{bouget2019semantic} & 15 & CT & Mediastinal Structures & \url{https://github.com/dbouget/ct_mediastinal_structures_segmentation} \\
CT Lymph Nodes~\cite{roth2014new} & 175 & CT & Lymph Nodes & \url{https://www.cancerimagingarchive.net/collection/ct-lymph-nodes} \\
MAMA MIA~\cite{mama_mia} & 1506 & MRI & Breast Lesions & \url{https://www.synapse.org/Synapse:syn60868042/wiki/628716} \\
ATM2022~\cite{atm22} & 300 & CT & Airway Tree & \url{https://atm22.grand-challenge.org} \\
Pediatric CT SEG~\cite{jordan2022pediatric} & 353 & CT & Organs & \url{https://doi.org/10.7937/TCIA.X0H0-1706} \\
Atlas Bourgogne~\cite{quinton2023tumour} & 60 & MRI & Liver, Tumor & \url{https://atlas-challenge.u-bourgogne.fr } \\
CC Tumor Heterogeneity~\cite{Mayr2023} & 63 & MRI & Cervix, Tumor & \url{https://www.cancerimagingarchive.net/collection/cc-tumor-heterogeneity} \\
CURVAS~\cite{RieraMarin2024} & 60 & CT & Pancreas, Kidney, Liver & \url{https://curvas.grand-challenge.org/curvas-dataset} \\
Emidec~\cite{lalande2020emidec} & 100 & MRI & Heart Structures & \url{https://emidec.com} \\
HVSMR-2.0~\cite{pace2024hvsmr} & 60 & MRI & Heart, Vessel & \url{https://segchd.csail.mit.edu} \\
Kipa22~\cite{he2021meta} & 70 & CT & Kidney, Vessel, Tumor & \url{https://kipa22.grand-challenge.org} \\
MrBrains18~\cite{kuijf2024mr} & 30 & MRI & Brain Structures & \url{https://mrbrains18.isi.uu.nl/index.html} \\
OrCaScore~\cite{wolterink2016evaluation} & 32 & CT & Calcifications & \url{https://orcascore.grand-challenge.org} \\
Parse22~\cite{luo2023efficient} & 100 & CT & Pulmonary Artery & \url{https://parse2022.grand-challenge.org/Parse2022} \\
PDDCA~\cite{raudaschl2017evaluation} & 47 & CT & Head and Neck Structures & \url{https://www.imagenglab.com/newsite/pddca} \\
ProstateEdgeCases~\cite{kanwar2023stress} & 131 & CT & Bladder, Prostate, Rectum & \url{https://www.cancerimagingarchive.net/collection/prostate-anatomical-edge-cases} \\
SKI10~\cite{ski10} & 100 & MRI & Cartilage, Bone  & \url{https://ski10.grand-challenge.org} \\
Soft Tissue Sarcoma~\cite{vallieres2015radiomics} & 102 & MRI & Edema, Tumor & \url{https://www.cancerimagingarchive.net/collection/soft-tissue-sarcoma} \\
Spider~\cite{spider} & 447 & MRI & Lumbar Spine & \url{https://zenodo.org/records/10159290} \\
VALDO Task 2~\cite{sudre2024valdo} & 72 & MRI & Cerebral Microbleed & \url{https://valdo.grand-challenge.org/Task2} \\
ToothFairy 2~\cite{toothfairy2} & 480 & CT & Dental Structures & \url{https://ditto.ing.unimore.it/toothfairy2} \\
UPENN-GBM~\cite{upenn-gbm} & 147 & MRI & Edema, Tumor & \url{https://www.cancerimagingarchive.net/collection/upenn-gbm} \\
ReMIND~\cite{remind} & 213 & MRI & Brain Resection & \url{https://www.cancerimagingarchive.net/collection/remind} \\
Prostate158~\cite{prostate158} & 188 & MRI & Gland, Tumor & \url{https://zenodo.org/records/6481141} \\
TotalSegmentator MRI~\cite{totalseg_mri} & 298 & MRI & Organs & \url{https://zenodo.org/records/11367005} \\
Instance2022~\cite{instance22} & 100 & CT & Intracranial Hemorrhage  & \url{https://instance.grand-challenge.org} \\
LAPD Mouse~\cite{lapd_mouse} & 34 & Fl. Microscopy & Airway & \url{https://cebs-ext.niehs.nih.gov/cahs/report/lapd/web-download-links} \\
Deep Lesion~\cite{deeplesion} & 1093 & CT & Multiple Types of Lesions & \url{https://nihcc.app.box.com/v/DeepLesion} \\ 
COVID-19 CT Lung~\cite{covid19_junma} & 10 & CT & COVID -19 & \url{https://zenodo.org/records/3757476} \\ 
LNDb~\cite{lndb} & 229 & CT & Lymph Nodes & \url{https://lndb.grand-challenge.org} \\ 
NIH Lymph~\cite{nih_lymph} & 176 & CT & Lymph Nodes & \url{https://www.cancerimagingarchive.net/collection/ct-lymph-nodes} \\ 
NSCLC Pleural Effusion~\cite{kiser2020plethora} & 78 & CT & Pleural Effusion & \url{https://www.cancerimagingarchive.net/analysis-result/plethora} \\ 
NSCLC Radiomics~\cite{NSCLC-Radiomics} & 503 & CT & Lung Lesions & \url{https://www.cancerimagingarchive.net/collection/nsclc-radiomics} \\ 
COVID-19-20~\cite{covid19_challenge} & 199 & CT & COVID-19 & \url{https://covid-segmentation.grand-challenge.org/COVID-19-20} \\ 
\bottomrule
\end{tabular}
}
\end{table*}

\begin{table*}
\caption{\textbf{Overview of held-out test datasets} spanning diverse anatomical structures, pathologies, and imaging modalities. These datasets introduce significant domain shifts in resolution, contrast, target structures, and anatomical scale, serving as a rigorous benchmark for model robustness. We use the filtered datasets from the RadioActive benchmark~\cite{radioactive} along with four additional out-of-distribution (OOD) datasets, visually inspected before benchmarking. None were part of the training data, except for SegVol, which included HanSeg.}
\label{tab:datasets_test}
\resizebox{0.96\linewidth}{!}{
\begin{tabular}[]{c lllr}
\toprule
& \textbf{Dataset}            & \textbf{Modality} & \textbf{Targets}         & \textbf{Images}       \\
\midrule
\multirow{10}{*}{\shortstack{RadioActive\\Benchmark\\Datasets~\cite{radioactive}}}  & MS Lesion \citep{muslim2022ms}            & MRI (T2 Flair)                & MS Lesions     & 60 \\
& HanSeg   \citep{podobnik2023han}   & MRI (T1)      &  30 Organs at Risk  & 42               \\
& HNTSRMFG    \citep{hntsmrg2024wahid}        &    MRI (T2)          & Oropharyngeal Cancer and Metastatic Lymph Nodes  & 135             \\
& RiderLung \citep{rider_lung} & CT & Lung Lesions & 58 \\
& LNQ    \citep{lnq2023challenge}                 & CT                & Mediastinal Lymph Nodes & 513\\
& LiverMets    \citep{livermets}                 & CT                & Liver Metastases & 171 \\
& Adrenal ACC    \citep{adrenalacc}                 & CT                &  Adrenal Tumors & 53\\
& HCC Tace        \citep{hcctace}             & CT                &  Liver and Liver Tumors & 65\\
& Pengwin    \citep{liu2023pelvic}           & CT                &   Bone Fragments          &100      \\
& SegRap \citep{luo2023segrap2023}& CT & 45 Organs at Risk & 30 \\
\midrule
\multirow{4}{*}{\shortstack{Additional\\OOD\\Datasets}} & MouseTumor \citep{mouseCTtumor} & MicroCT & Subcutaneous Tumors in Mice & 452 \\
& InsectAnatomy \citep{ant_brain} & MicroCT & Insect Brain & 84 \\
& ACRIN H\&N \citep{Kinahan2019acrin} & PET & Head and Neck Tumors & 67 \\
& Stanford Knee \citep{stanford_knee_mri} & MRI & Patellar, Femoral, Tibial Cartilages and Meniscus & 155 \\
\bottomrule
\end{tabular}}
\end{table*}

\section{Ambiguity}\label{appendix:ambiguity}
\begin{figure*}[!t]
  \begin{subfigure}[t]{0.32\textwidth}
    \includegraphics[width=\linewidth]{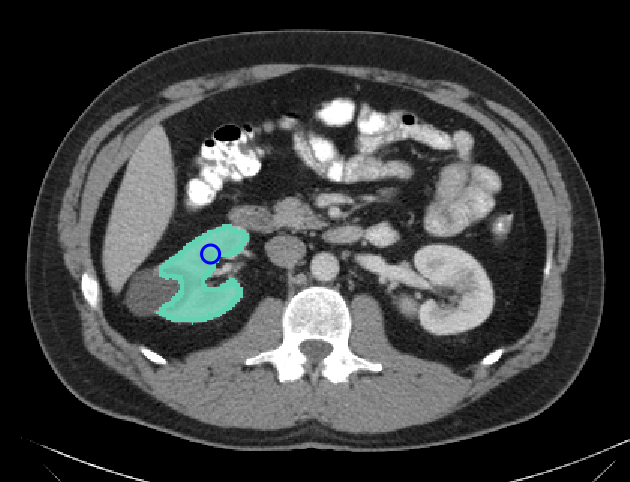}
    \caption{Target: Kidney without tumor}
  \end{subfigure}
  \hfill
  \begin{subfigure}[t]{0.32\textwidth}
    \includegraphics[width=\linewidth]{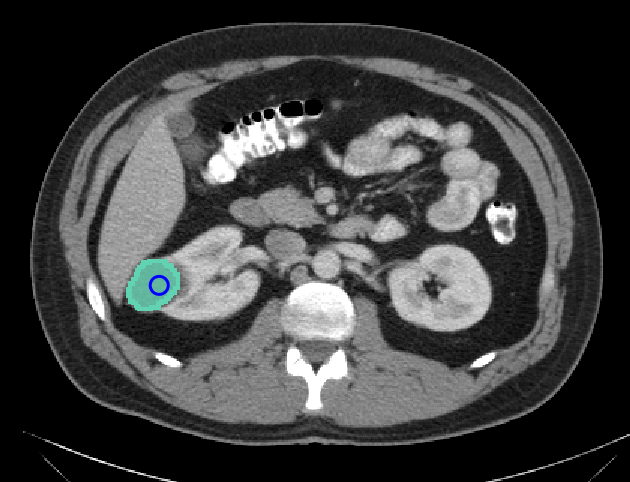}
    \caption{Target: Kidney tumor}
  \end{subfigure}
  \hfill
  \begin{subfigure}[t]{0.32\textwidth}
    \includegraphics[width=\linewidth]{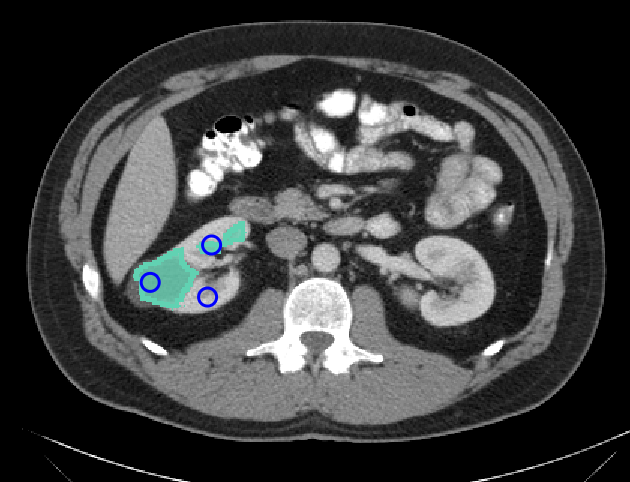}
    \caption{Target: Kidney with tumor}
  \end{subfigure}
  \caption{Vista 3D \cite{vista3d} cannot resolve ambiguities. When prompted to segment the kidney or the kidney tumor (a and b), Vista3D yields plausible results. However, Vista3D is unable to segment the kidney including the tumor (c), as this setting contradicts its training labels. Results generated using Nvidias online demo (\url{https://build.nvidia.com/nvidia/vista-3d}) and their provided \textit{Abdomen CT} example image.}
  \label{fig:vista_fail_kidney_and_tumor}
\end{figure*}

\begin{figure*}[!t]
  \begin{subfigure}[t]{0.32\textwidth}
    \includegraphics[width=\linewidth]{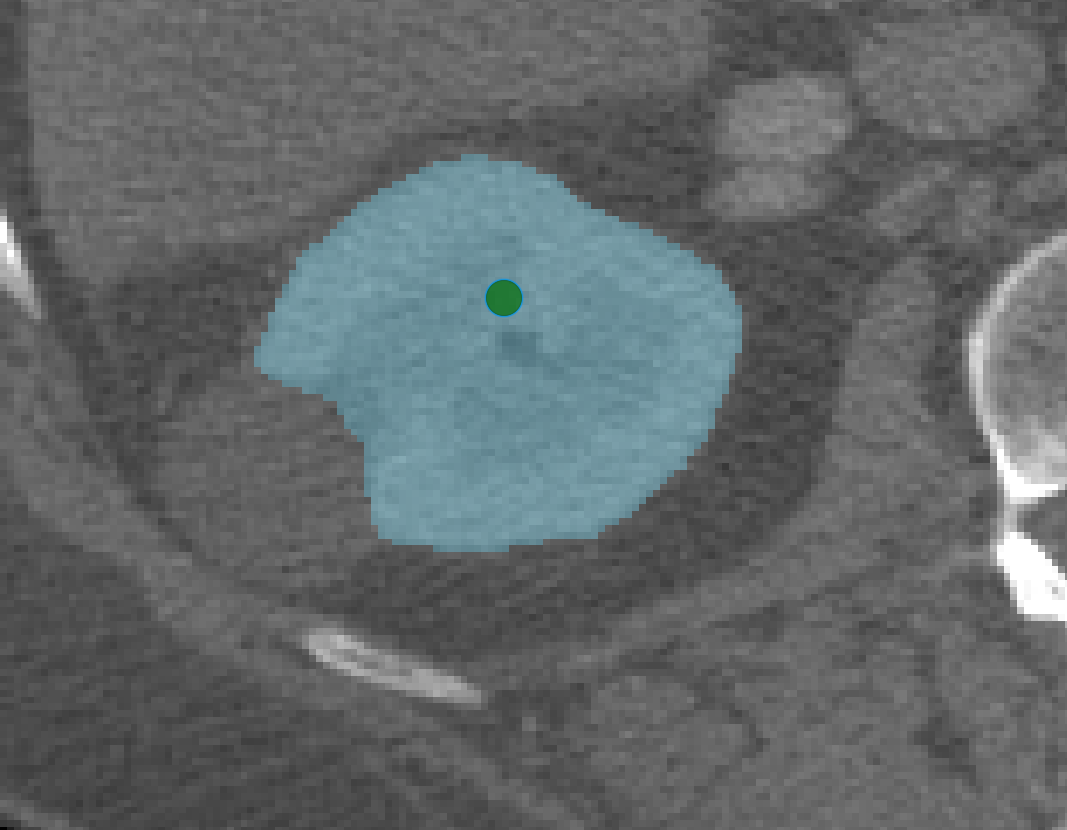}
    \caption{Target: Kidney without tumor}
  \end{subfigure}
  \hfill
  \begin{subfigure}[t]{0.32\textwidth}
    \includegraphics[width=\linewidth]{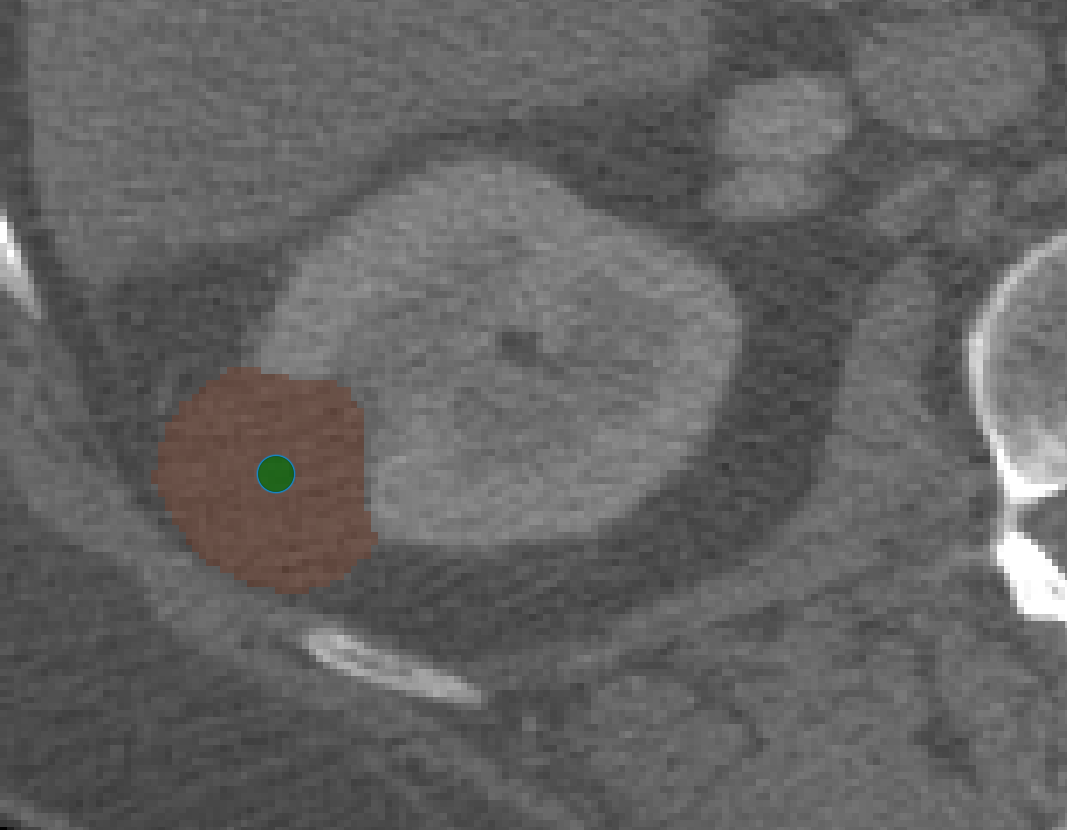}
    \caption{Target: Kidney tumor}
  \end{subfigure}
  \hfill
  \begin{subfigure}[t]{0.32\textwidth}
    \includegraphics[width=\linewidth]{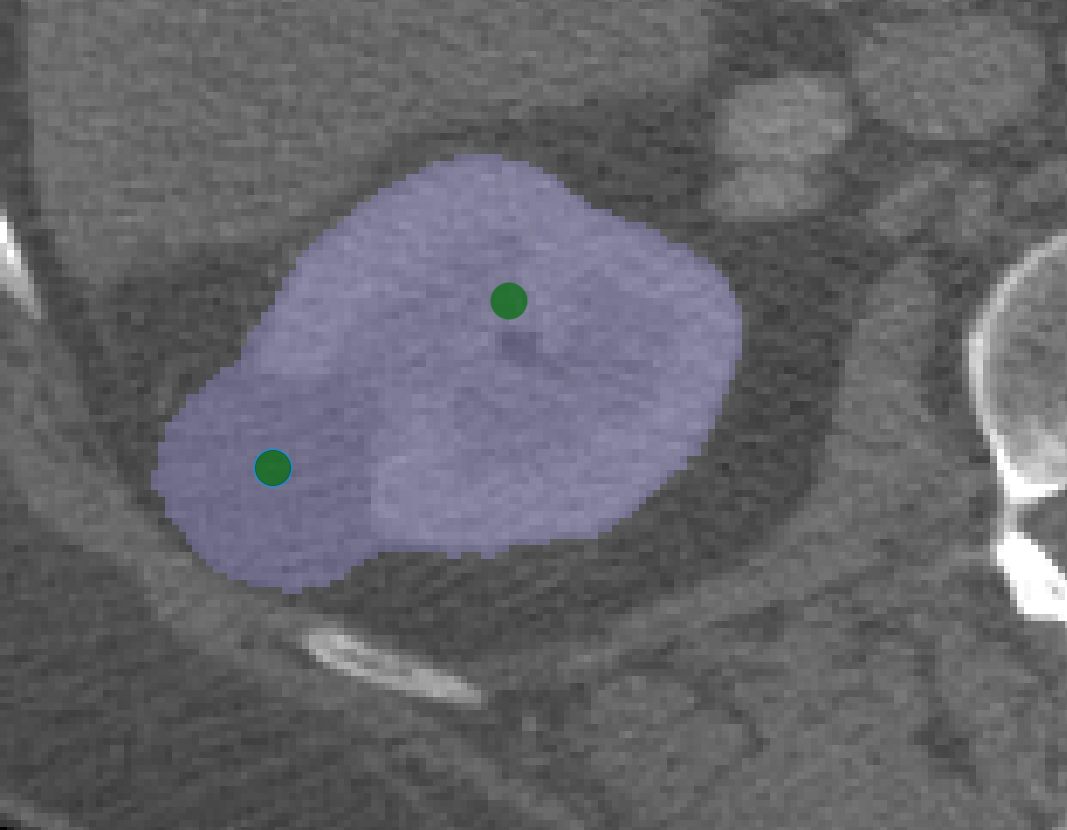}
    \caption{Target: Kidney including tumor}
  \end{subfigure}

  \begin{subfigure}[t]{0.32\textwidth}
    \includegraphics[width=\linewidth]{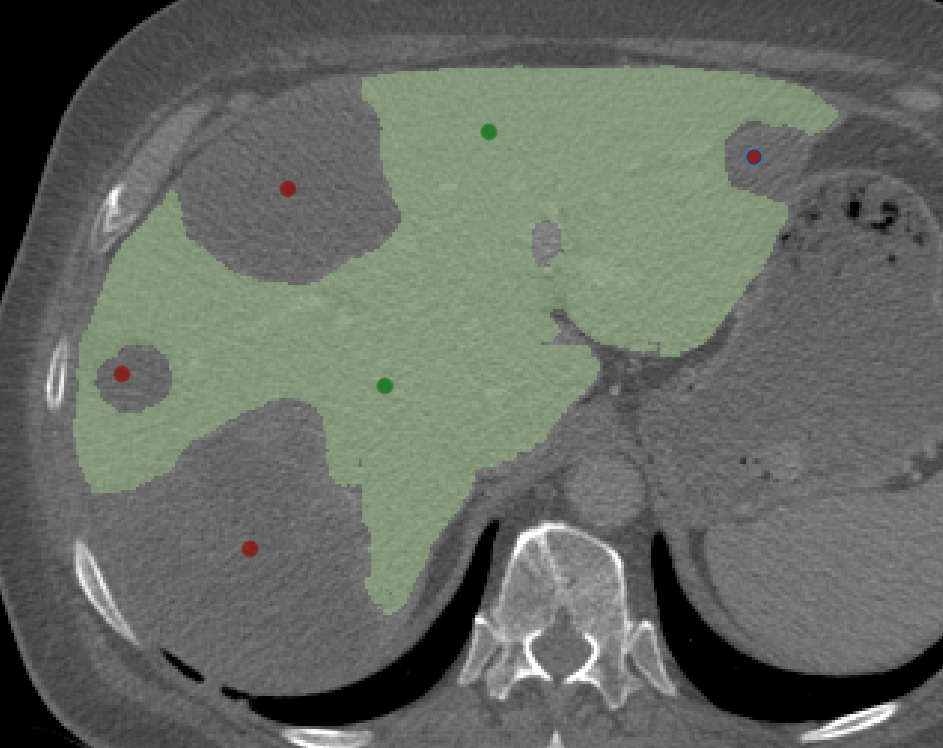}
    \caption{Target: Liver without tumors}
  \end{subfigure}
  \hfill
  \begin{subfigure}[t]{0.32\textwidth}
    \includegraphics[width=\linewidth]{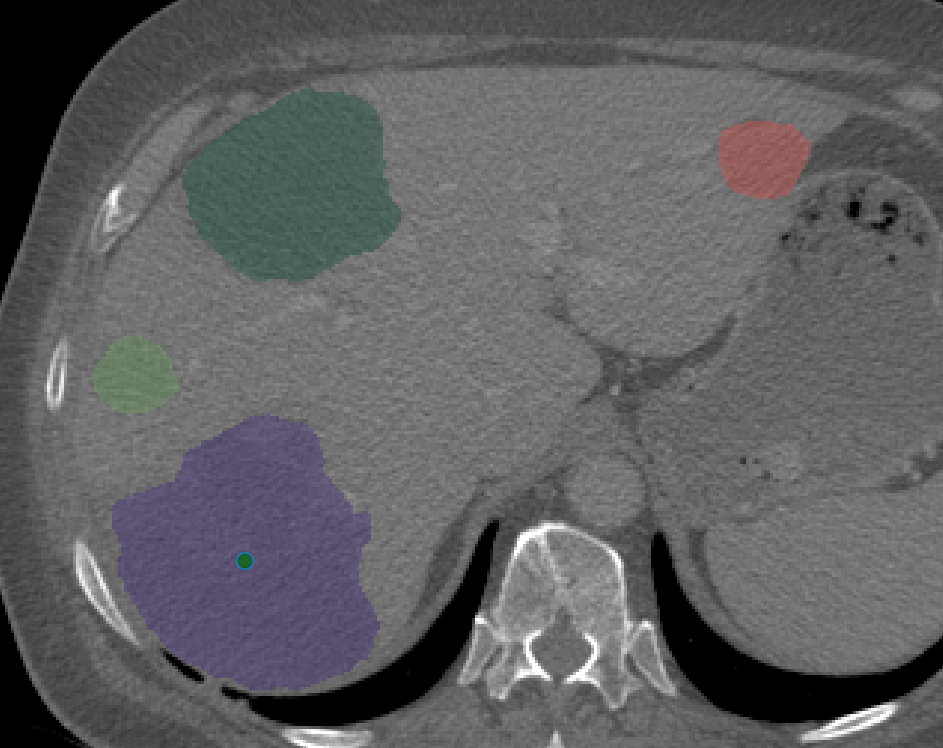}
    \caption{Target: Tumors (instances)}
  \end{subfigure}
  \hfill
  \begin{subfigure}[t]{0.32\textwidth}
    \includegraphics[width=\linewidth]{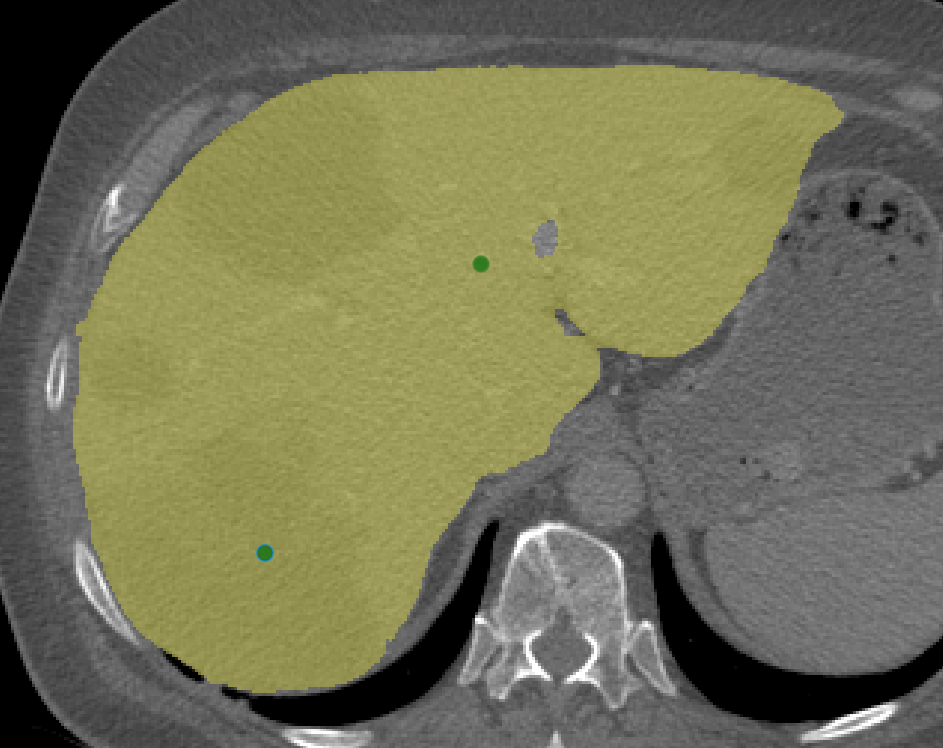}
    \caption{Target: Liver including tumors}
  \end{subfigure}
  \caption{nnInteractive successfully resolves ambiguities. Image is \textit{liver\_190} from the MSD Task 3 test set.}
  \label{fig:nninteractive_ambiguities_win}
\end{figure*}

nnInteractive dynamically adapts to user input and is able to efficiently resolve ambiguities with minimal interaction Fig. \ref{fig:nninteractive_ambiguities_win}. This is in stark contrast to competing methods that are overfitted to the training labels and lack specific ambiguity enabling training schemes (Fig. \ref{fig:vista_fail_kidney_and_tumor}).

\section{Run Time}\label{appendix:run_time}

\label{appendix:runtime}
\begin{table*}[h!]
\caption{\textbf{Inference run times.} Measured using our Napari Plugin on a system with an Nvidia RTX 4090, AMD Ryzen 5800X3D and 32GB RAM. Multiple numbers per box indicate several refinement steps by the user. Inference time is measured as wall time from triggering the prediction until the prediction is completed. This includes copying data between devices, and all steps of the inference pipeline, including the computation of zoom steps, determining what boxes to use for refinement, etc.}
\centering
\label{table:appendix_runtime}
\resizebox{\textwidth}{!}{
\begin{tabular}{l|llrrrrr}
Image & Target & Interaction & \multicolumn{1}{l}{Zoom out} & \multicolumn{1}{l}{Refinement Boxes} & \multicolumn{1}{l}{Time AutoZoom (ms)} & \multicolumn{1}{l}{Time Refinement (ms)} & \multicolumn{1}{l}{Time Total (ms)} \\ \hline
\multirow{4}{*}{\begin{tabular}[c]{@{}l@{}}Image: Dolichoderus\_mariae2\\ Dataset: InsectAnatomy\\ Modality: MicroCT\\ Size: 260x260x99\end{tabular}} & \multirow{4}{*}{Insect Brain} & lasso & 1.38 & 5 & 240 & 610 & 850 \\
 &  & scribble & 1.5 & 5 & 360 & 700 & 1050 \\
 &  & 2D bbox & 1.39 & 6 & 160 & 890 & 1050 \\
 &  & point & 1; 1; 2.25; 1 & 0; 0; 4; 0 & 130; 120; 450; 150 & 0; 0; 540; 0 & 130; 120; 990; 150 \\ \hline
\multirow{4}{*}{\begin{tabular}[c]{@{}l@{}}Image: lnq\_0006\\ Dataset: LNQ\\ Modality: CT\\ Size, spacing: 512x512x118, 0.9x0.9xmm\end{tabular}} & \multirow{4}{*}{Lymph Node} & lasso & 1 & 0 & 160 & 0 & 160 \\
 &  & scribble & 1 & 0 & 180 & 0 & 180 \\
 &  & 2D bbox & 1 & 0 & 150 & 0 & 150 \\
 &  & point & 1 & 0 & 160 & 0 & 160 \\ \hline
\multirow{4}{*}{\begin{tabular}[c]{@{}l@{}}Image: Mets\_019\\ Dataset: BrainMetShare\\ Modality: T1 spin-echo post\\ Size, spacing: 256x256x133, 0.94x0.94x1mm\end{tabular}} & \multirow{4}{*}{Brain Metastasis} & lasso & 1 & 0 & 160 & 0 & 160 \\
 &  & scribble & 1 & 0 & 140 & 0 & 140 \\
 &  & 2D bbox & 1 & 0 & 130 & 0 & 130 \\
 &  & point & 1 & 0 & 150 & 0 & 150 \\ \hline
\multirow{20}{*}{\begin{tabular}[c]{@{}l@{}}Image: s0360\\ Dataset: Totalsegmentator MRI v2\\ Modality: MRI\\ Size, spacing: 512x512x96, 0.72x0.72x1.7mm\end{tabular}} & \multirow{4}{*}{Liver} & lasso & 1.59 & 4 & 190 & 580 & 770 \\
 &  & scribble & 1.73 & 5 & 340 & 720 & 1060 \\
 &  & 2D bbox & 1.54 & 5 & 190 & 720 & 910 \\
 &  & point & 1.5 & 5 & 380 & 660 & 1040 \\
 & \multirow{4}{*}{Spleen} & lasso & 1.39 & 3 & 160 & 450 & 610 \\
 &  & scribble & 1.52 & 3 & 290 & 440 & 740 \\
 &  & 2D bbox & 1.36 & 3 & 170 & 400 & 570 \\
 &  & point & 1.5 & 3 & 380 & 460 & 850 \\
 & \multirow{4}{*}{Aorta} & lasso & 1 & 0 & 170 & 0 & 170 \\
 &  & scribble & 1 & 0 & 150 & 0 & 150 \\
 &  & 2D bbox & 1 & 0 & 150 & 0 & 150 \\
 &  & point & 1 & 0 & 160 & 0 & 160 \\
 & \multirow{4}{*}{Pancreas} & lasso & 1.5; 1.5; 1 & 3; 1; 0 & 320; 310; 200 & 450; 130; 0 & 770; 440; 200 \\
 &  & scribble & 1.5; 1.5; 1 & 3; 2; 0 & 290; 270; 190 & 360; 260; 0 & 650; 540; 190 \\
 &  & 2D bbox & 1.5; 1; 1; 1 & 3; 0; 0; 0 & 290; 130; 170; 130 & 360; 0; 0; 0 & 650; 130; 170; 130 \\
 &  & point & 2.25; 1; 1; 1; 1 & 1; 0; 0; 0; 0 & 620; 150; 120; 150; 130; & 210; 0; 0; 0; 0 & 830; 150; 120; 150; 130 \\
 & \multirow{4}{*}{Kidney L} & lasso & 1 & 0 & 170 & 0 & 170 \\
 &  & scribble & 1 & 0 & 150 & 0 & 150 \\
 &  & 2D bbox & 1 & 0 & 160 & 0 & 160 \\
 &  & point & 1 & 0 & 160 & 0 & 160 \\ \hline
\multirow{20}{*}{\begin{tabular}[c]{@{}l@{}}Image: liver\_201\\ Dataset: Task 3 MSD\\ Modality: CT\\ Size, spacing: 512x512x186, 0.77x0.77x2.5mm\end{tabular}} & \multirow{4}{*}{Liver} & lasso & 1.91 & 5 & 400 & 760 & 1160 \\
 &  & scribble & 2.21 & 5 & 560 & 660 & 1210 \\
 &  & 2D bbox & 1.77 & 5 & 330 & 780 & 1110 \\
 &  & point & 2.25 & 4 & 740 & 540 & 1290 \\
 & \multirow{4}{*}{Spleen} & lasso & 1 & 0 & 130 & 0 & 130 \\
 &  & scribble & 1 & 0 & 190 & 0 & 190 \\
 &  & 2D bbox & 1 & 0 & 160 & 0 & 160 \\
 &  & point & 1 & 0 & 180 & 0 & 180 \\
 & \multirow{4}{*}{Vertebra} & lasso & 1 & 0 & 170 & 0 & 170 \\
 &  & scribble & 1 & 0 & 160 & 0 & 160 \\
 &  & 2D bbox & 1 & 0 & 170 & 0 & 170 \\
 &  & point & 1 & 0 & 160 & 0 & 160 \\
 & \multirow{4}{*}{Pancreas} & lasso & 1 & 0 & 160 & 0 & 160 \\
 &  & scribble & 1 & 0 & 220 & 0 & 220 \\
 &  & 2D bbox & 1.04 & 1 & 210 & 190 & 400 \\
 &  & point & 1.5 & 1 & 390 & 170 & 560 \\
 & \multirow{4}{*}{Kidney L} & lasso & 1 & 0 & 160 & 0 & 160 \\
 &  & scribble & 1 & 0 & 180 & 0 & 180 \\
 &  & 2D bbox & 1 & 0 & 170 & 0 & 170 \\
 &  & point & 1 & 0 & 170 & 0 & 170 \\ \hline
\multirow{20}{*}{\begin{tabular}[c]{@{}l@{}}Image: liver\_141\\ Dataset: MSD Task 3\\ Modality: CT\\ Size, spacing: 512x512x971 0.7x0.7x0.5mm\end{tabular}} & \multirow{4}{*}{Liver} & lasso & 2.91 & 14 & 1520 & 2180 & 3700 \\
 &  & scribble & 3.54 & 14 & 1650 & 1890 & 3540 \\
 &  & 2D bbox & 2.69 & 14 & 1300 & 2300 & 3600 \\
 &  & point & 3.36; 1.5 & 7; 6 & 1980; 500 & 1160; 980 & 3150; 1490 \\
 & \multirow{4}{*}{Spleen} & lasso & 1.63 & 4 & 500 & 640 & 1130 \\
 &  & scribble & 1.5 & 4 & 530 & 640 & 1180 \\
 &  & 2D bbox & 1.67 & 3 & 600 & 570 & 1170 \\
 &  & point & 1.5 & 4 & 490 & 650 & 1140 \\
 & \multirow{4}{*}{Vertebra} & lasso & 1 & 0 & 180 & 0 & 180 \\
 &  & scribble & 1 & 0 & 150 & 0 & 150 \\
 &  & 2D bbox & 1 & 0 & 260 & 0 & 260 \\
 &  & point & 1 & 0 & 200 & 0 & 200 \\
 & \multirow{4}{*}{Pancreas} & lasso & 1.5 & 2 & 570 & 340 & 910 \\
 &  & scribble & 1.5 & 2 & 590 & 370 & 860 \\
 &  & 2D bbox & 1 & 1 & 190 & 0 & 190 \\
 &  & point & 1 & 1 & 190 & 0 & 190 \\
 & \multirow{4}{*}{Kidney L} & lasso & 1.5 & 3 & 490 & 560 & 1020 \\
 &  & scribble & 1.5 & 3 & 510 & 560 & 1070 \\
 &  & 2D bbox & 1.5 & 3 & 510 & 650 & 1130 \\
 &  & point & 1.5 & 3 & 500 & 510 & 1010
\end{tabular}
}
\end{table*}

\noindent Table \ref{table:appendix_runtime} shows inference times for nnInteractive as measured in our Napari Plugin. We test several images across multiple modalities, image sizes and voxel spacings. Tested objects are quite diverse ranging from known objects (organs, lymph nodes, tumors) to unknown ones (insect brain). Since nnInteractive processes objects at the original image resolution (no resampling) the size of objects in pixels is linked to the inference time. Small objects that fit within the patch size of 192x192x192 are predicted rapidly with inference times ranging from 120-200 ms. As objects become larger, AutoZoom is required to capture them in its entirety. This raises the inference time as a function of object size. In \textit{liver\_201}, the liver requires a zoom out factor of 1.77-2.25 and 4-5 refinement boxes. Given that the time required to process a refinement box is approximately the same as processing one object without zoom, it is no surprise that the total inference type ranges rises to 1110--1290ms. As a worst case, in images such as \textit{liver\_141} inference time can exceed 3500ms for large organs (here liver).

\noindent Row with multiple values in them indicate that several refinement steps were necessary to achieve acceptable segmentation accuracy for the respective structure. Notably, the pancreas of image \textit{s0360} was notoriously difficult to segment. Throughout user refinement, smaller areas of the object of interest need to be changed, requiring less zoom out and consequently fewer bounding boxes for refinement. Thus, the initial interaction typically takes the longest while further interactions are much quicker.

\noindent As can be seen in the point interaction on \textit{Dolichoderus mariae2}, the initial predictions were too localized and did not trigger AutoZoom, hinting at a deficit of point interactions in conveying informative guidance. Only after 2 additional interactions the model 'noticed' that the intended structure is much larger and triggered AutoZoom.

\noindent Throughout all datasets and targets we did not observe a measurable difference in inference speed across the supported interaction types.

\section{Additional results}\label{appendix:additional_results}

\subsection{AutoZoom}
\label{appendix:autozoom}
\begin{figure*}[!h]
  \begin{subfigure}[t]{0.32\textwidth}
    \includegraphics[width=\linewidth]{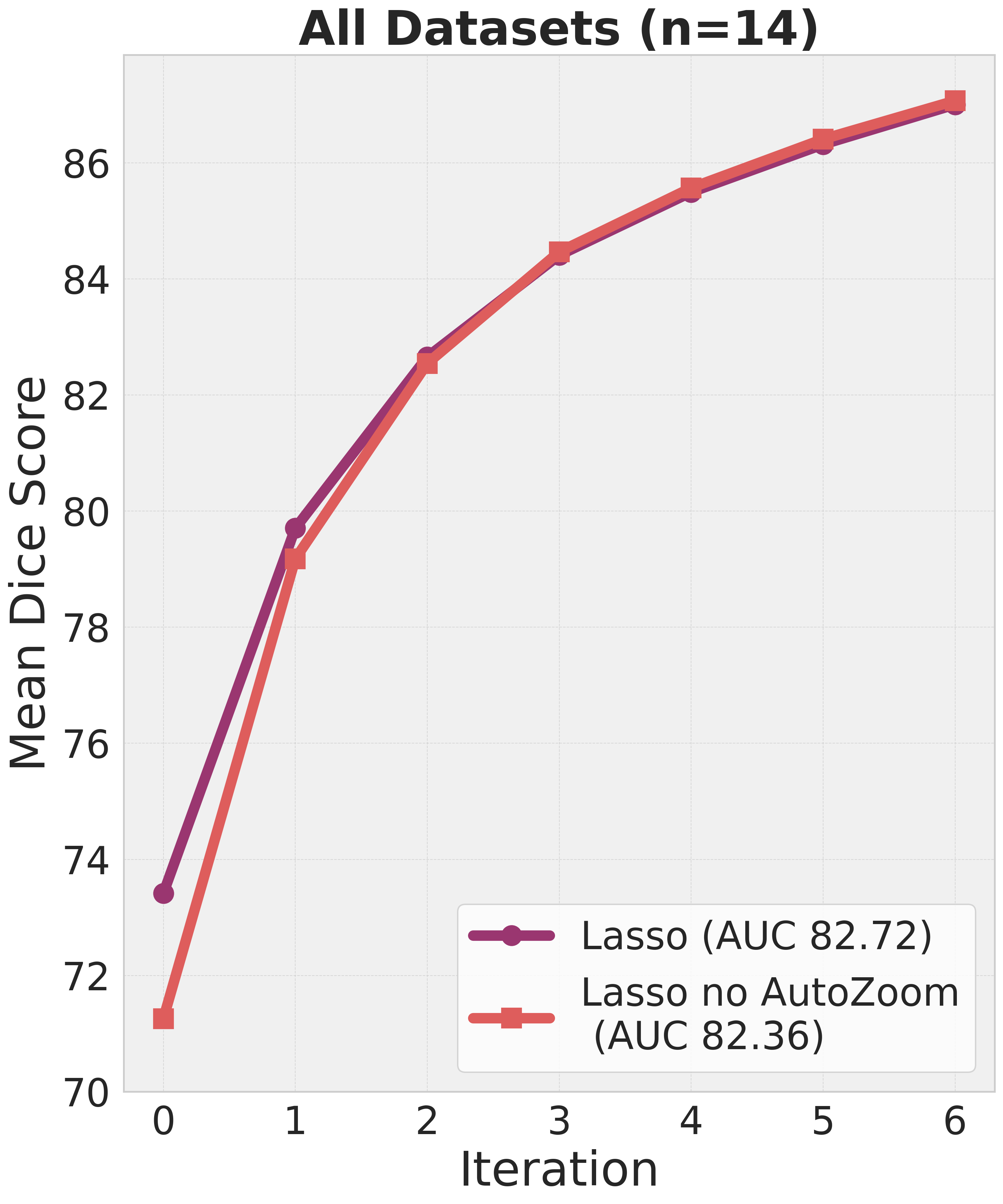}
    \caption{All test and OOD datasets}
    \label{fig:lasso_zoom_all}
  \end{subfigure}
  \hfill
  \begin{subfigure}[t]{0.32\textwidth}
    \includegraphics[width=\linewidth]{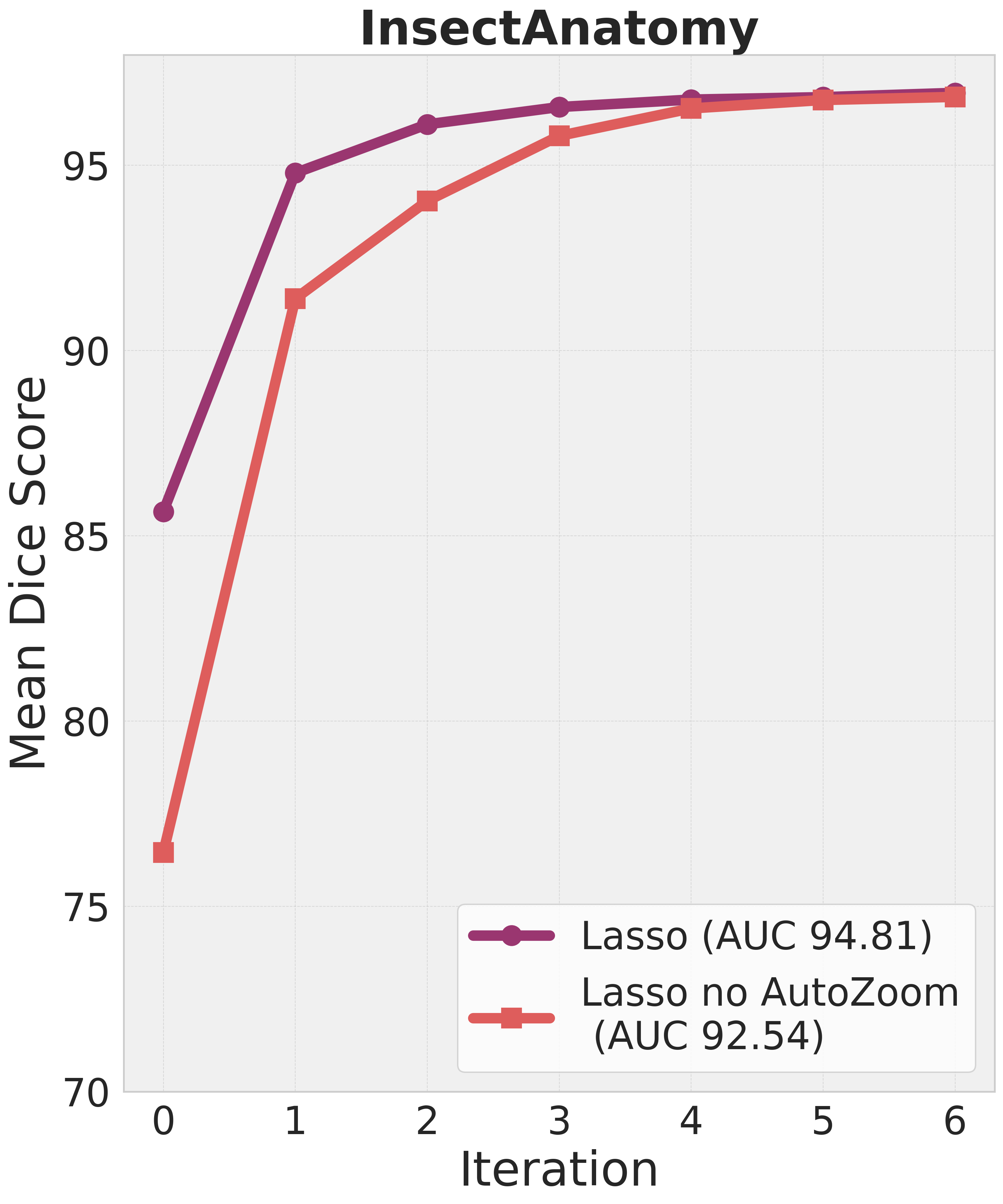}
    \caption{Insect brain only}
    \label{fig:lasso_zoom_d132}
  \end{subfigure}
  \hfill
  \begin{subfigure}[t]{0.32\textwidth}
    \includegraphics[width=\linewidth]{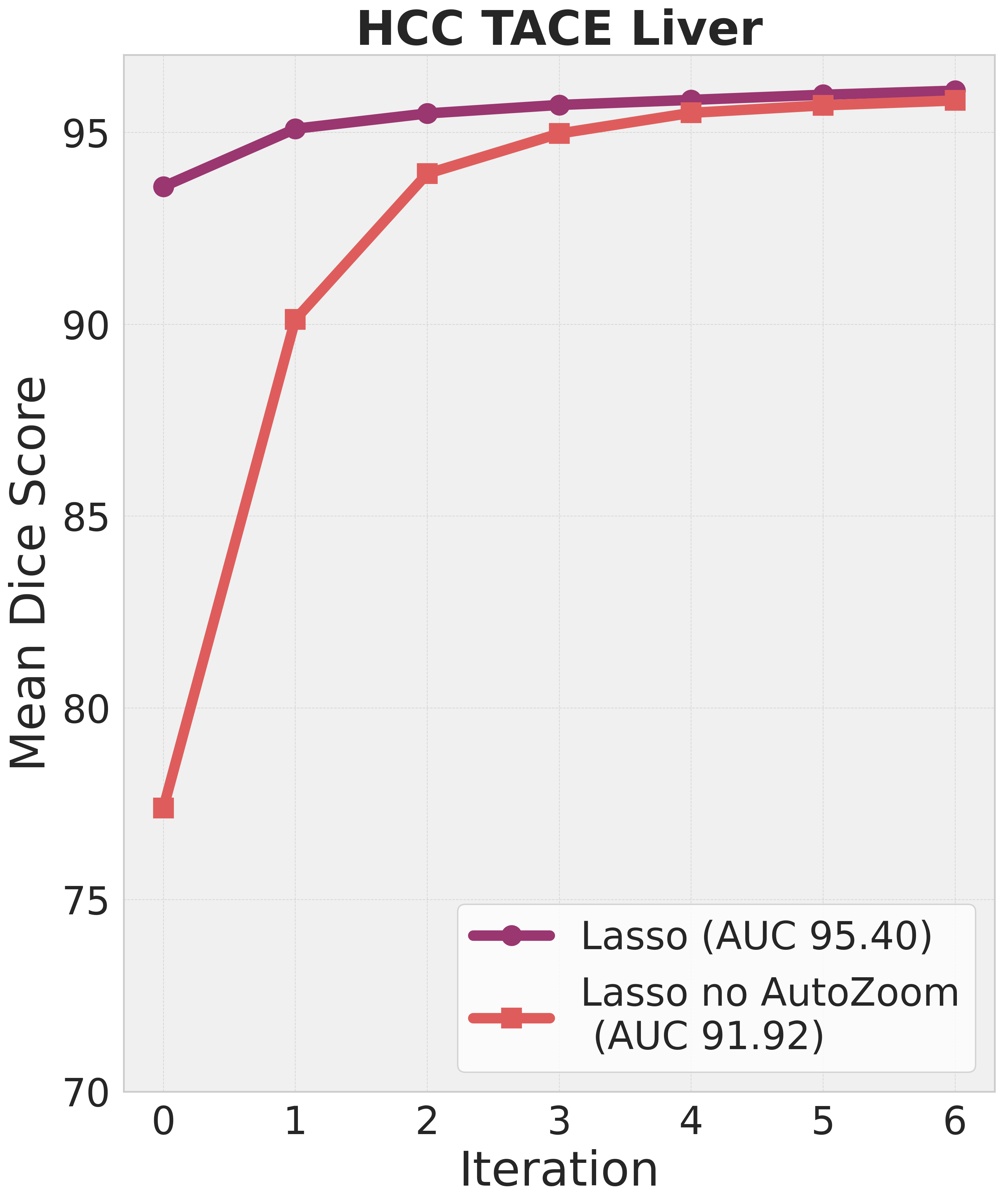}
    \caption{HCC TACE Liver only}
    \label{fig:lasso_zoom_d137}
  \end{subfigure}
  \caption{\textbf{The effect of AutoZoom on segmentation performance.} Averaged over all test and OOD datasets (a), the effect of AutoZoom is small, but noticeable. Since most target objects in these datasets are smaller than the nnInteractive patch size of 192x192x192 pixels they do not require AutoZoom, understating the impact of this contribution on large objects. When focussing on Datasets with large target objects, such as the Insect Brain of Ants (b) or the Liver in CT scans (c) AutoZoom becomes an essential feature. When AutoZoom is active, nnInteractive produces substantially improved results with fewer user interactions, achieving saturation much sooner than with AutoZoom disabled.}
  \label{fig:appendix_autozoom}
\end{figure*}

As shown in Fig. \ref{fig:appendix_autozoom}, AutoZoom offers substantial benefits for large objects where high Dice scores are achieved with fewer iterations. Particularly on large objects such as livers in CT scans, AutoZoom is an essential feature.

\subsection{Qualitative Results}
\begin{figure*}[h!]
\begin{center}
\includegraphics[width=0.93\linewidth]{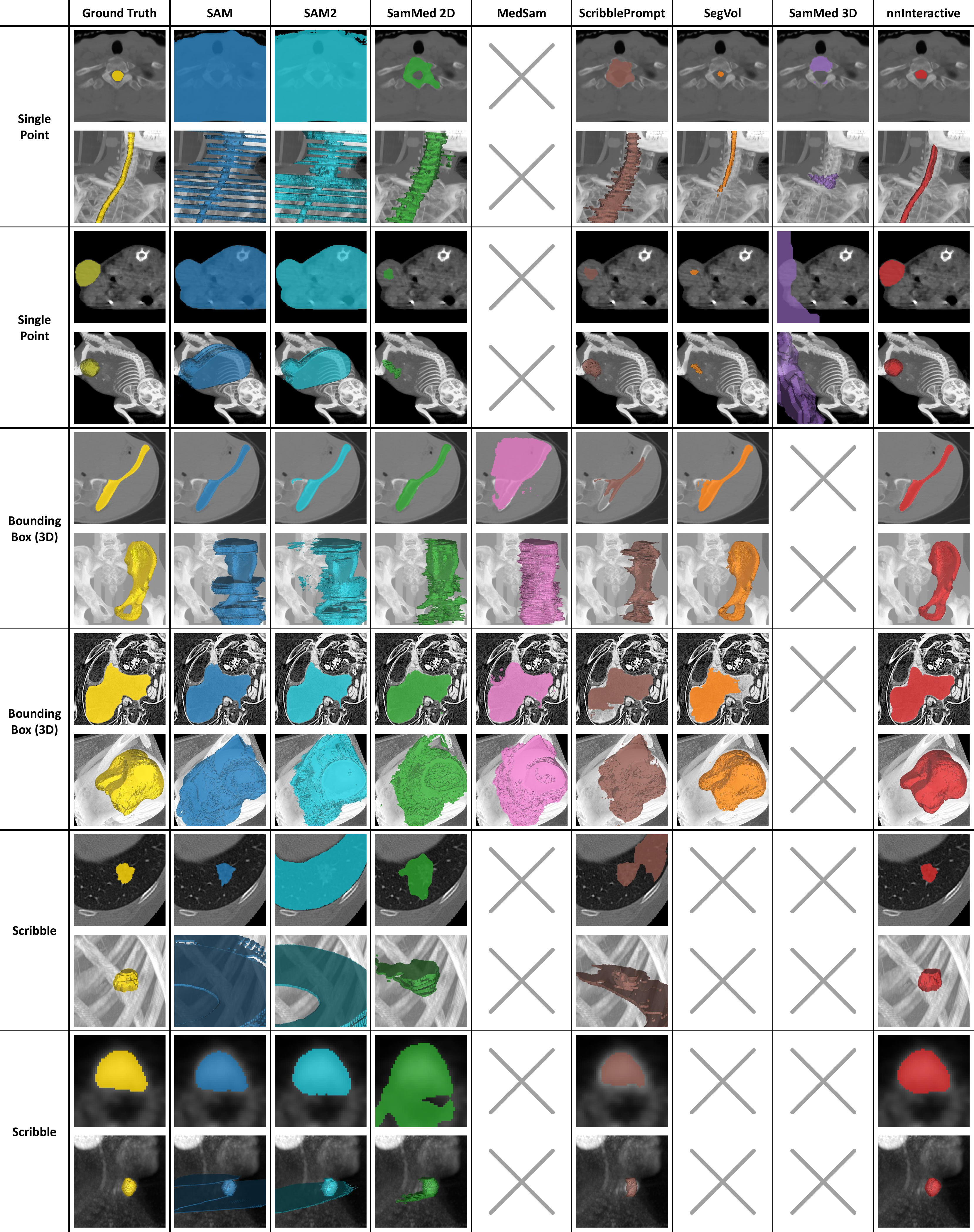}
\end{center} 
\caption{\textbf{Qualitative Results.} For each example the top row shows a 2D cross-section of the target and the bottom row shows a 3D rendering of the segmented structure. Crossed fields \(X\) denote unsupported prompting style for the respective method}
\label{fig:qualitative_appendix}
\end{figure*}

Additional qualitative results are shown in Fig. \ref{fig:qualitative_appendix}. Throughout all objects and tested prompting styles, nnInteractive consistently delivers maximum segmentation accuracy while competing methods fall behind, often producing severe artifacts. 2D methods in particular suffer from inconsistent results between slices, causing major drop in measured performance. Following nnInteractive, SegVol achieves the second best results.

\begin{figure*}[t]
  \centering
  \includegraphics[width=\textwidth]{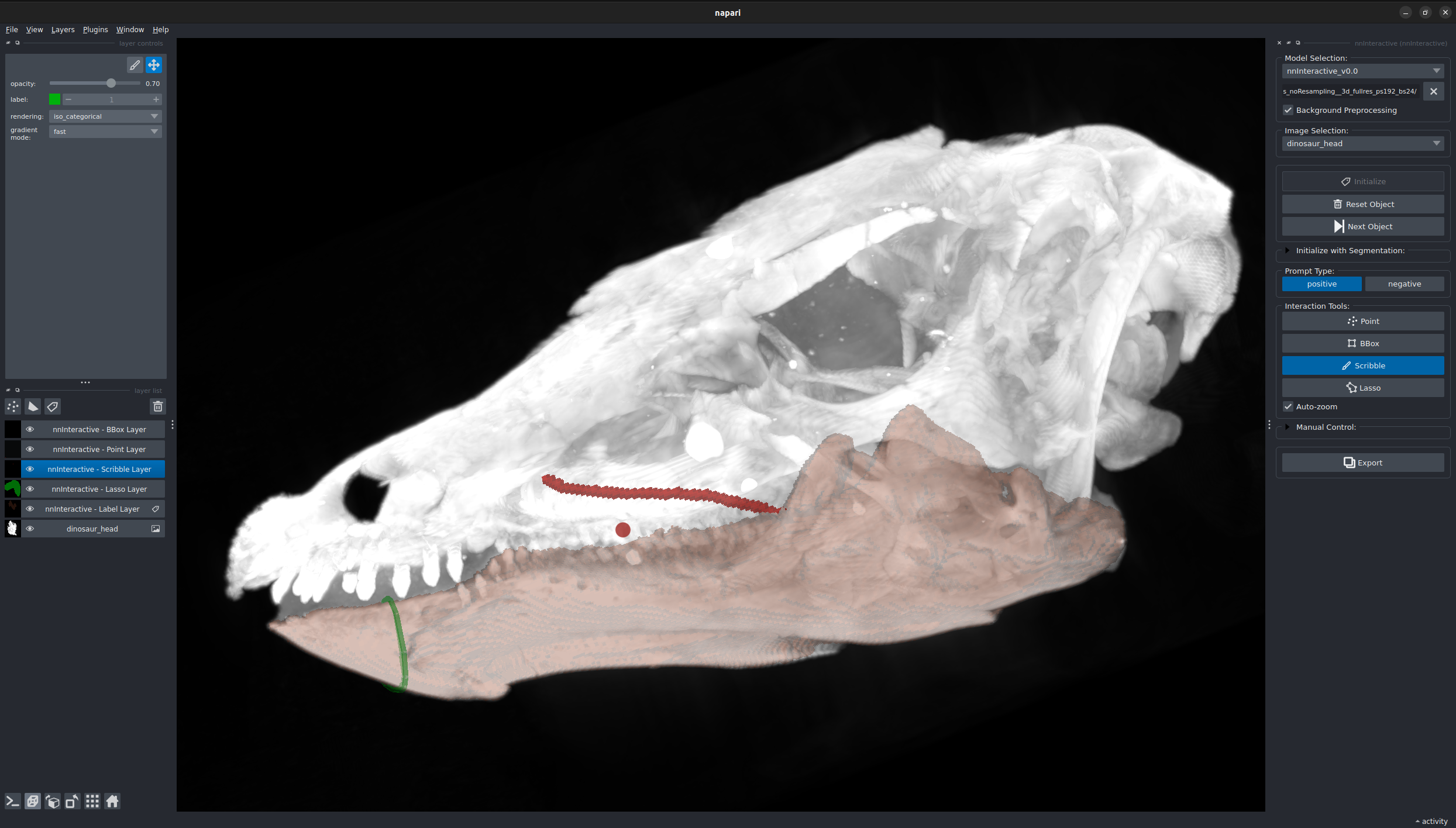}
  \caption{\textbf{nnInteractive on out-of-distribution data.} Segmentation of dinosaur (\textit{Thescelosaurus neglectus}~\cite{dino_head}) jawbones using the proposed Napari plugin with a few intuitive prompts (lasso, scribbles, points). Green areas indicate positive prompts while dark red highlights negative prompting. This demonstrates nnInteractive’s ability to generalize to unseen, non-medical data.}
  \label{fig:dino_head}
\end{figure*}

\begin{figure*}[h!]
  \centering
  \includegraphics[width=\textwidth]{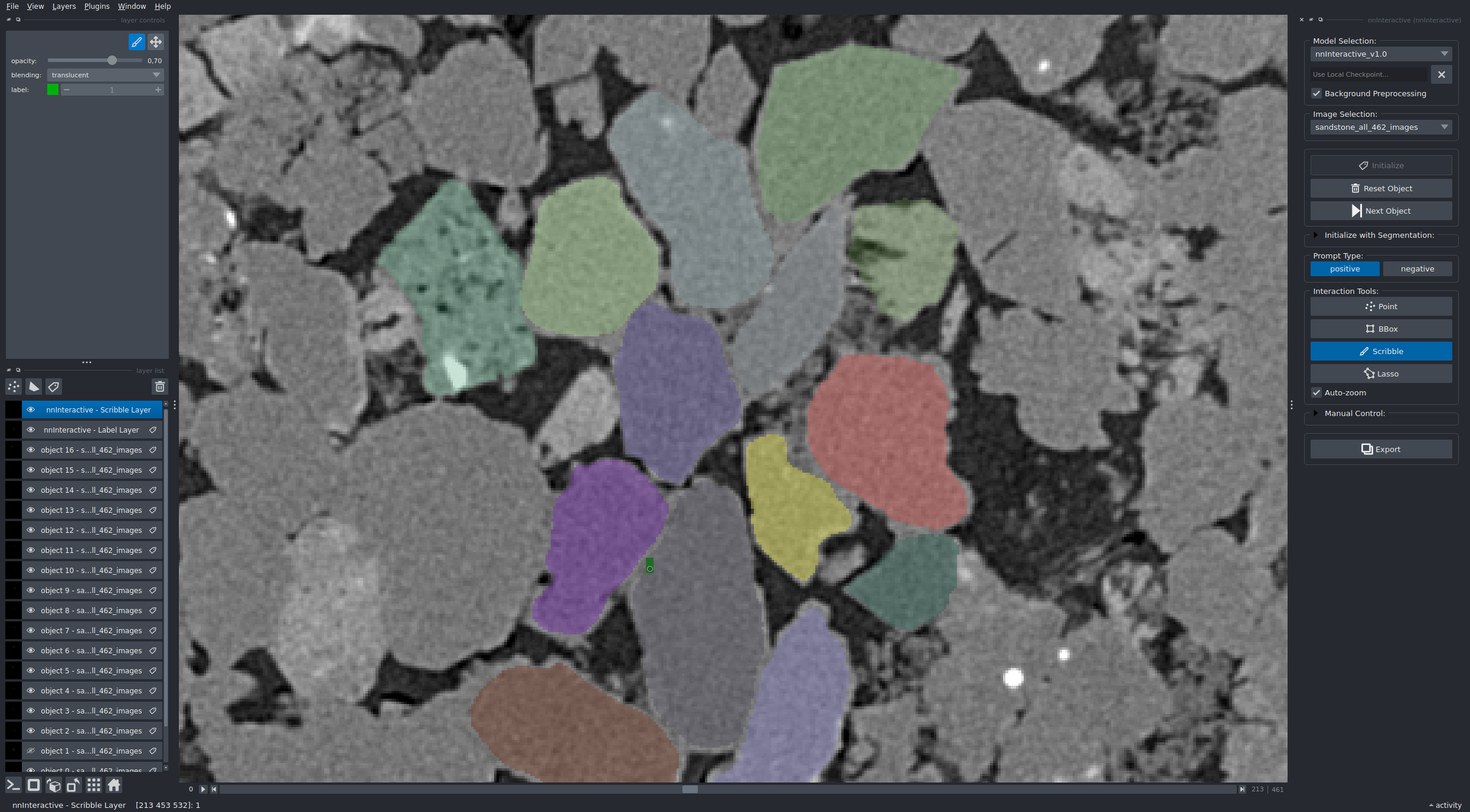}
  \caption{\textbf{nnInteractive on out-of-distribution data.} MicroCT of sandstone (data source: \href{http://youtube.com/post/Ugw86-dalhRybHVqq0J4AaABCQ?si=cZx9zOlSazMPXnQF}{here}). nnInteractive successfully separates individual grains despite the borders being barely visible to the human eye.}
  \label{fig:sandstone}
\end{figure*}

\noindent nnInteractive excels even in far OOD cases such as segmenting the jaw bone of a dinosaur fossil (Fig. \ref{fig:dino_head}) and individual grains from sandstone (Fig. \ref{fig:sandstone}).

\end{document}